\documentclass[journal]{IEEEtran}
\ifCLASSINFOpdf
\else
\fi
\usepackage{graphicx}
\usepackage{amsmath,amssymb,amsthm}
\usepackage{epstopdf}
\usepackage{algorithm}
\usepackage{algpseudocode}
\usepackage{epsfig}
\usepackage{stackengine}
\newtheorem{definition}{Definition}
\begin{document}

\title{Detecting Approximate Reflection Symmetry in a Point Set using Optimization on Manifold}

\author{\IEEEauthorblockN{Rajendra Nagar\IEEEauthorrefmark{1} and
		Shanmuganathan Raman\IEEEauthorrefmark{2}\\}
	\IEEEauthorblockA{Electrical Engineering,
		Indian Institute of Technology Gandhinagar, India, 382355\\
		Email: \IEEEauthorrefmark{1}rajendra.nagar@iitgn.ac.in,
		\IEEEauthorrefmark{2}shanmuga@iitgn.ac.in}}

\maketitle
\begin{abstract}
We propose an algorithm to detect approximate reflection symmetry present in a set of volumetrically distributed points belonging to $\mathbb{R}^d$ containing a distorted reflection symmetry pattern. We pose the problem of detecting approximate reflection symmetry as the problem of establishing correspondences between the points which are reflections of each other and we determine the reflection symmetry transformation. We formulate an optimization framework in which the problem of establishing the correspondences amounts to solving a linear assignment problem and the problem of determining the reflection symmetry transformation amounts to solving an optimization problem on a smooth Riemannian product manifold. The proposed approach estimates the symmetry from the geometry of the points and is descriptor independent. We evaluate the performance of the proposed approach on the standard benchmark dataset and achieve the state-of-the-art performance. We further show the robustness of our approach by varying the amount of distortion in a perfect reflection symmetry pattern where we perturb each point by a different amount of perturbation. We demonstrate the effectiveness of the method by applying it to the problem of 2-D and 3-D reflection symmetry detection along with comparisons.
\end{abstract}


\IEEEpeerreviewmaketitle

\section{Introduction}\label{sec:introduction}
\begin{figure}[htbp]
\centering
	\stackunder{\includegraphics[width=0.44\linewidth]{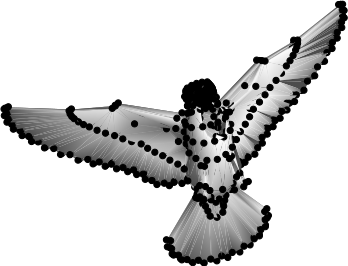}}{(a)}
	\stackunder{\includegraphics[width=0.54\linewidth]{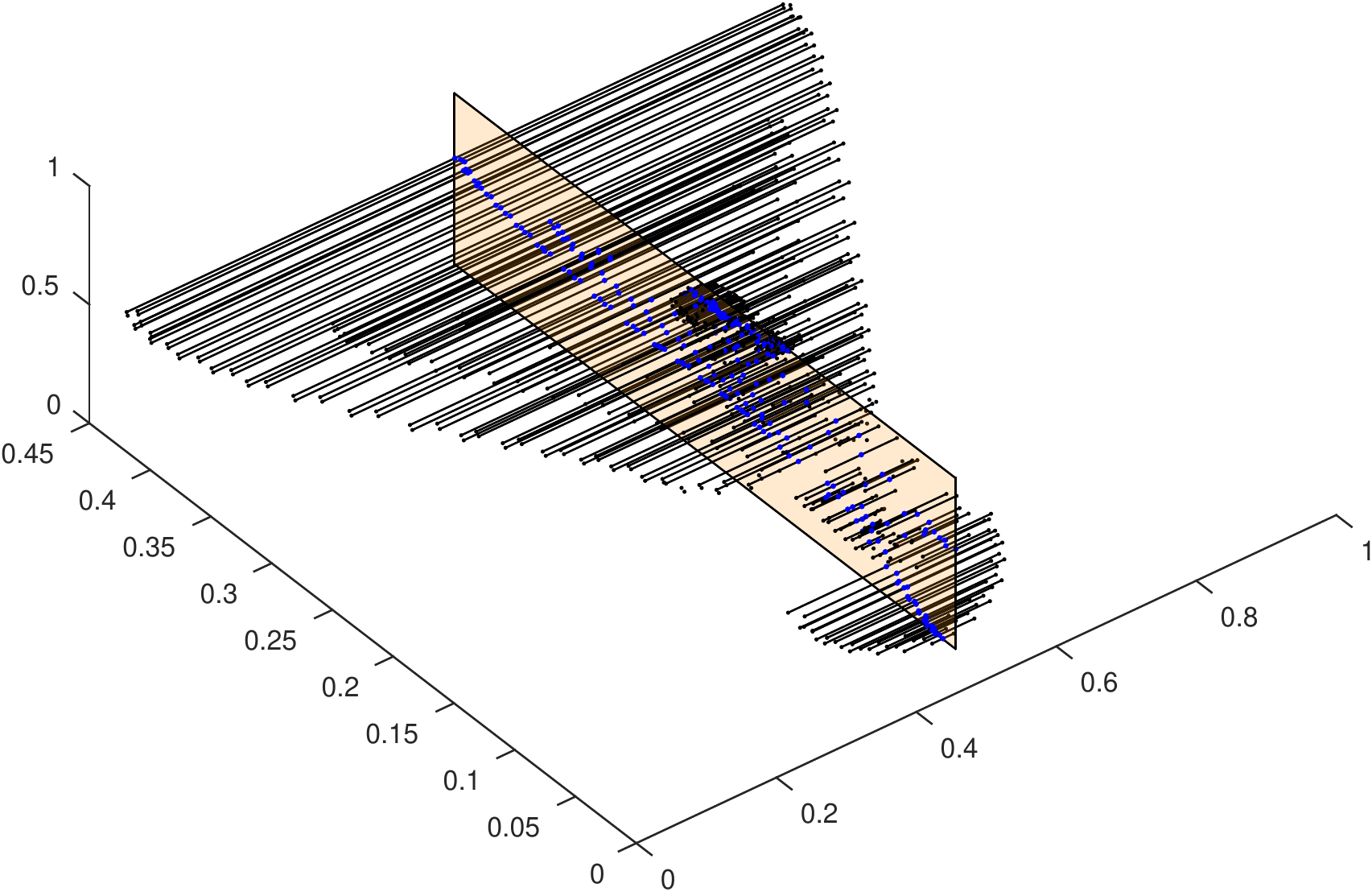}}{(b)}
	\caption{ Established correspondences (shown in (b)) between the reflection symmetry points sampled from the input model (shown in (a)) using the proposed approach.}
	\label{fig:intro_fig}
\end{figure}

\IEEEPARstart{S}{ymmetry} present in natural and man-made objects enriches the objects to be physically balanced, beautiful, easy to recognize, and easy to understand. Characterizing and finding the symmetry has been an active topic of research in computer vision and computer graphics as physical objects form the basis for these research areas. The digitized objects are mainly represented in the form of meshes, volumes, sets of points, and images. The primary objective has been to detect symmetry in objects depicted through these different representations. We particularly aim to detect reflection symmetry present in objects represented by a set of finite number of points belonging to $\mathbb{R}^d$. In Fig. \ref{fig:intro_fig}, we present an example result of the proposed approach for illustration. 
 
 The motivation behind detecting symmetry in higher dimensional spaces ($d>3$) is inspired by the fact that many physical data points reside in the space of dimensions greater than three. For example, an RGB-D image captured using a Kinect sensor, which has become a major tool for interaction of human with machine, has four dimensions at each pixel location. Another example is the embedding of feature points or shapes into a higher-dimensional space. In the scale invariant feature transform (SIFT) algorithm, each keypoint is represented in a 128-dimensional space \cite{lowe2004distinctive}. We not only target data residing in 2-D (image) and 3-D (point cloud), but also develop a generic  framework to detect symmetry in higher dimensional data.

The problem of establishing correspondences between reflection symmetry points and determining the hyperplane of reflection symmetry has been extensively studied due to its astounding applications such as compression of objects, symmetrization, shape matching, and symmetry aware segmentation of shapes \cite{mitra2013symmetry}. Most of the existing algorithms attempt this problem by using surface signatures such as Gaussian curvature, eigenbases of the Laplace-Beltrami operator, and heat kernels for the points sampled on a given surface (\cite{berner2008graph,mitra2006partial,mitra2013symmetry}). The challenge we face is that, we only have a set of discrete points in $\mathbb{R}^d$. We can not take benefits from local surface signatures by fitting a surface over these points. For the case $d=2$, an explanation could be that the prominent surface signatures, such as Gaussian curvatures, are meaningful only if the surface is non-linear. For the case $d\geq 3$, an explanation could be that if the point set represents a volumetric shape, fitting a surface could be hard and eigenbases of Laplace-Beltrami operator are not defined for a set of finite points since it is not a compact manifold without the boundary \cite{ovsjanikov2008global}. Prominent methods such as \cite{lipman2010symmetry} and \cite{xu2012multi} are independent of surface features and employ randomized algorithms to establish correspondences between the reflection symmetry points. However, they require fine tuning of a hyper-parameter to handle the reflection symmetry patterns perturbed by an unknown source of noise and an improper choice of this parameter could lead to higher time complexity. 

Both these categories of algorithms are sequential in the sense that they first establish the correspondences between the reflection symmetry points and then determine the reflection symmetry hyperplane. Therefore, many outlier correspondences could be detected along with the correct correspondences. In summary, detecting symmetry in a set containing a finite number of points is a non-trivial problem.  In this work, we propose an optimization framework where we jointly establish correspondences between reflection symmetry points and determine the reflection symmetry hyperplane in a set of points containing a distorted reflection symmetry pattern. In order to design the cost function, we introduce an affine transformation to obtain the reflection point of a point in $\mathbb{R}^d$. The main intuition behind forming this cost function is that the reflection point of a point obtained through the optimal reflection hyperplane should be present closest to its ground-truth reflection point.

The primary contributions of this work are listed below.
\begin{enumerate}
	\item We propose an optimization based algorithm to establish correspondences between the reflection symmetry points and determine the reflection symmetry transformation in a set of discrete points residing in $\mathbb{R}^d$ containing a distorted reflection symmetry pattern.
	\item We show that the proposed optimization framework is convex in translation and correspondences matrix, and \textit{locally convex} in each of the rotation matrices. 
	\item The proposed approach is shown to not use any shape descriptors and can be applied to point sets obtained by sampling any shape residing in  $\mathbb{R}^d$.
	\item  We demonstrate the effectiveness of the proposed approach by detecting symmetry in 2-D images and 3-D point clouds.
\end{enumerate}
We organize the remainder of the paper as follows. In \S \ref{sec:rw}, we present the related works to our approach. In \S\ref{subsec:PF}, we formulate the energy minimization problem. In \S\ref{subsec:Rt}, we find the optimal rotations and translation. In \S\ref{subsec:PI}, we find the optimal mirror symmetric correspondences. In \S\ref{subsec:CA}, we prove the convergence properties. In \S\ref{sec:cc}, we report the computational complexity of our algorithm. In \S\ref{sec:RD}, we report the results and the evaluation of the proposed approach. In \S\ref{sec:CF}, we conclude the work with future directions.
\section{Related Works}
\label{sec:rw}
The problem of characterizing and detecting the reflection symmetry in digitized objects has been extensively studied. The works \cite{liu2010computational} and \cite{mitra2013symmetry} provide a survey of the symmetry detection algorithms. The symmetry detection algorithms can be categorized based on either the form of the input data or whether the algorithm is dependent or independent of the surface features. General forms of the input data are: set of points, mesh, volume, and image. Most of the methods for symmetry detection in meshes first extract salient keypoints on the surface and then describe each point using local surface features. The prominent surface features are: Gaussian curvatures, slippage features, moments, geodesic distances, and extended Gaussian images  (\cite{liu2010computational,mitra2013symmetry}). 

\textbf{ Symmetry detection in a set of points without features.}
These algorithms detect reflection symmetry in a set of points without using surface features. Our work also falls in this category. In the work by Zabrodsky \emph{et al.}, the authors find the closest shape to a given shape represented by a set of points in $\mathbb{R}^2$  and it requires point correspondences \cite{zabrodsky1995symmetry}. However, our goal is different in the sense that we find reflection correspondences within the given set of points in $\mathbb{R}^d$. In the work by Lipman \emph{et al.}, the authors propose the concept of symmetry factored embedding where they represent pairs of points which are in the same orbit in a new space and propose the concept of symmetry factored distance to find the distance between such pairs \cite{lipman2010symmetry}. In the work by Xu \emph{et al.}, the authors detect multi-scale symmetry \cite{xu2012multi}. The authors use a randomized algorithm to detect the correspondences efficiently. However, performance degrades as the perfect pattern gets perturbed due to noisy measurements. We compare the correspondences established by our method to that of this method and show that our method performs better than this method when the patterns are perturbed. It is fair to compare with this method on the perturbed patterns because most of the real world patterns are not perfectly symmetric, e.g., human face and butterfly wings. In the works by Comb{\`e}s \emph{et al.} \cite{combes2008automatic}, Speciale \emph{et al.} \cite{speciale2016symmetry}, Ecins \emph{et al.} \cite{ecins2017detecting},  Cicconet \emph{et al.} \cite{Cicconet_2017_ICCV_Workshops}, Li \emph{et al.} \cite{li2016efficient}, and Sipiran \emph{et al.} \cite{sipiran2014approximate}, the authors automatically detect the symmetry plane in a point cloud. But, the methods in \cite{combes2008automatic}, \cite{speciale2016symmetry}, \cite{Cicconet_2017_ICCV_Workshops}, \cite{li2016efficient}, and \cite{sipiran2014approximate} do not establish correspondences. However, correspondences are an important aspect as shown in (\cite{xu2012multi,lipman2010symmetry}). Ecins \emph{et al.} proposed an ICP based approach \cite{ecins2017detecting} where they used the normals at each point to determine the symmetry. However, this method is applicable only to non-volumetric point clouds, i.e., points sampled from a surface.\\
\textbf{Symmetry detection in meshes using surface features.} 
These algorithms either directly use surface patches described using local features or first detect the salient keypoints on the surface and describe them using the local surface features. Here, we review only the salient works to give an idea of these algorithms. Mitra \emph{et al.} detect partial and approximate symmetries in 3D models \cite{mitra2006partial}. They start with sampling salient keypoints on the surface and pair them up using their local principal curvatures. Then using the Hough transformation, they find the pairs of reflection symmetry points. Then in the Hough transformation space, they perform the clustering of the pairs to determine all the partial symmetries.  Martinet \emph{et al.} detect symmetries by generalized moment functions where the shape symmetry gets inherited as symmetry in these functions \cite{martinet2006accurate}. Raviv \emph{et al.} detect symmetry in non-rigid shapes by observing that the intrinsic geometry of a shape is invariant under non-rigid shape transformations \cite{RavivBBK07}. Berner \emph{et al.} start with constructing a graph based on the similarity of slippage features detected on the surface \cite{berner2008graph}. Then, they detect the structural regularities by matching the sub-graphs. Cohen \emph{et al.} detect symmetry using  geometric and image cues \cite{cohen2012discovering}. They use it to reconstruct accurate 3D models. We refer the reader to some of the pioneering works for more details on this category (\cite{bokeloh2009symmetry,lasowski2009probabilistic,mitra2007symmetrization,pauly2008discovering,thomas2014multiscale,kazhdan2002reflective,liu2017detection,shi2016symmetry,wang2017group}). There exist algorithms which find symmetry in meshes and volume without sampling keypoints. The  works described in  (\cite{kim2010mobius,mitra2010intrinsic,ovsjanikov2008global,podolak2006planar,raviv2010full,thrun2005shape,xu2009partial,sun19973d,jiang2013skeleton}) belong to this category.

\textbf{Symmetry detection algorithm for real images.} These algorithms primarily rely on the local image features such as edge orientations, curvatures, and gradients. The recent works such as (\cite{Lukac17-SIG,levinshtein2013multiscale,sie2013detecting,loy2006detecting,teo2015detection,hauagge2012image,wang2015reflection,patraucean2013detection,tsogkas2012learning,chertok2010spectral,atadjanov2016reflection,koser2011dense}) present excellent algorithms for reflection symmetry detection in images. Given the accurate detection of keypoints, the algorithm developed in this work can be used to detect reflection symmetry in images without using local features.

Our algorithm is similar to  Iterative Closest Point (ICP) algorithm (\cite{besl1992method,rusinkiewicz2001efficient}) only in the sense that we also follow the alternation between the optimization of reflection transformation (rotation and translation in ICP) and the correspondences between the mirror symmetric points (correspondences between the points of two different shapes in ICP). Our algorithm differs from the ICP algorithm  since ICP has a different error function in the transformation parameters than the error function of our problem. Furthermore, our matching is bijective since we impose the bijectivity constraints in our optimization framework. These constraints ensure that each point has exactly one mirror image point.

\section{Proposed Approach}
\label{sec:PA}
\begin{figure}
	\centering
	\epsfig{figure=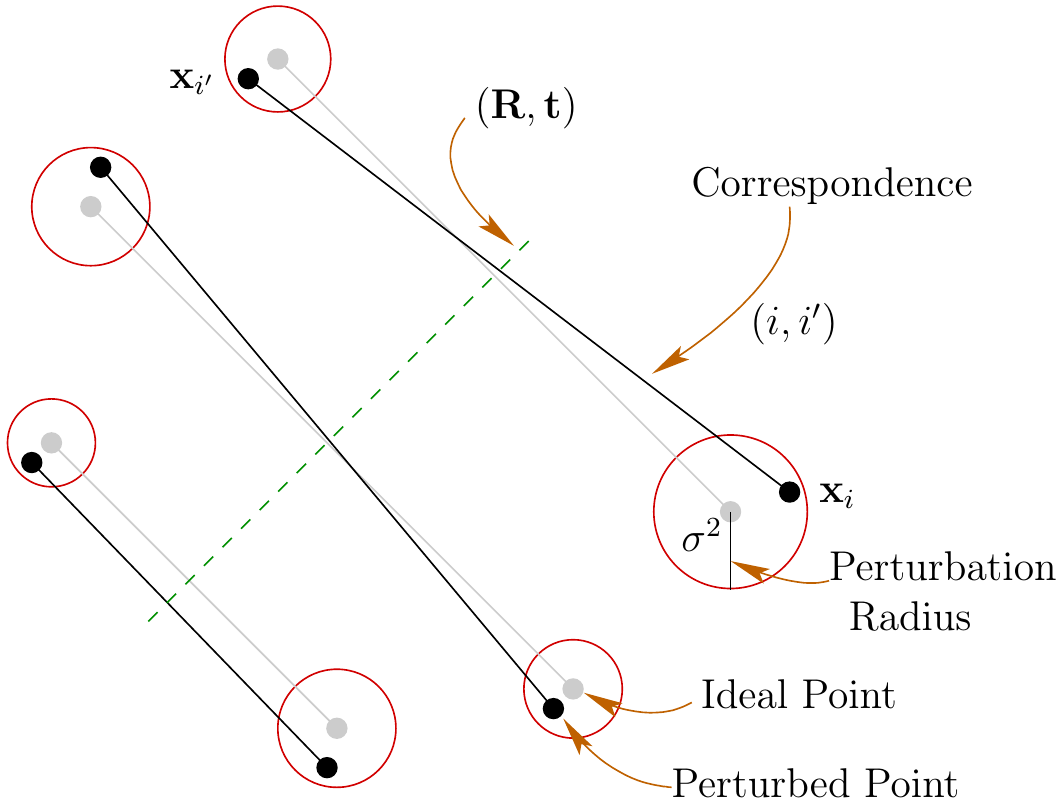,width=0.9\linewidth}
	\caption{Each point of a perfect pattern, shown in gray color, is perturbed to a point within a circular region around it where the radius is different for all the points and is unknown. Our goal is to determine the correspondences $(i,i^\prime)$ and the reflection transformation $(\mathbf{R,t})$.}
	\label{fig:intro}
\end{figure}

Consider a set $\mathcal{S}=\{\mathbf{x}_i\}_{i=1}^{n}$  of points, where $\mathbf{x}_i\in\mathbb{R}^d$, containing a distorted reflection symmetry pattern. Our goal is to determine the reflection symmetry transformation and establish the correspondences between the reflection symmetry points. In Fig. \ref{fig:intro}, we show the graphical representation of our problem. We formulate an optimization framework in which both the correspondences between reflection symmetry points and the reflection symmetry transformation are variables as described below. We use the notation $[k]$ for the set $\{1,2,\ldots,k\}$, where $k$ is a natural number.
\subsection{Problem Formulation}
\label{subsec:PF}
We introduce reflection transformation in $\mathbb{R}^d$ in order to obtain the reflection of a point through a hyperplane $\boldsymbol{\pi}$, not necessarily passing through the origin. The intuition is based on the fact that any hyperplane in $\mathbb{R}^d$ is a $d-1$ dimensional \textit{subspace}. Therefore, it can be made parallel to the subspace spanned by any $d-1$ coordinate axes by translating the origin of the coordinate system on the hyperplane $\boldsymbol{\pi}$ and then rotating these $d-1$ axes sequentially (by the angle between the hyperplane $\boldsymbol{\pi}$ and the axis). In this new coordinate system, the reflection of a point through the hyperplane $\boldsymbol{\pi}$ can be obtained by multiplying the coordinate corresponding to the remaining axis of the point by $-1$. Then the reflection in the original coordinate system is obtained by applying the inverse procedure.
\begin{definition}
The reflection point $\mathbf{x}_{i^\prime}\in \mathbb{R}^d$ of a point $\mathbf{x}_i\in \mathbb{R}^d$ through the reflection symmetry hyperplane $\boldsymbol{\pi}$ is determined by an affine transformation as shown in Equation \ref{eq:1}.
\begin{eqnarray}
\mathbf{x}_{i^\prime}&=&\left(\prod_{u=1}^{d-1}\mathbf{R}_u\right)\mathbf{E}\left(\prod_{u=1}^{d-1}\mathbf{R}_u\right)^\top(\mathbf{x}_i-\mathbf{t})+\mathbf{t}.
\label{eq:1}
\end{eqnarray}
\end{definition}
Here, $i,i^\prime\in[n]$, $\mathbf{t}\in\mathbb{R}^d$ is the translation vector which translates the origin of the coordinate system on the hyperplane $\boldsymbol{\pi}$, $\mathbf{R}_u$ is a rotation matrix of size $d\times d$ that rotates the $u^\text{th}$ axis about the origin such that it becomes perpendicular to the normal of the hyperplane $\boldsymbol{\pi}$, and the matrix $\mathbf{E}$ is defined as
$\mathbf{E}=\begin{bmatrix}
\mathbf{I}_{d-1}&\mathbf{0}_{d-1}\\\mathbf{0}_{d-1}^\top&-1
\end{bmatrix}$ and satisfies $\mathbf{E}^\top=\mathbf{E}$, $\mathbf{E}^\top\mathbf{E}=\mathbf{I}_d$. The matrix $\mathbf{R}_u$ is an orthogonal matrix ($\mathbf{R}_u^\top\mathbf{R}_u=\mathbf{R}_u\mathbf{R}_u^\top=\mathbf{I}_d$) with determinant equal to +1, $\forall u\in\{1,2,\ldots,d-1\}$.  Here, $\mathbf{0}_{d-1}$ is a vector of size $(d-1)\times 1$ with all the coordinates equal to zero and $\mathbf{I}_{d-1}$ is an identity matrix of size $(d-1)\times(d-1)$.

Now, we introduce the essential properties of this transformation in order to formulate the problem. We show that the rotation matrices $(\mathbf{R}_1,\ldots,\mathbf{R}_{d-1})$ and the translation vector $\mathbf{t}$ uniquely determine the reflection hyper-plane $\boldsymbol{\pi}$. We let $\mathbf{T}=\prod_{u=1}^{d-1}\mathbf{R}_u$ throughout this paper and note that it is again an orthogonal matrix with determinant equal to +1.

\textbf{Theorem 1:}
\textit{The point $\mathbf{x}_{i^\prime}$ is the reflection of the point $\mathbf{x}_i$ through the hyperplane $\boldsymbol{\pi}$ if and only if the point $\mathbf{x}_{i}$ is the reflection of the point $\mathbf{x}_{i^\prime}$ through the hyperplane $\boldsymbol{\pi}$.}\\
\textbf{Proof:}
We prove the forward direction of the Theorem 1, since the backward direction can be proved in a similar way. Let us assume that the point $\mathbf{x}_{i^\prime}$ is the reflection of the point $\mathbf{x}_i$. Therefore, Equation \eqref{eq:1} holds true. Now, we multiply Equation \eqref{eq:1} by $\mathbf{T}\mathbf{E}\mathbf{T}^\top$  from left and use the identities $\mathbf{E}^\top=\mathbf{E}, \mathbf{E}\mathbf{E}=\mathbf{I}_d,\mathbf{T}^\top\mathbf{T}=\mathbf{T}\mathbf{T}^\top=\mathbf{I}_d$ to achieve,
\begin{equation}
\nonumber\mathbf{T}\mathbf{E}\mathbf{T}^\top\mathbf{x}_{i^\prime}=\mathbf{x}_i-\mathbf{t}+\mathbf{T}\mathbf{E}\mathbf{T}^\top\mathbf{t}
\end{equation}
\begin{equation}
\Rightarrow\mathbf{x}_i=\mathbf{T}\mathbf{E}\mathbf{T}^\top(\mathbf{x}_{i^\prime}-\mathbf{t})+\mathbf{t}.
\label{eq:2}
\end{equation}\qed

\textbf{Theorem 2:}
\textit{The normal vector of the reflection hyper-plane $\boldsymbol{\pi}$ lies in the null space of the matrix $\mathbf{I}_d+\mathbf{T}\mathbf{E}\mathbf{T}^\top$, the hyper-plane $\boldsymbol{\pi}$ passes through $\mathbf{t}$, and the null space of the matrix $\mathbf{I}_d+\mathbf{T}\mathbf{E}\mathbf{T}^\top$ is an one-dimensional subspace of $\mathbb{R}^d$}.\\
\textbf{Proof:}
We subtract Equation \eqref{eq:1} from Equation \eqref{eq:2} to achieve
$$
\mathbf{x}_i-\mathbf{x}_{i^\prime}=\mathbf{T}\mathbf{E}\mathbf{T}^\top(\mathbf{x}_{i^\prime}-\mathbf{x}_i)\Rightarrow(\mathbf{I}_d+\mathbf{T}\mathbf{E}\mathbf{T}^\top)(\mathbf{x}_i-\mathbf{x}_{i^\prime})=\mathbf{0}.
$$
Therefore, the normal to the reflection hyperplane $\boldsymbol{\pi}$, which is in the direction of the vector $(\mathbf{x}_i-\mathbf{x}_{i^\prime})$, lies in the null space of the matrix $(\mathbf{I}_d+\mathbf{T}\mathbf{E}\mathbf{T}^\top)$. It is easy to show that the reflection hyperplane $\boldsymbol{\pi}$ passes through the translation $\mathbf{t}$ by noting that the reflection point of the point $\mathbf{t}$ is $\mathbf{t}$. This is possible only if the point $\mathbf{t}$ lies on the reflection hyperplane $\boldsymbol{\pi}$. We prove that the null space of the matrix $\mathbf{I}_d+\mathbf{T}\mathbf{E}\mathbf{T}^\top$ is an one-dimensional subspace of $\mathbb{R}^d$ in order to show that there exists an unique hyperplane $\boldsymbol{\pi}$. The nullspace of a matrix is the space spanned by the eigenvectors corresponding to the zero eigenvalue. Let $\mathbf{p}=\begin{bmatrix}
p_1&p_2&\ldots&p_d\end{bmatrix}^\top\in\mathbb{R}^d$ be any vector. If $\mathbf{p}$ is an eigenvector corresponding to the zero eigenvalue of the matrix $\mathbf{I}_d+\mathbf{T}\mathbf{E}\mathbf{T}^\top$, then we must have
$$
\mathbf{p}^\top(\mathbf{I}_d+\mathbf{T}\mathbf{E}\mathbf{T}^\top)\mathbf{p}=0
\Rightarrow\mathbf{p}^\top\mathbf{I}_d\mathbf{p}+(\mathbf{T}^\top\mathbf{p})^\top\mathbf{E}(\mathbf{T}^\top\mathbf{p})=0
$$
$$
\Rightarrow\mathbf{p}^\top\mathbf{I}_d\mathbf{p}+\mathbf{b}^\top\mathbf{E}\mathbf{b}=0
\Rightarrow\sum_{u=1}^{d}p_u^2+\sum_{u=1}^{d-1}b_u^2-b_d^2=0
$$
\begin{equation}
\Rightarrow\sum_{u=1}^{d}p_u^2+\sum_{u=1}^{d}b_u^2-2b_d^2=0.
\label{eq:3}
\end{equation}

Here, $\mathbf{b}=\mathbf{T}^\top\mathbf{p}$. We note that $\|\mathbf{b}\|_2^2=(\mathbf{T}^\top\mathbf{p})^\top(\mathbf{T}^\top\mathbf{p})=\mathbf{p}^\top\mathbf{p}=\|\mathbf{p}\|_2^2$.  Therefore, from Equation \eqref{eq:3} we have
$$
\sum_{u=1}^{d}p_u^2+\sum_{u=1}^{d}p_u^2-2b^2_d=0\Rightarrow 
\sum_{u=1}^{d}p_u^2=b^2_d\Rightarrow\sum_{u=1}^{d-1}b_u^2=0.
$$
Therefore, $b_1=b_2=\ldots=b_{d-1}=0$ and $b_d\in\mathbb{R}$. Hence, the vector $\mathbf{b}$ lies in the one dimensional space $\{\mathbf{q}:q_1=q_2=\ldots=q_{d-1}=0, q_d\in\mathbb{R}\}$. Since $\mathbf{b}=\mathbf{T}^\top\mathbf{p}\Rightarrow\mathbf{p}=\mathbf{T}\mathbf{b}$. Since the rotation does not change the dimension of a linear space, the vector $\mathbf{p}$ also lies in one dimensional space. \qed

Given the set $\mathcal{S}$, our goal is to find all the correct reflection correspondences $(i,i^\prime)\in[n]\times[n]$ and the matrices $(\mathbf{R}_1,\mathbf{R}_2,\dots,\mathbf{R}_{d-1},\mathbf{t})$ which define the  reflection symmetry hyperplane $\boldsymbol{\pi}$. We represent all the correspondences by a permutation matrix $\mathbf{P}\in\{0,1\}^{n\times n}$, such that $\mathbf{P}_{ii^\prime}=1$ if the point $\mathbf{x}_{i^\prime}$ is the reflection point of the point $\mathbf{x}_{i}$ and $\mathbf{P}_{ii^\prime}=0$, otherwise. Here, we note from Theorem 1 that $\mathbf{P}_{ii^\prime}=1\Leftrightarrow \mathbf{P}_{i^\prime i}=1$. 

Now, we let $\mathbf{R}=(\mathbf{R}_1,\mathbf{R}_2,\dots,\mathbf{R}_{d-1})\in\mathbb{V}$. Here, $ \mathbb{V}=\mathbb{R}^{d\times d}\times\mathbb{R}^{d\times d}\times\ldots\times\mathbb{R}^{d\times d}$. Let $\mathbf{X}=\begin{bmatrix}\mathbf{x}_1&\mathbf{x}_2&\ldots&\mathbf{x}_n\end{bmatrix}\in\mathbb{R}^{d\times n}$ be the matrix containing all the points of the set $\mathcal{S}$ as its columns. Since the $i^{\text{th}}$ column of the matrix $\mathbf{X}\mathbf{P}$ is the reflection point of the point $\mathbf{x}_i$, the reflection transformation $(\mathbf{R},\mathbf{t})$ maps the matrix $\mathbf{X}$ to the reflected points matrix $\mathbf{X}\mathbf{P}$. Using Equation \ref{eq:1}, we write the reflected points in the form of the matrix $\mathbf{T}\mathbf{E}\mathbf{T}^\top(\mathbf{X}-\mathbf{t}\mathbf{e}^\top)+\mathbf{t}\mathbf{e}^\top$, where $\mathbf{e}=\begin{bmatrix} 1&1&\ldots&1\end{bmatrix}^\top$ is a vector of size $n\times1$. Therefore, Equation \eqref{eq:4} holds true  when the input set contains a perfect reflection symmetry pattern.
\begin{equation}
\mathbf{T}\mathbf{E}\mathbf{T}^\top\big(\mathbf{X}-\mathbf{t}\mathbf{e}^\top\big)+\mathbf{t}\mathbf{e}^\top=\mathbf{X}\mathbf{P}.
\label{eq:4}
\end{equation}
In practice, a reflection symmetry pattern might have been distorted. Therefore, we would be able to find only the \textit{approximate reflection symmetry}. We find the reflection transformation $(\mathbf{R},\mathbf{t})$ and the correspondences matrix $\mathbf{P}$ in such a way that the symmetry error, which we define as $\|\mathbf{T}\mathbf{E}\mathbf{T}^\top(\mathbf{X}-\mathbf{t}\mathbf{e}^\top)+\mathbf{t}\mathbf{e}^\top-\mathbf{X}\mathbf{P}\|_\text{F}^2$, is minimized. Here $\|.\|_\text{F}$ is the Frobenius norm operator. We frame this problem in an optimization framework as shown in Equation \eqref{eq:5}. 
\begin{eqnarray}
\nonumber\underset{\mathbf{R},\mathbf{t},\mathbf{P}}{\min}&\left\|(\prod\limits_{u=1}^{d-1}\mathbf{R}_u)\mathbf{E}(\prod\limits_{u=1}^{d-1}\mathbf{R}_u)^\top(\mathbf{X}-\mathbf{t}\mathbf{e}^\top)+\mathbf{t}\mathbf{e}^\top-\mathbf{X}\mathbf{P}\right\|_\text{F}^2\\
\nonumber\text{s.t.}&\mathbf{P}\mathbf{e}=\mathbf{e},\mathbf{P}^\top\mathbf{e}=\mathbf{e},\mathbf{P}\in\{0,1\}^{n\times n},\\
\nonumber&\mathbf{R}_u^\top\mathbf{R}_u=\mathbf{I}_d=\mathbf{R}_u\mathbf{R}_u^\top,\det(\mathbf{R}_u)=1,\mathbf{R}_u\in\mathbb{R}^{d\times d},\\
&\forall u\in[d-1],\mathbf{t}\in\mathbb{R}^{d}.\label{eq:5}
\end{eqnarray}
By imposing the constraints $\mathbf{P}\mathbf{e}=\mathbf{e}$ and $\mathbf{P}^\top\mathbf{e}=\mathbf{e}$, we ensure that each point has only one reflection point.
We adopt an alternating optimization approach to solve the problem defined in Equation \eqref{eq:5}. We start with initializing the reflection transformation $(\mathbf{R,t})$ and solve for the optimal correspondences $\mathbf{P}$ and then for this optimal $\mathbf{P}$, we solve for optimal the $\mathbf{(R,t)}$. We continue to alternate till convergence.

Once $\mathbf{P}$ is fixed, if we minimize the cost over the set $\mathbb{V}\times\mathbb{R}^d$, then we have to make sure that the orthogonality and the unit determinant constraints hold true for the matrices $\mathbf{R}_u, \forall u\in[d-1]$. One approach could be the Lagrange augmentation which requires us to handle $3d-3$ additional Lagrange multipliers. However, we observe that the set $\mathcal{M}=\{(\mathbf{R}_1,\ldots,\mathbf{R}_{d-1},\mathbf{t}):\mathbf{R}_u^\top\mathbf{R}_u=\mathbf{R}_u\mathbf{R}_u^\top=\mathbf{I}_d,\det(\mathbf{R}_u)=1,\mathbf{R}_u\in\mathbb{R}^{d\times d}, \forall u\in[d-1],\mathbf{t}\in\mathbb{R}^{d}\}$ of constraints is a \textit{smooth Riemannian product manifold} over which the optimization algorithms are well studied \cite{absil2009optimization}. 

We solve the sub-optimization problem for optimal $\mathbf{(R,t)}$ on a manifold which we discuss in Section \ref{subsec:Rt}. We observe that the optimization of Equation \eqref{eq:5} for $\mathbf{P}$ is a standard \textit{linear assignment problem} for which we formulate an integer linear program which we discuss in Section \ref{subsec:PI}.

\subsection{Optimizing reflection transformation ($\mathbf{R,t}$)}
\label{subsec:Rt}
In this step, we fix the correspondences matrix $\mathbf{P}$ and find the optimal reflection transformation $(\mathbf{R},\mathbf{t})$ by taking advantages from the \textit{differential structure} of the set $\mathcal{M}$. We shall now briefly introduce the differential geometry of the set  $\mathcal{M}$. 

\textbf{Differential geometry of the set $\mathcal{M}$ of constraints}.
In order to introduce the essential differential geometry of the set $\mathcal{M}$, we follow \cite{absil2009optimization}. The elements of the set $\mathcal{M}$ are of the form $(\mathbf{R},\mathbf{t})\simeq(\mathbf{R_1},\ldots,\mathbf{R_{d-1}},\mathbf{t})$. All the orthogonal matrices (each for rotation along a single axis) with determinant +1 form a Lie group, also known as \textit{special orthogonal group}, which is a smooth Riemannian manifold. The Euclidean space $\mathbb{R}^d$ is also a smooth Riemannian manifold. Therefore the set $\mathcal{M}$ is a product manifold, $\mathcal{SO}(2,d)\times\ldots\times\mathcal{SO}(2,d)\times\mathbb{R}^d$, the product of $d-1$ special orthogonal groups $\mathcal{SO}(2,d)$ and an Euclidean space $\mathbb{R}^d$. Each rotation matrix  performs rotation about a single axis. Therefore, all the possible rotation matrices about a particular axis form a $\mathcal{SO}(2)$ embedded in the Euclidean space $\mathbb{R}^{d\times d}$. We denote this group as $\mathcal{SO}(2,d)$.

The tangent space $\mathcal{T}_{(\mathbf{R},\mathbf{t})}\mathcal{M}$ at the point $(\mathbf{R,t})\in\mathcal{M}$ is 
\begin{equation}
\{(\mathbf{R\Omega},\mathbf{t}):\mathbf{\Omega}_u^\top=-\mathbf{\Omega}_u,\mathbf{\Omega}_u\in\mathbb{R}^{d\times d},\forall u\in[d-1],\mathbf{t}\in\mathbb{R}^d\}.
\label{eq:6}
\end{equation}
Here, $\mathbf{R\Omega}=(\mathbf{R}_1\mathbf{\Omega}_1,\ldots,\mathbf{R}_{d-1}\mathbf{\Omega}_{d-1})$. The Riemannian metric $\left\langle ., .\right\rangle_{\mathbf{(R,t)}}$ on the product manifold $\mathcal{M}$, which gives the intrinsic distance between two elements $(\mathbf{R\Omega},\boldsymbol{\eta}_\mathbf{t})$ and $(\mathbf{R}\mathbf{\Omega}^\prime,\boldsymbol{\eta}^\prime_\mathbf{t})$ of the tangent space at the point $\mathbf{(R,t)}$ of the manifold $\mathcal{M}$, is defined in Equation \eqref{eq:7}. 
\begin{equation}
\left\langle(\mathbf{R\Omega},\boldsymbol{\eta}_\mathbf{t}),(\mathbf{R}\mathbf{\Omega}^\prime,\boldsymbol{\eta}^\prime_\mathbf{t})\right\rangle_{\mathbf{(R,t)}}=\boldsymbol{\eta}^\top_\mathbf{t}\boldsymbol{\eta}^\prime_\mathbf{t}+\sum_{u=1}^{d-1}\text{trace}(\mathbf{\Omega}_u^\top\mathbf{\Omega}_u^\prime).
\label{eq:7}
\end{equation}
Let $\bar{f}:\mathbb{V}\times\mathbb{R}^d\rightarrow\mathbb{R}$ be a scalar function. Let the function $f=\bar{f}\mid_\mathcal{M}$ be the restriction of the function $\bar{f}$ on the product manifold $\mathcal{M}$. Since the product manifold $\mathcal{M}$ is a \textit{submanifold} of the Riemannian manifold $\mathbb{V}\times\mathbb{R}^d$,  the Riemannian gradient of the function $f$ at the point $(\mathbf{R,t})$ is obtained by projecting the Riemannian gradient of the function $\bar{f}$ at the point $(\mathbf{R,t})\in\mathbb{V}\times\mathbb{R}^d$ on the tangent space at the point $(\mathbf{R,t})\in\mathcal{M}$. Therefore, the Riemannian gradient of the function $f$  at the point $(\mathbf{R,t})$ is defined in Equation \eqref{eq:8}.
\begin{equation}
\text{grad }f(\mathbf{R},\mathbf{t})=(\mathbb{P}_{\mathbf{R}}(\nabla_\mathbf{R}\bar{f}),\mathbb{P}_{\mathbf{t}}(\nabla_\mathbf{t}\bar{f}))\in \mathcal{T}_{(\mathbf{R},\mathbf{t})}\mathcal{M}.
\label{eq:8}
\end{equation} 
Since the tangent space at a point in an Euclidean space is again an Euclidean space, the second component is given by $\mathbb{P}_\mathbf{t}(\nabla_\mathbf{t}\bar{f})=\nabla_\mathbf{t}\bar{f}$. The first component is defined as $$\mathbb{P}_{\mathbf{R}}(\nabla_\mathbf{R}\bar{f})=(\mathbb{P}_{\mathbf{R}_1}(\nabla_{\mathbf{R}_1}\bar{f}),\ldots,\mathbb{P}_{\mathbf{R}_{d-1}}(\nabla_{\mathbf{R}_{d-1}}\bar{f})).$$ Here, $$\mathbb{P}_{\mathbf{R}_j}(\nabla_{\mathbf{R}_j}\bar{f})=\mathbf{R}_j\text{skew}(\mathbf{R}_j^\top\nabla_{\mathbf{R}_j}\bar{f}),$$ where $\text{skew}(\mathbf{A})=0.5(\mathbf{A}-\mathbf{A}^\top)$. We define $\boldsymbol{\xi}_{\mathbf{R}_j}(\mathbf{R}_j)=\mathbb{P}_{\mathbf{R}_j}(\nabla_{\mathbf{R}_j}\bar{f})$. The Riemannian Hessian of the function $f$ at a point $(\mathbf{R},\mathbf{t})$ is a linear map, $\text{Hess }f:\mathcal{T}_{(\mathbf{R},\mathbf{t})}\mathcal{M}\rightarrow \mathcal{T}_{(\mathbf{R},\mathbf{t})}\mathcal{M}$ and is defined as shown in Equation \eqref{eq:9}.
\begin{equation}
\text{Hess }f(\mathbf{R},\mathbf{t})[\boldsymbol{\eta}_\mathbf{R},\boldsymbol{\eta}_\mathbf{t}]=(\mathbb{P}_{\mathbf{R}}(\text{D} \boldsymbol{\xi}_\mathbf{R}(\mathbf{R})[\boldsymbol{\eta}_\mathbf{R}]),\mathbb{P}_{\mathbf{t}}(\text{D} \boldsymbol{\xi}_\mathbf{t}(\mathbf{t})[\boldsymbol{\eta}_\mathbf{t}])).
\label{eq:9}
\end{equation}
Here, the first component $\mathbb{P}_{\mathbf{R}}(\text{D} \boldsymbol{\xi}_\mathbf{R}(\mathbf{R})[\boldsymbol{\eta}_\mathbf{R}])$ is equal to $$(\mathbb{P}_{\mathbf{R}_1}(\text{D}\boldsymbol{\xi}_{\mathbf{R}_1}(\mathbf{R}_1)[\boldsymbol{\eta}_{\mathbf{R}_1}]),\ldots, \mathbb{P}_{\mathbf{R}_{d-1}}(\text{D}\boldsymbol{\xi}_{\mathbf{R}_{d-1}}(\mathbf{R}_{d-1})[\boldsymbol{\eta}_{\mathbf{R}_{d-1}}])),$$ where $\boldsymbol{\eta}_{\mathbf{R}_j}=\mathbf{R}_j\mathbf{\Omega}_j$. The term $$\text{D}\boldsymbol{\xi}_\mathbf{x}(\mathbf{x})[\boldsymbol{\eta}_\mathbf{x}]=\lim\limits_{t\rightarrow 0}\frac{\boldsymbol{\xi}(\mathbf{x}+t\boldsymbol{\eta}_\mathbf{x})-\boldsymbol{\xi}(\mathbf{x})}{t}$$ is the classical derivative of the vector field $\boldsymbol{\xi}(\mathbf{x})$ in the direction $\boldsymbol{\eta}_\mathbf{x}$.

\textbf{The Riemannian trust region method.} Our goal is to minimize the function $f(\mathbf{R,t})$ over the product manifold $\mathcal{M}$. There exists a generalization of the popular optimization methods on the Riemannian manifolds. Since our problem is locally convex in each variable $\mathbf{R}_j$, which we prove in Theorem 6, we employ the Riemannian trust region approach \cite{absil2007trust}. It requires the Riemannian gradient and the Riemannian Hessian operator for the function $f$, which we find as follows.
Let $\bar{f}$ be a function from the set $\mathbb{V}\times\mathbb{R}^d$ to $\mathbb{R}$ and defined as $ \bar{f}(\mathbf{R},\mathbf{t})=\|\mathbf{T}\mathbf{E}\mathbf{T}^\top(\mathbf{X}-\mathbf{t}\mathbf{e}^\top)+\mathbf{t}\mathbf{e}^\top-\mathbf{X}\mathbf{P}\|_\text{F}^2.$ Its classical gradients with respect to both the variables are given in the Equations \eqref{eq:10} and \eqref{eq:11}. The detailed derivation is given in the Appendices \S A1 and \S A2.
\begin{equation}
\nabla_\mathbf{t}\bar{f}=2\big(\mathbf{I}_d-\mathbf{T}\mathbf{E}\mathbf{T}^\top\big)(2\mathbf{e}^\top\mathbf{e}\mathbf{t}-\mathbf{Xe}-\mathbf{XPe}).
\label{eq:10}
\end{equation}
\begin{equation}
\nabla_{\mathbf{R}_j}\bar{f}=-2\big(\prod_{u=1}^{j-1}\mathbf{R}_u)^\top\mathbf{A}\big(\prod_{u=1}^{d-1}\mathbf{R}_u\big)\mathbf{E}\big(\prod_{u=j+1}^{d-1}\mathbf{R}_u\big)^\top.
\label{eq:11}
\end{equation}
Here, $$\mathbf{A}=(\mathbf{X}\mathbf{P}-\mathbf{t}\mathbf{e}^\top)(\mathbf{X}-\mathbf{t}\mathbf{e}^\top)^\top+(\mathbf{X}-\mathbf{t}\mathbf{e}^\top)(\mathbf{X}\mathbf{P}-\mathbf{t}\mathbf{e}^\top)^\top$$ which satisfies $\mathbf{A}^\top=\mathbf{A}$. Now let the function $f=\bar{f}\mid_\mathcal{M}$ be the restriction of the function $\bar{f}$ on the set $\mathcal{M}$.  We obtain the Riemannian gradient of the function $f$ at a point $(\mathbf{R,t})$ by projecting the Riemannian gradient of the function $\bar{f}$ over the tangent space $\mathcal{T}_{(\mathbf{R,t})}$ at the point $(\mathbf{R,t})$. Since the manifold $\mathbb{V}\times\mathbb{R}^d$ is an Euclidean space, the Riemannian gradient of the function $\bar{f}$ is equal to its classical gradient. Therefore, we apply the definition given in Equation \eqref{eq:8} in order to find the Riemannian gradient $\text{grad} f(\mathbf{R,t})$ of the function $f$ which we denote as $(\boldsymbol{\xi}_{\mathbf{R}_1}(\mathbf{R}_1),\ldots,\boldsymbol{\xi}_{\mathbf{R}_{d-1}}(\mathbf{R}_{d-1}),\boldsymbol{\xi}_{\mathbf{t}})$ and define in Equations \eqref{eq:12} and \eqref{eq:13}. The detailed derivation is given in the Appendices \S A3 and \S A4.
\begin{equation}
\boldsymbol{\xi}_\mathbf{t}(\mathbf{t})=2\big(\mathbf{I}_d-\mathbf{T}\mathbf{E}\mathbf{T}^\top\big)(2\mathbf{e}^\top\mathbf{e}\mathbf{t}-\mathbf{Xe}-\mathbf{XPe}),
\label{eq:12}
\end{equation}
\begin{equation*}
\boldsymbol{\xi}_{\mathbf{R}_j}(\mathbf{R}_j)=-\mathbf{R}_j\big(\prod\limits_{u=1}^{j}\mathbf{R}_u\big)^\top\mathbf{A}\big(\prod\limits_{u=1}^{d-1}\mathbf{R}_u\big)\mathbf{E}\big(\prod\limits_{u=j+1}^{d-1}\mathbf{R}_u\big)^\top
\end{equation*}
\begin{equation}
+\mathbf{R}_j\big(\prod\limits_{u=j+1}^{d-1}\mathbf{R}_u\big)\mathbf{E}\big(\prod\limits_{u=1}^{d-1}\mathbf{R}_u\big)^\top\mathbf{A}^\top\big(\prod\limits_{u=1}^{j}\mathbf{R}_u\big).
\label{eq:13}
\end{equation}

We determine the Riemannian Hessian of the function $f$ using the definition given in Equation \eqref{eq:9}. In order to determine the $j^\text{th}$ component $\text{Hess}_{\mathbf{R}_j}(f(\mathbf{R,t}))[\mathbf{R}_j\Omega_j]$ of the Riemannian Hessian, which is equal to $\mathbb{P}_{\mathbf{R}_j}(\text{D} \boldsymbol{\xi}_{\mathbf{R}_j}(\mathbf{R}_j)[\mathbf{R}_j\mathbf{\Omega}_j])$, we first find the classical derivative $\text{D} \boldsymbol{\xi}_{\mathbf{R}_j}(\mathbf{R}_j)[\mathbf{R}_j\mathbf{\Omega}_j]$ of the Riemannian gradient $\boldsymbol{\xi}_{\mathbf{R}_j}(\mathbf{R}_j)$ in the direction $\mathbf{R}_j\mathbf{\Omega}_j$ and then apply the projection operator $\mathbb{P}_{\mathbf{R}_j}$. Therefore, the $j^\text{th}$ component $\text{Hess}_{\mathbf{R}_j}(f(\mathbf{R,t}))[\mathbf{R}_j\Omega_j]$ of the Riemannian Hessian is equal to
\begin{equation}
\frac{1}{2}\mathbf{R}_j([\mathbf{B}_1,[\mathbf{R}_j^\top\mathbf{B}_2\mathbf{R}_j, \mathbf{\Omega}_j]]+[[\mathbf{\Omega}_j,\mathbf{B}_1],\mathbf{R}_j^\top\mathbf{B}_2\mathbf{R}_j]).
\label{eq:14}
\end{equation}

The detailed derivation is given in the Appendix \S A5. Here $[.,.]$ is the Lie bracket and defined as $[\mathbf{U},\mathbf{V}]=\mathbf{UV}-\mathbf{VU}$ for any two matrices $\mathbf{U}$ and $\mathbf{V}$, $$\mathbf{B}_1=\big(\prod_{u=j+1}^{d-1}\mathbf{R}_u)\mathbf{E}\big(\prod_{u=j+1}^{d-1}\mathbf{R}_u)^\top,$$ and 
$$\mathbf{B}_2=\big(\prod_{u=1}^{j-1}\mathbf{R}_u\big)^\top\mathbf{A}\big(\prod_{u=1}^{j-1}\mathbf{R}_u\big).$$ In a similar way, we determine the component, $\mathbb{P}_{\mathbf{t}}(\text{D} \boldsymbol{\xi}_\mathbf{t}(\mathbf{t})[\boldsymbol{\eta}_\mathbf{t}])$, of the Riemannian Hessian which is shown in Equation \eqref{eq:15}.
\begin{equation}
\text{Hess}_{\mathbf{t}}(f(\mathbf{R,t}))[\boldsymbol{\eta}_\mathbf{t}]=4n\big(\mathbf{I}_d-\mathbf{T}\mathbf{E}\mathbf{T}^\top\big)\boldsymbol{\eta}_\mathbf{t}.
\label{eq:15}
\end{equation}

The detailed derivation is given in the Appendix \S A6. Now, we apply the Riemannian-trust-region method using the Riemannian gradient and Hessian defined in Equations \eqref{eq:12}, \eqref{eq:13}, \eqref{eq:14}, and \eqref{eq:15} in order to obtain the optimal solution.  We use the \textit{manopt} toolbox in order to implement the optimization problem given in Equation \eqref{eq:5} for a fixed $\mathbf{P}$ \cite{manopt}.

\textbf{Determining the reflection symmetry hyperplane $\boldsymbol{\pi}$}. In order to determine the reflection hyperplane $\boldsymbol{\pi}$, we use Theorem 2 which states that the normal vector of $\boldsymbol{\pi}$ lies in the null space of the matrix $\mathbf{I}_d+\big(\prod_{u=1}^{d-1}\mathbf{R}_u\big)\mathbf{E}\big(\prod_{u=1}^{d-1}\mathbf{R}_u\big)^\top$ and the optimal translation $\mathbf{t}$ lies on the hyperplane.

\subsection{Optimizing Correspondences $\mathbf{P}$}
\label{subsec:PI}
After obtaining the current estimate of the reflection transformation $(\mathbf{R,t})$, we improve the correspondences matrix $\mathbf{P}$ by solving the problem given in Equation \eqref{eq:5} while fixing $ (\mathbf{R,t})$. We show that this sub-problem is equivalent to a linear assignment problem, where an \textit{assignment} is a pair $(i,i^\prime)$ of reflection symmetry points.

\textbf{Claim 1:} \textit{The optimization problem given in Equation \eqref{eq:5} is a linear assignment problem in $\mathbf{P}$, for a fixed  $(\mathbf{R,t})$.}

\textbf{Proof:} Let us consider the cost function in Equation \eqref{eq:5} and let $\mathbf{X}_\text{m}=\mathbf{T}\mathbf{E}\mathbf{T}^\top(\mathbf{X}-\mathbf{t}\mathbf{e}^\top)+\mathbf{t}\mathbf{e}^\top$. We have 
$$
\|\mathbf{X}_\text{m}-\mathbf{X}\mathbf{P}\|_\text{F}^2=\text{trace}((\mathbf{X}_\text{m}-\mathbf{X}\mathbf{P})^\top(\mathbf{X}_\text{m}-\mathbf{X}\mathbf{P}))$$$$
=\text{trace}(\mathbf{X}_\text{m}^\top\mathbf{X}_\text{m}-2\mathbf{X}_\text{m}^\top\mathbf{X}\mathbf{P}+\mathbf{X}^\top\mathbf{X}\mathbf{P}\mathbf{P}^\top).$$
Since, the first and the third terms (using the fact that the permutation matrices are orthogonal) are not the functions of $\mathbf{P}$,  the problem of finding the point of minimum of the function $\|\mathbf{X}_\text{m}-\mathbf{X}\mathbf{P}\|_\text{F}^2$ is identical to the problem of finding the point of maximum of the function $\text{trace}(\mathbf{X}_\text{m}^\top\mathbf{X}\mathbf{P})$. Using the identity $\text{trace}(\mathbf{A}^\top\mathbf{B})=\mathtt{vec}(\mathbf{A})^\top\mathtt{vec}(\mathbf{B})$,  we have that $\text{trace}(\mathbf{X}_\text{m}^\top\mathbf{X}\mathbf{P})=\mathtt{vec}(\mathbf{X}^\top\mathbf{X}_\text{m})^\top\mathtt{vec}(\mathbf{P})$, where the operator $\mathtt{vec}$ vectorizes a matrix by stacking all the columns successively in a column vector.

Therefore, for a fixed reflection transformation, the problem defined in Equation \eqref{eq:5} is equivalent to the problem defined in Equation \eqref{eq:16}.
\begin{eqnarray}
\nonumber\underset{\mathbf{P}\in\{0,1\}^{n\times n}}{\max}&\text{trace}(\mathbf{X}_\text{m}^\top\mathbf{X}\mathbf{P})=\mathtt{vec}(\mathbf{X}_\text{m}^\top\mathbf{X})^\top\mathtt{vec}(\mathbf{P})\\
\text{subject to}&\mathbf{P}\mathbf{e}\leq\mathbf{e},\;\;\mathbf{P}^\top\mathbf{e}\leq\mathbf{e},
\label{eq:16}
\end{eqnarray}
which is a standard linear assignment problem.$\square$

\textbf{Claim 2: }\textit{The problem defined in Equation \eqref{eq:16} is an integer linear program.}

\textbf{Proof:} Let $\mathbf{v}_1$ be a vector of size $n^2\times 1$ with the first $n$ coordinates equal to one and the last $n(n-1)$ coordinates equal to zero. Let $\mathbf{e}_1$  be a vector of size $n\times 1$ with all the coordinates equal to zero except the first coordinate which is equal to one. Let $\mathbf{v}_2=\begin{bmatrix} \mathbf{e}_1^\top&\mathbf{e}_1^\top&\ldots&\mathbf{e}_1^\top\end{bmatrix}^\top
$ be a vector of size $n^2\times 1$. Now let us construct the matrices $\mathbf{A}_1$ and $\mathbf{A}_2$, each of size $n\times n^2$, such that the $i^\text{th}$ row of the matrix $\mathbf{A}_1$ is equal to the row vector $\mathtt{cs}(\mathbf{v}_1^\top,n(i-1))$ and the $i^\text{th}$ row of the matrix $\mathbf{A}_2$ is equal to the row vector $\mathtt{cs}(\mathbf{v}_2^\top,i-1)$. Here $\mathtt{cs}(\mathbf{v}^\top,i)$ is a row vector obtained by circularly shifting any row vector $\mathbf{v}^\top$ right by $i$ coordinates.

Now, it is trivial to verify that the constraint $\mathbf{P}^\top\mathbf{e}\leq\mathbf{e}$ is equivalent to $\mathbf{A}_1\mathtt{vec}(\mathbf{P})\leq \mathbf{e} $ and the constraint $\mathbf{P}\mathbf{e}\leq\mathbf{e}$ is equivalent to $\mathbf{A}_2\mathtt{vec}(\mathbf{P})\leq \mathbf{e}$. Therefore, the problem defined in Equation \eqref{eq:16} is equivalent to the problem defined in Equation \eqref{eq:17}.
\begin{eqnarray}
\nonumber\underset{\mathbf{a}\in\{0,1\}^{n^2\times 1}}{\max}&\mathtt{vec}(\mathbf{X}_\text{m}^\top\mathbf{X})^\top\mathbf{a}\\
\text{subject to}&\begin{bmatrix} \mathbf{A}_1^\top&\mathbf{A}_2^\top\end{bmatrix}^\top\mathbf{a}\leq\begin{bmatrix} \mathbf{e}^\top&\mathbf{e}^\top\end{bmatrix}^\top
\label{eq:17}
\end{eqnarray}
which is an integer linear program with $\mathbf{a}=\mathtt{vec}(\mathbf{P}).\square$

\textbf{Solving the ILP.} Since  ILP is an NP-complete problem, there may not exist a polynomial time algorithm to find the optimal solution. We relax this ILP to a linear program by converting the constraint $\mathbf{a}\in\{0,1\}^{n^2\times 1}$ into $\mathbf{a}\in[0,1]^{n^2\times 1}$. Now, the above ILP becomes a linear program. We first solve this LP using the Karmarkar's algorithm in \cite{karmarkar1984new} which takes $O(n^{3.5})$ time. The solution $\mathbf{a}^\star=\begin{bmatrix}a^\star_1&a^\star_2&\ldots&a^\star_{n^2}\end{bmatrix}^\top$ of this LP belongs to $[0,1]^{n^2\times 1}$ which is a continuous solution. However, our final solution $\mathbf{a}^{f}=\begin{bmatrix}a^f_1&a^f_2&\ldots&a^f_{n^2}\end{bmatrix}^\top$ of the proposed ILP should be a discrete solution. We follow the rounding approach, as explained in  (\cite{kleinberg2006algorithm}, ch. 11). The $i$-th element $a^f_i$ of the final solution is equal to $1$, if $a^\star_i\geq0.5$ and equal to 0, if $a^\star_i<0.5$. This solution $\mathbf{a}^f$ may not be the optimal solution because according to \cite{kleinberg2006algorithm}, $\mathtt{vec}(\mathbf{X}_\text{m}^\top\mathbf{X})^\top\mathbf{a}^f\geq\frac{1}{2}\times \mathtt{vec}(\mathbf{X}_\text{m}^\top\mathbf{X})^\top\mathbf{a}^{OPT}$. Here, $\mathbf{a}^{OPT}$ is the optimal solution of the above ILP.
\subsection{Convergence Analysis}
\label{subsec:CA}
We derive the essential results in order to prove that the alternating optimization framework converges.

\textbf{Theorem 3: }
\textit{The cost function $f(\mathbf{R,t,P})$ is convex in the variable $\mathbf{t}$.
}

\textbf{Proof:} In order to prove this, we prove that the Riemannian Hessian of the function $f$ with respect to the variable $\mathbf{t}$ is a positive semi-definite (PSD) matrix. Let $\boldsymbol{\eta}_\mathbf{t}=\begin{bmatrix}\eta_1&\eta_2&\ldots&\eta_d\end{bmatrix}^\top\in\mathbb{R}^d$. Then using the definition of Riemannian metric, we have $$\nonumber\left\langle\boldsymbol{\eta}_\mathbf{t},\text{Hess}_{\mathbf{t}}(f)[\boldsymbol{\eta}_\mathbf{t}]\right\rangle_\mathbf{t}=\boldsymbol{\eta}_\mathbf{t}^\top\text{Hess}_{\mathbf{t}}(f)[\boldsymbol{\eta}_\mathbf{t}].$$
Now, using the Riemannian Hessian $\text{Hess}_{\mathbf{t}}(f)[\boldsymbol{\eta}_\mathbf{t}]$ defined in Equation \eqref{eq:15}, we have that
$$\boldsymbol{\eta}_\mathbf{t}^\top\text{Hess}_{\mathbf{t}}(f)[\boldsymbol{\eta}_\mathbf{t}]=\boldsymbol{\eta}_\mathbf{t}^\top\boldsymbol{\eta}_\mathbf{t}-\big(\mathbf{T}^\top\boldsymbol{\eta}_\mathbf{t}\big)^\top\mathbf{E}\big(\mathbf{T}^\top\boldsymbol{\eta}_\mathbf{t}\big)$$
Now let  $\mathbf{q}=\mathbf{T}^\top\mathbf{\boldsymbol{\eta}_\mathbf{t}}$. Then, we obtain
$$\boldsymbol{\eta}_\mathbf{t}^\top\text{Hess}_{\mathbf{t}}(f)[\boldsymbol{\eta}_\mathbf{t}]=\boldsymbol{\eta}_\mathbf{t}^\top\boldsymbol{\eta}_\mathbf{t}-\mathbf{q}^\top\mathbf{E}\mathbf{q}$$
$$
\nonumber=\|\boldsymbol{\eta}_\mathbf{t}\|_2^2-\sum_{u=1}^{d-1}q_u+q^2_d=\|\boldsymbol{\eta}_\mathbf{t}\|_2^2-\|\mathbf{q}\|_2^2+2q_d^2.
$$
 Now, we know that $\mathbf{T}\mathbf{T}^\top=\mathbf{I}$. Hence, we have $$\|\mathbf{q}\|_2^2=\mathbf{q}^\top\mathbf{q}=\boldsymbol{\eta}_\mathbf{t}^\top\mathbf{T}\mathbf{T}^\top\boldsymbol{\eta}_\mathbf{t}=\boldsymbol{\eta}_\mathbf{t}^\top\boldsymbol{\eta}_\mathbf{t}=\|\boldsymbol{\eta}_\mathbf{t}\|_2^2.$$ Therefore,
$$\|\boldsymbol{\eta}_\mathbf{t}\|_2^2-\|\mathbf{q}\|_2^2=0
\Rightarrow\boldsymbol{\eta}_\mathbf{t}^\top\text{Hess}_{\mathbf{t}}(f)[\boldsymbol{\eta}_\mathbf{t}]=2q_d^2\geq 0.\square
$$

\textbf{Theorem 4: }
\textit{At the critical point, the matrix $\mathbf{T}^\star=\prod_{u=1}^{d}\mathbf{R}_u^\star$ contains the eigenvectors of the matrix $\mathbf{A}$ as columns.}

\textbf{Proof:}
At the critical point, the Riemannian gradient given in Equation \eqref{eq:13} vanishes. Therefore, $\boldsymbol{\xi}_{\mathbf{R}_j}(\mathbf{R}_j)=\mathbf{0}_{d\times d}$. Now pre-multiplying it with $\big(\prod_{u=1}^j\mathbf{R}_u\big)\mathbf{R}_j^\top$ and then post-multiplying with $\big(\prod_{u=j+1}^{d-1}\mathbf{R}_u\big)$, we achieve  
$$
\mathbf{A}\mathbf{T}^\star\mathbf{E}=\mathbf{T}^\star\mathbf{E}(\mathbf{T}^\star)^\top\mathbf{A}\mathbf{T}^\star\Rightarrow(\mathbf{T}^\star)^\top\mathbf{A}\mathbf{T}^\star\mathbf{E}=\mathbf{E}(\mathbf{T}^\star)^\top\mathbf{A}\mathbf{T}^\star.
$$
Now, let $\mathbf{Q}=\begin{bmatrix}
\mathbf{Q}_1&\mathbf{q}_2\\\mathbf{q}^\top_3&q_4
\end{bmatrix}=(\mathbf{T}^\star)^\top\mathbf{A}\mathbf{T}^\star$ be a matrix. Then, we have $\mathbf{QE}=\mathbf{EQ}$. Therefore,
$$
\begin{bmatrix}
\mathbf{Q}_1&\mathbf{q}_2\\\mathbf{q}^\top_3&q_4
\end{bmatrix}\begin{bmatrix}
\mathbf{I}_{d-1}&\mathbf{0}_{d-1}\\\mathbf{0}^\top_{d-1}&-1
\end{bmatrix}=\begin{bmatrix}
\mathbf{I}_{d-1}&\mathbf{0}_{d-1}\\\mathbf{0}^\top_{d-1}&-1
\end{bmatrix}\begin{bmatrix}
\mathbf{Q}_1&\mathbf{q}_2\\\mathbf{q}^\top_3&q_4
\end{bmatrix}
$$
$$
\Rightarrow \mathbf{q}_2=\mathbf{0}_{d-1},\mathbf{q}_3=\mathbf{0}_{d-1},\mathbf{Q}_1\mathbf{I}_{d-1}=\mathbf{I}_{d-1}\mathbf{Q}_1.
$$
Since, $\mathbf{I}_{d-1}$ is a diagonal matrix and the equality $\mathbf{Q}_1\mathbf{I}_{d-1}=\mathbf{I}_{d-1}\mathbf{Q}_1$ holds true, it is easy to prove that $\mathbf{Q}_1$ is a diagonal matrix. 
Therefore, the matrix $\mathbf{Q}$ is also diagonal. The \textit{spectral theorem} states that every real symmetric matrix has eigenvalue decomposition with real eigenvalues and orthogonal eigenvectors. Here, we have observed that the matrix $\mathbf{A}$ is a real symmetric matrix and satisfies $\mathbf{Q}=(\mathbf{T}^\star)^\top\mathbf{A}\mathbf{T}^\star$, where the matrix $\mathbf{Q}$ is a diagonal matrix and the matrix $\mathbf{T}^\star$ is an orthogonal matrix. Therefore, the matrix $\mathbf{T}^\star$ is the matrix containing the eigenvectors of the matrix $\mathbf{A}$. In Theorem 5, we prove that the order of stacking eigenvectors of $\mathbf{A}$ as columns of $\mathbf{T}^\star$ affects the convexity of the problem. $\square$

\textbf{Theorem 5:}
\textit{The cost function $f(\mathbf{R,t,P})$ is locally convex in each rotation matrix $\mathbf{R}_j$.
}

\begin{figure}[htbp]
\centering
\includegraphics[width=0.95\linewidth]{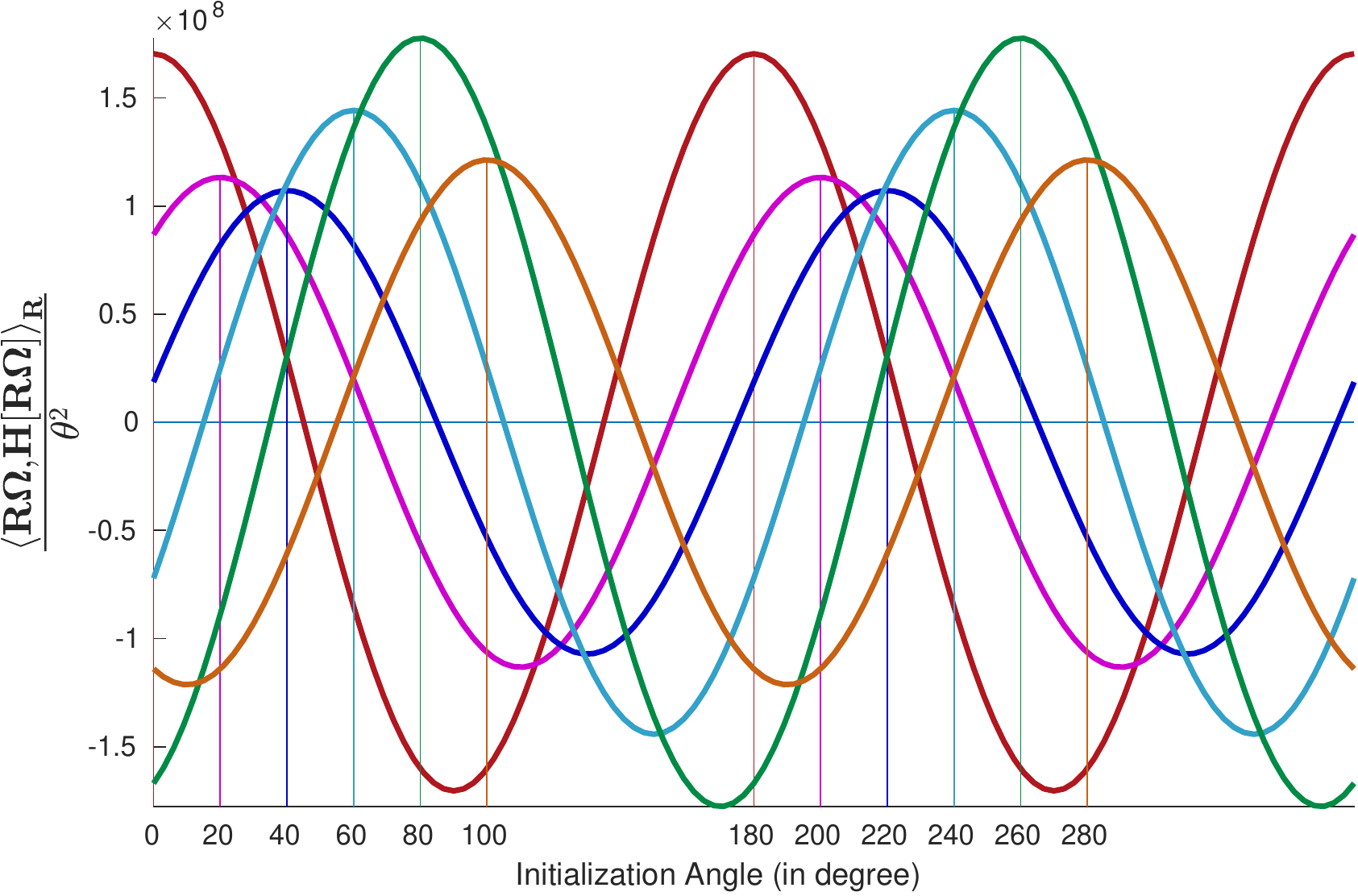}
\caption{Illustration of the local convexity. The value $\frac{\left\langle \mathbf{R\Omega}, \mathbf{H}[\mathbf{R\Omega}]\right\rangle_{\mathbf{R}}}{\theta^2}$ against the initialization angle $\theta$ for 6 reflection symmetry patterns having different orientations, $\{0^\circ,20^\circ, 40^\circ,60^\circ,80^\circ,100^\circ\}$ for symmetry axis. The PSD value (divided by $\theta^2$) is positive in the proximity of the optimal angle.}
\label{fig:lc}
\end{figure}

\textbf{Proof:}
In order to show the local convexity in  $\mathbf{R}_j$, we have to show that the value $\left\langle \mathbf{R}_j\mathbf{\Omega}_j, \mathbf{H}[\mathbf{R}_j\mathbf{\Omega}_j]\right\rangle_{\mathbf{R}_j}\geq0$ in the neighborhood of the optimal angle $\theta_j^\star$. Here, $\mathbf{H}[\mathbf{R}_j\mathbf{\Omega}_j]=\text{Hess}_{\mathbf{R}_j}(f(\mathbf{R,t}))[\mathbf{R}_j\mathbf{\Omega}_j]$. By using the Riemannian metric defined in Equation \eqref{eq:7}, we have 
$$\left\langle \mathbf{R}_j\mathbf{\Omega}_j, \mathbf{H}[\mathbf{R}_j\mathbf{\Omega}_j]\right\rangle_{\mathbf{R}_j}=\text{trace}(\mathbf{\Omega}_j^\top\mathbf{R}_j^\top\mathbf{H}[\mathbf{R}_j\mathbf{\Omega}_j]).$$ 
By using Equation \eqref{eq:14}, the matrix  $\mathbf{R}_j^\top\mathbf{H}[\mathbf{R}_j\mathbf{\Omega}_j]$ is equal to
$$
0.5[\mathbf{B}_1,[\mathbf{R}_j^\top\mathbf{B}_2\mathbf{R}_j, \mathbf{\Omega}_j]]+0.5[[\mathbf{\Omega}_j,\mathbf{B}_1],\mathbf{R}_j^\top\mathbf{B}_2\mathbf{R}_j].
$$
In the Appendix \S A7, we show that the $\text{trace}(\mathbf{\Omega}_j^\top\mathbf{R}_j^\top\mathbf{H}[\mathbf{R}_j\mathbf{\Omega}_j])$ is equal to
\begin{equation}4\times\text{trace}(\mathbf{R}_j^\top\mathbf{B}_2\mathbf{R}_j(\mathbf{\Omega}_j\mathbf{B}_1\mathbf{\Omega}_j-\mathbf{\Omega}_j\mathbf{\Omega}_j\mathbf{B}_1)).\label{eq:18}\end{equation}
We visualize this term for $d=2$. For $d=2$, the matrix $\mathbf{\Omega}=\begin{bmatrix}0&-\theta\\\theta &0\end{bmatrix}$,  $\mathbf{E}=\begin{bmatrix}1&0\\0 &-1\end{bmatrix}$,  and let $\mathbf{A}=\begin{bmatrix}a_1&a_2\\a_2 &a_3\end{bmatrix}$ and $\mathbf{R}=\begin{bmatrix}\cos\theta&-\sin\theta\\\sin\theta &\cos\theta\end{bmatrix}$.
We have that $$\left\langle \mathbf{R\Omega}, \mathbf{H}[\mathbf{R\Omega}]\right\rangle_{\mathbf{R}}=8a_2\theta^2\sin(2\theta)+4\theta^2\cos(2\theta)(a_1-a_3).$$ In Fig. \ref{fig:lc}, we plot the value $\frac{\left\langle \mathbf{R\Omega}, \mathbf{H}[\mathbf{R\Omega}]\right\rangle_{\mathbf{R}}}{\theta^2}$ against the initialization angle $\theta$ for six reflection symmetry patterns having different orientations for symmetry axis. We observe that the PSD values are positive in the proximity of the optimal angles. Therefore, it is locally convex. We further observe that this quantity is maximum at the optimal angle. We also observe that, if $\theta $ is the symmetry axis orientation, then the PSD value becomes positive in the proximity of $\theta$ and  $\theta+180^\circ$. The reason for the second range is that, if $\theta$ is the slope of a line, then $\theta+180^\circ$ is also the slope of the same line.

In Theorem 4, we claimed that the order in which the eigenvectors are stacked  as columns of the matrix $\mathbf{R}$ affects the local convexity. We prove it as follows. At the critical point, we have that $\mathbf{R}^\top\mathbf{A}\mathbf{R}=\text{diag}(d_1,d_2)$. We note that $\mathbf{\Omega}_j\mathbf{B}_1\mathbf{\Omega}_j-\mathbf{\Omega}_j\mathbf{\Omega}_j\mathbf{B}_1=\mathbf{E}$ for $d=2$. Now from Equation \eqref{eq:18}, we achieve $$\left\langle \mathbf{R\Omega}, \mathbf{H}[\mathbf{R\Omega}]\right\rangle_{\mathbf{R}}=d_1-d_2\Rightarrow d_1\geq d_2.$$ Therefore, the first column of the matrix $\mathbf{R}^\star$ should be the eigenvector corresponding to the maximum eigenvalue and the second column of the matrix $\mathbf{R}^\star$ should be the eigenvector corresponding to the minimum eigenvalue of the matrix $\mathbf{A}$.$\square$

\textbf{Theorem 6: }
\textit{The proposed alternating framework converges to the global minimum if the initialization of the rotation matrices $\mathbf{R}_1$, $\ldots$, $\mathbf{R}_{d-1}$ are within the proximity of the optimal rotation matrices and initialization of the translation $\mathbf{t}$ is any random vector.}

\begin{figure}[htbp]
\centering
\includegraphics[width=0.93\linewidth]{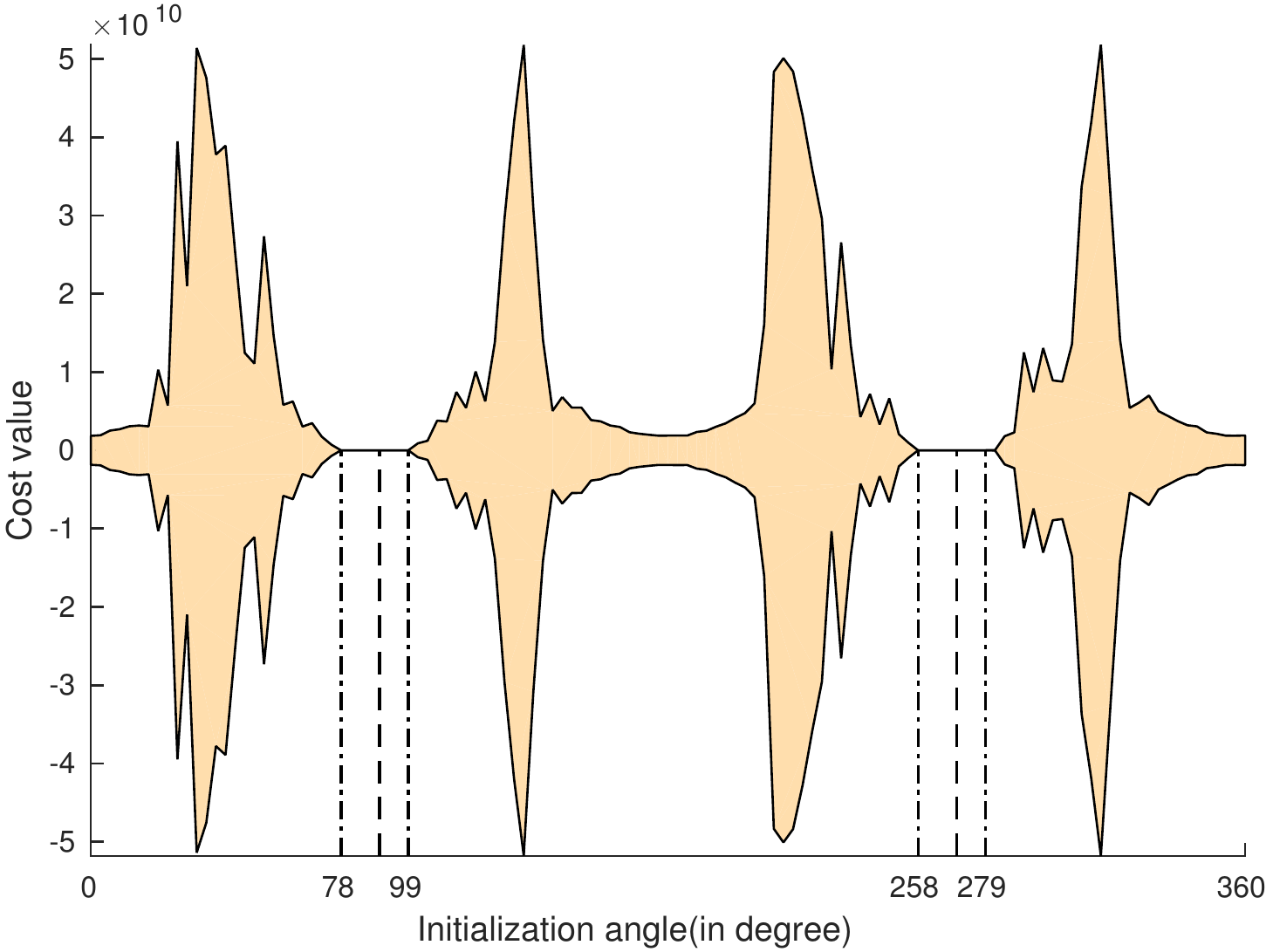}
\caption{ We plot the error at the convergence point against the initialization angles for the case $d=2$ (we shift the error vectors for different optimal angles so that the optimal angle is always $90^\circ$). We observe that the variance in the error becomes zeros for initialization angle $\theta_0\in(90^\circ-12^\circ,90^\circ+9^\circ)$ and $\theta_0\in(270^\circ-12^\circ,270^\circ+9^\circ)$.}
\label{fig:exp_ang}
\end{figure}

\textbf{Proof:}
We observe that the proposed alternation framework is basically the block coordinate descent (BCD) method, where $(\mathbf{R}_1,\ldots,\mathbf{R}_{d-1},\mathbf{t})$ and $\mathbf{P}$ are two blocks of coordinates. According to \cite{tseng2001convergence}, the BCD method converges if the cost function is convex in each block of coordinates. Here, we have seen that the cost function is convex in the coordinates $\mathbf{t}$ (Theorem 3), convex in the coordinates $\mathbf{P}$ on the relaxed domain $[0,1]^{n\times n}$, and locally convex in the coordinates $(\mathbf{R}_1,\ldots,\mathbf{R}_{d-1})$ (Theorem  5). This implies that if the initialization of $(\mathbf{R}_1,\ldots,\mathbf{R}_{d-1})$ is within the proximity of the optimal solution, then the alternating framework converges to the global minimum. We experimentally show this theorem for the case $d=2$. We use the dataset for $d=2$ with $\sigma=0$ as mentioned in \S \ref{subsec:rps}. In Fig. \ref{fig:exp_ang}, we plot the error (averaged over all optimal angles) at the convergence point against the initialization angles for the case $d=2$ (we shift the error vectors for different optimal angles so that the optimal angle is always $90^\circ$). We observe that the variance becomes zero for initialization angle $\theta_0\in(90^\circ-12^\circ,90^\circ+9^\circ)$ and $\theta_0\in(270^\circ-12^\circ,270^\circ+9^\circ)$. The reason for the second range is that, if $\theta$ is the slope of a line, then $\theta+180^\circ$ is also the slope of the same line.$\square$

In summary, in order to obtain the optimal $\mathbf{(R^\star,t^\star,\mathbf{P^\star})}$, we follow Algorithm \ref{alg:1}.

\begin{algorithm}[htbp]
\caption{}\label{alg:1}
\begin{algorithmic}[1]
\State Input:  Set of points $\mathcal{S}= \{\mathbf{x}_i \}_{i=1}^n$.
\State Initialize angles $\boldsymbol{\theta}_0$ and translation $\mathbf{t}$.
\State Solve the ILP defined in Equation \eqref{eq:17} for $\mathbf{P}$.
\State For this $\mathbf{P}$, solve for $\mathbf{(R,t)}$ using the Riemannian-trust-region method using the Riemannian gradient and Hessian defined in Equations \eqref{eq:12}, \eqref{eq:13}, \eqref{eq:14}, and \eqref{eq:15}.
\State Keep iterating steps 3 and 4 till convergence.
\State Output: The optimal $\mathbf{R}_1^\star,\mathbf{R}_2^\star,\ldots,\mathbf{R}_{d-1}^\star$ and $\mathbf{t}^\star$.
\end{algorithmic}
\end{algorithm}
\textbf{Initialization Strategy:} In the Theorem 5, we have shown that $f(\mathbf{R,t,P})$ is locally convex in rotation matrix $\mathbf{R}$. Therefore, Algorithm 1 converges to the global minimum if we initialize the rotation matrix in the proximity of the global solution. Hence, we approximate the initial $\mathbf{R}$ by finding a small set of candidate pairs of mirror symmetric points. We discuss the proposed approach for finding a small set of candidate pairs of mirror symmetric points as follows.

Let us consider the input set $\mathcal{S}=\{\mathbf{x}_i\}_{i=1}^n$. We propose a randomized approach to find a small set of candidate pairs of mirror symmetric points. We select two points, $\mathbf{x}_p$ and $\mathbf{x}_q$, uniformly at random from the set $\mathcal{S}$. Let $\mathbf{x}_{p^\prime}$ and $\mathbf{x}_{q^\prime}$ be their actual mirror images, respectively. We then construct two sets, $
\mathcal{P}=\{(\mathbf{x}_p,\mathbf{x}_i)\}_{i=1,i\neq p,q}^{n}$ and $\mathcal{Q}=\{(\mathbf{x}_q,\mathbf{x}_i)\}_{i=1,i\neq q,p}^{n}
$ of pairs of points. Given the sets $\mathcal{P}$ and $\mathcal{Q}$, our goal is to find the pairs $(\mathbf{x}_p,\mathbf{x}_{p^\prime})$ and $(\mathbf{x}_q,\mathbf{x}_{q^\prime})$. It is trivial to observe that $(\mathbf{x}_p,\mathbf{x}_{p^\prime})\in\mathcal{P}$ and $(\mathbf{x}_q,\mathbf{x}_{q^\prime})\in\mathcal{Q}$. We note that each pair of points define its own symmetry plane, the one which is perpendicular to the line segment joining the two points and passing through the mid-point of this line segment. Now, if the pairs $(\mathbf{x}_p,\mathbf{x}_{p^\prime})$ and $(\mathbf{x}_q,\mathbf{x}_{q^\prime})$ are true pairs then both the reflection planes, defined by these two pairs, should be the same. For each pair $(\mathbf{x}_p,\mathbf{x}_{i})\in\mathcal{P}$, we keep sampling a pair $(\mathbf{x}_q,\mathbf{x}_{j})\in\mathcal{Q}$ uniformly at random without replacement until the reflection planes defined by these two pairs are the same. We determine whether the two reflection planes, defined by these two pairs, $\boldsymbol{\pi}_{pi}:\boldsymbol{\eta}_{pi}^\top\mathbf{x}-c_{pi}=0$ and $\boldsymbol{\pi}_{qj}:\boldsymbol{\eta}_{qj}^\top\mathbf{x}-c_{qj}=0$ are the same if the conditions, $
\cos^{-1}(\boldsymbol{\eta}_{pi}^\top\boldsymbol{\eta}_{qj})\leq\epsilon_\theta$ and $ \frac{\min\{d_q,d_j\}}{\max\{d_q,d_j\}}\geq1-\epsilon_d
$ are true.  Here, $\boldsymbol{\eta}_{pi}=\frac{\mathbf{x}_p-\mathbf{x}_i}{\|\mathbf{x}_p-\mathbf{x}_i\|_2}$ is the normal vector to the plane $\boldsymbol{\pi}_{pi}$, $c_{pi}=\boldsymbol{\eta}_{pi}^\top(\frac{\mathbf{x}_p+\mathbf{x}_i}{2})$ is the  distance of the origin from the plane $\boldsymbol{\pi}_{pi}$, $\boldsymbol{\eta}_{qj}=\frac{\mathbf{x}_q-\mathbf{x}_j}{\|\mathbf{x}_q-\mathbf{x}_j\|_2}$ is the normal vector to the plane $\boldsymbol{\pi}_{qj}$, $c_{qj}=\boldsymbol{\eta}_{qj}^\top(\frac{\mathbf{x}_q+\mathbf{x}_j}{2})$ is the distance of the origin from the plane $\boldsymbol{\pi}_{qj}$, $d_q=|\boldsymbol{\eta}_{pi}^\top\mathbf{x}_{q}-c_{pi}|$, and $d_j=|\boldsymbol{\eta}_{pi}^\top\mathbf{x}_{j}-c_{pi}|$.

We repeat the above procedure ten times. With this, we get a set of 20 (2 for each run) candidate pairs of mirror symmetric points. Since we consider the case where only a single symmetric object is present in the input set, we consider the median plane of the 20 planes defined by the above computed 20 candidate pairs. Now, we use the normal $\boldsymbol{\eta}$ to this median plane for initialization. 
We also initialize the initial translation vector $\mathbf{t}$ as the median of the mid-points of the line segment joining the points of the candidate pairs of the mirror symmetric points.

First, we subtract each data point of the point cloud from the estimated $\mathbf{t}$ of the point cloud. This ensures that the reflection symmetry plane passes through the origin. Now, we know the unit normal to the reflection symmetry plane. Therefore, we use the Householder transform to reflect each point which is $\mathbf{x}_{i^\prime}=(\mathbf{I}-2\boldsymbol{\eta}\boldsymbol{\eta}^\top)\mathbf{x}_i$. Therefore, we have the matrix $\mathbf{X}$ containing the original point cloud and the matrix $\mathbf{X}_\text{m}$ containing the reflected point cloud about the estimated reflection symmetry plane. Now, using $\mathbf{X}$ and $\mathbf{X}_\text{m}$, we solve the linear assignment problem, defined in Equation \eqref{eq:17} to find the matrix $\mathbf{P}$. Now, we use these approximate correspondences to estimate the reflection symmetry plane as step 4 of Algorithm 1.

\section{Time Complexity}
\label{sec:cc}
There are two main steps involved in our algorithm. The first one is to solve for reflection symmetry transformation matrices $\mathbf{R}_1,\mathbf{R}_2,\ldots,\mathbf{R}_{d-1},\mathbf{t}$  using the Riemannian trust region \cite{absil2007trust}. The second step is to find the pairs of reflective symmetric points using an integer linear program. The time complexity of Riemannian trust region method is $O(nd^{2})$. Since solving integer linear program is an $NP$-complete problem, we first relax it to a linear program (as discussed at the end of \S \ref{subsec:PI}).  The time complexity of solving a linear program is polynomial in the number of points in the point cloud. We use the Karmarkar's algorithm in \cite{karmarkar1984new} which has the time complexity of $O(n^{3.5})$. Therefore, the overall complexity of our approach is polynomial in the number of points in the point cloud which is equal to $O(nd^2)+O(n^{3.5})\approx O(n^{3.5})$, since $d<<n$. It takes around 38.4 seconds ($d=3$) to find the symmetry plane and all the pairs of mirror symmetric points in a point cloud with 500 points using MATLAB on a Linux machine with i7-7500U CPU @ 2.70GHz, and 16GB RAM.

\begin{figure}[!htbp]
	\centering
	\stackunder{\epsfig{figure=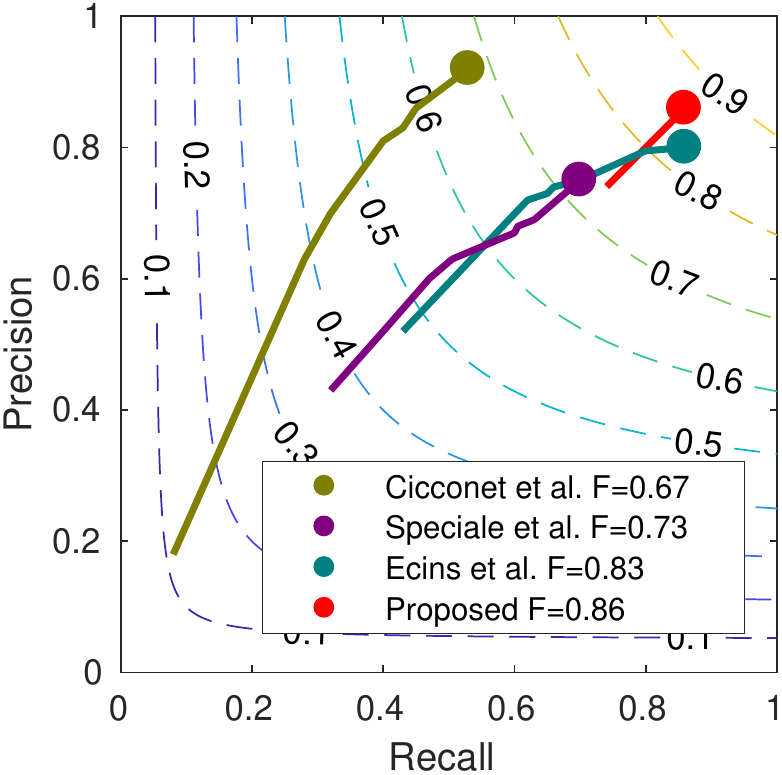,width=1\linewidth}}{}
	\caption{Recall vs Precision curves for methods Cicconet \emph{et al.} \cite{Cicconet_2017_ICCV_Workshops}, Ecins \emph{et al.} \cite{ecins2017detecting}, Speciale \emph{et al.} \cite{speciale2016symmetry}, and the proposed approach on the dataset given \cite{Funk_2017_ICCV_Workshops}. We show the maximum F-score for each method in the legends and corresponding points on the precision vs. recall curve using the same colored point.}
	\label{fig:eval}
\end{figure}
\section{Evaluation and Results}
\label{sec:RD}
\subsection{Evaluation of Reflection Symmetry Plane}
In order to evaluate the performance of reflection symmetry plane detection, we compare the performance of our approach with the performance of the methods in \cite{ecins2017detecting}, \cite{Cicconet_2017_ICCV_Workshops}, and \cite{speciale2016symmetry}. We compare the detected plane of reflection symmetry to that of these methods on the dataset in \cite{Funk_2017_ICCV_Workshops} with F-score as the evaluation metric proposed in \cite{Funk_2017_ICCV_Workshops}.  The dataset given in \cite{Funk_2017_ICCV_Workshops} contains models of 1354 3D real world objects in which the ground-truth plane of reflection symmetry is provided for all the objects.  
 \begin{figure}[!htbp]
 	\centering
 	\stackunder{\includegraphics[width=0.8\linewidth]{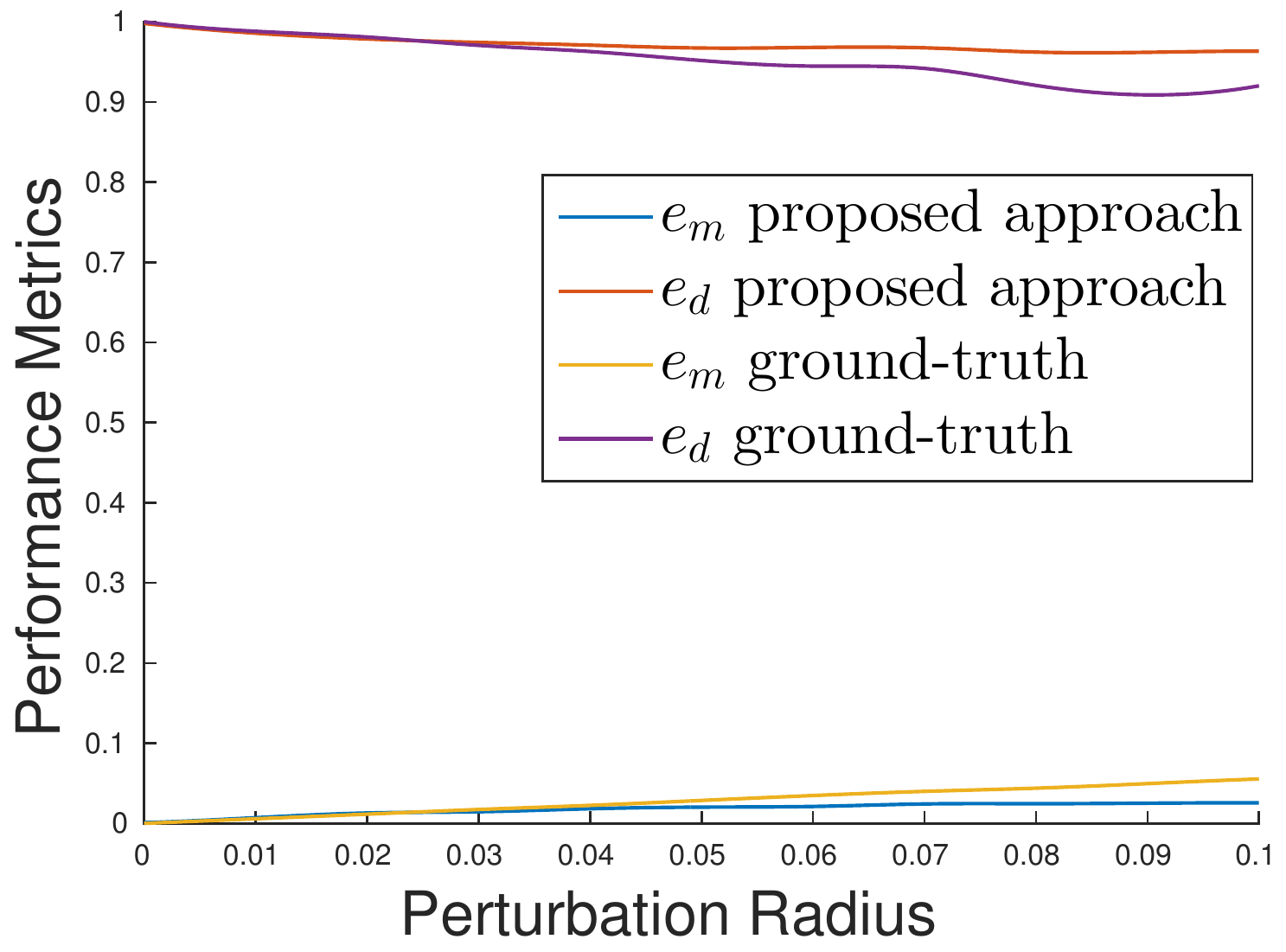}}{(a)}
 	\stackunder{\includegraphics[width=0.8\linewidth]{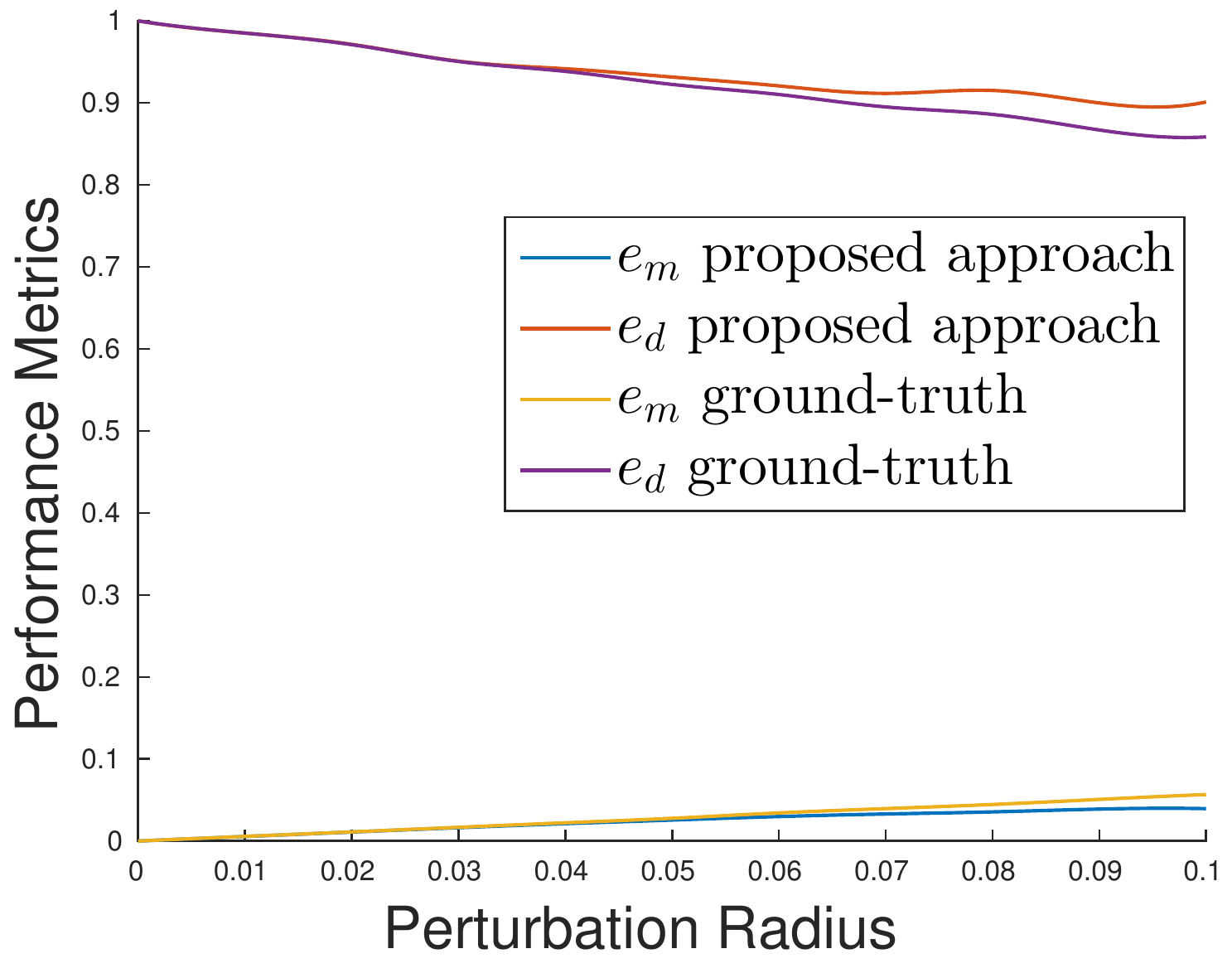}}{(b)}
 	\caption{The values $e_\text{d}$ and $e_\text{m}$ vs the perturbation radius $\sigma^2$. (a) d=2, and (b) d=3. We observe that the performance measure quantities $e_\text{d}$ and $e_\text{m}$ remain close to that of the ground truth quantities.}
 	\label{fig:perf}
 \end{figure}
 
 Speciale \emph{et al.} proposed a Hough transform voting based approach \cite{speciale2016symmetry}. Ecins \emph{et al.} proposed an ICP based approach \cite{ecins2017detecting}. First, they initialize the reflection symmetry plane and then iteratively update the reflection symmetry plane using the Levenberg-Marquardt solver till convergence. They  have further used the normals at each point to reject outliers points. Therefore, they need oriented point clouds, i.e., normal at each point be given.  Cicconet \emph{et al.} first reflected the original point cloud about an arbitrary reflection plane and then used the ICP algorithm to align the original point cloud and the reflected point cloud \cite{Cicconet_2017_ICCV_Workshops}. Then, they determine the reflection symmetry plane.
    
 In order to evaluate the accuracy of detecting reflection symmetry plane for each method, we find the precision and recall rates  and the $F$-score. According to \cite{Funk_2017_ICCV_Workshops}, the precision and the recall rates are defined as $P=\frac{TP}{TP+FP}$, $R=\frac{TP}{TP+FN}$, respectively. The $F$-Score is defined as $F=\frac{2RP}{R+P}$. According to \cite{Funk_2017_ICCV_Workshops}, $TP$ is equal to the number of correctly  estimated reflection symmetry planes, $FP$ is  equal to the number of incorrectly estimated reflection symmetry planes, and $FN$ is equal to the number of  ground-truth reflection symmetry planes which are not detected. According to \cite{Funk_2017_ICCV_Workshops}, a detected plane of reflective symmetry is declared to be correct or incorrect as follows. Let $\mathbf{x}_1^\text{e}$, $\mathbf{x}_2^\text{e}$, and $\mathbf{x}_3^\text{e}$ be three points on the detected plane of reflection symmetry. Let  $\mathbf{x}_1^\text{g}$, $\mathbf{x}_2^\text{g}$, and $\mathbf{x}_3^\text{g}$  be three points on  the ground truth plane of reflection symmetry of the underlying symmetric object. These three points on the plane of reflection symmetry planes are any three points from the four points of intersection of the plane of reflection symmetry with the bounding box of the underlying reflective symmetric object. Now, according to \cite{Funk_2017_ICCV_Workshops}, the detected plane of reflection symmetry is declared correct if the angle between the normal of the detected plane of reflection symmetry, which is defined as $\boldsymbol{\eta}_\text{e}=(\mathbf{x}_1^\text{e}-\mathbf{x}_2^\text{e})\times(\mathbf{x}_1^\text{e}-\mathbf{x}_3^\text{e})$, and the normal of the ground truth plane of reflection symmetry, which is defined as $\boldsymbol{\eta}_\text{g}=(\mathbf{x}_1^\text{g}-\mathbf{x}_2^\text{g})\times(\mathbf{x}_1^\text{g}-\mathbf{x}_3^\text{g})$, is less than a predefined threshold, i.e., $\cos^{-1}\left(\frac{|\boldsymbol{\eta}_\text{e}^\top\boldsymbol{\eta}_\text{g}|}{\|\boldsymbol{\eta}_\text{e}\|_2\|\boldsymbol{\eta}_\text{g}\|_2}\right)<t_\theta$. Furthermore, according to \cite{Funk_2017_ICCV_Workshops}, the distance between the center of the detected plane of reflection symmetry, which defined as  $\mathbf{c}_\text{e}=\frac{\mathbf{x}_1^\text{e}+\mathbf{x}_2^\text{e}}{2}$,  from the ground truth plane of reflection symmetry is less than a predefined threshold, i.e., $\frac{|\mathbf{c}_\text{e}^\top\boldsymbol{\eta}_\text{g}-\boldsymbol{\eta}_\text{g}^\top\mathbf{x}_1^\text{g}|}{\|\boldsymbol{\eta}_\text{g}\|_2}<t_d$. In order to find the precision vs. recall curve,  we change the  threshold for angle as $t_\theta\in[0,45^\circ]$ and the threshold for distance as $t_d\in[0,2s]$. Here, $s=\min\{\|\mathbf{x}_1^\text{e}-\mathbf{x}_2^\text{e}\|_2,\|\mathbf{x}_1^\text{e}-\mathbf{x}_3^\text{e}\|_2,\|\mathbf{x}_1^\text{g}-\mathbf{x}_2^\text{g}\|_2,\|\mathbf{x}_1^\text{g}-\mathbf{x}_3^\text{g}\|_2\}$.  In Fig. \ref{fig:eval}, we plot the recall vs. precision curves for the methods in \cite{Cicconet_2017_ICCV_Workshops}, \cite{ecins2017detecting}, \cite{speciale2016symmetry}, and the proposed approach on the dataset given in \cite{Funk_2017_ICCV_Workshops}. We show the maximum $F$-score for each method in the legends. The maximum $F$-score for \cite{ecins2017detecting} is equal to 0.83, for  \cite{Cicconet_2017_ICCV_Workshops} is equal to 0.67, for  \cite{speciale2016symmetry} is equal to 0.73, and for the proposed approach is equal to 0.86. 
\subsection{Robustness to Perturbations} 
\label{subsec:rps}
\begin{figure}[!htbp]
	\centering
	\stackunder{\epsfig{figure=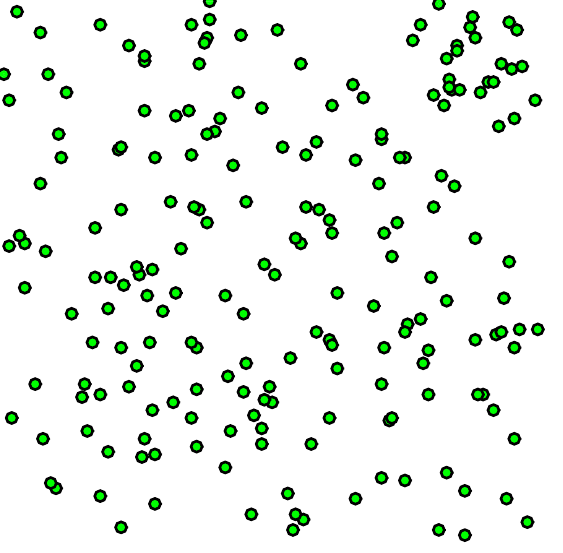,width=0.75\linewidth}}{}
	\caption{An example point set for $d=2$.}
	\label{fig:ex_2}
\end{figure}

In order to measure the qualitative performance of the proposed approach, we investigate the following two errors which are functions of the perturbation radius $\sigma^2$: 
$$e_{\text{d}}=\frac{1}{n}\sum_{i=1}^n\mid\left\langle\hat{\mathbf{z}}_i,\hat{\mathbf{v}}\right\rangle\mid\text{ and } e_{\text{m}}=\frac{1}{n}\sum_{i=1}^n\mid\hat{\mathbf{v}}^\top\mathbf{x}_i^\text{m}+w_0\mid.$$ 
The error $e_{\text{d}}$ represents how well the vectors, along the line segments joining the estimated reflection symmetry points, align with the normal to the hyperplane $\boldsymbol{\pi}$ at convergence. The error $e_{\text{m}}$ represents how well the mid-points of line segments joining reflection symmetry points lie on the estimated hyperplane $\boldsymbol{\pi}$. Here, $\hat{\mathbf{z}}_i=\frac{\mathbf{x}_i-\mathbf{x}_{i^\prime}}{\|\mathbf{x}_i-\mathbf{x}_{i^\prime}\|_2}$, $\hat{\mathbf{v}}$ is the unit normal to the hyperplane $\boldsymbol{\pi}$,  $\mathbf{x}_i^\text{m}=\frac{\mathbf{x}_i+\mathbf{x}_{i^\prime}}{2}$, and $w_0$ is the distance of the hyperplane $\boldsymbol{\pi}$ from the origin. In Fig. \ref{fig:perf}, we show the errors $e_\text{d}$ and $e_\text{m}$ against the perturbation radius $\sigma^2$.  We observe that the values $e_{\text{m}}$ and $e_{\text{d}}$ for the proposed approach are close to that of the ground-truth reflection symmetry even as the value of $\sigma^2$ increases.  We construct the following dataset to perform the above experiment. Let $\{\mathbf{x}_1,\mathbf{x}_2,\ldots,\mathbf{x}_{\frac{n}{2}}\}$  be the randomly chosen  $\frac{n}{2}$ points. Given the reflection transformations  $\{\mathbf{R}_1,\ldots,\mathbf{R}_{d-1},\mathbf{t}\}$, we reflect these points using the \textit{Definition 1} in order to get the final symmetric set  $\mathcal{S}=\{\mathbf{x}_1,\mathbf{x}_2,\ldots,\mathbf{x}_{\frac{n}{2}},\mathbf{x}_1^\prime,\mathbf{x}_2^\prime,\ldots,\mathbf{x}_{\frac{n}{2}}^\prime\}$. Then, we perturb each point with random noise  as $\mathbf{x}\gets\mathbf{x}+\mathcal{N}(\mathbf{0}_{d},\text{diag}(\sigma^2,\sigma^2,\ldots,\sigma^2)),\forall\mathbf{x}\in\mathcal{S}$. Here, $\sigma^2$ is the perturbation radius and the perturbation is different for each point. For the case $d=2$, we create sets containing reflection symmetry patterns with $n\in\{50, 100, 150, 200, 250, 300\}$ with $0\leq x,y\leq 1$. For each $n$, we take 19 different symmetry axis orientations in the range from $-90^\circ$ to $90^\circ$ with step size of $10^\circ$. We choose $\sigma^2\in\{0,0.01,0.02,\ldots,0.1\}$ to get 11 different perturbations. In total, we have 1254 sets for the evaluation. In Fig. \ref{fig:ex_2}, we show an example point set from this dataset.  For the case $d=3$, we create reflective symmetric sets  with $n\in\{50, 100, 150, 200, 250, 300\}$ with $0\leq x,y\leq 1$. For each $n$, we take 16 different symmetry plane orientations by considering $\theta_1\in\{-30^\circ, 0^\circ, 35^\circ, 80^\circ\}$ and  $\theta_2\in\{-30^\circ, 0^\circ, 35^\circ, 80^\circ \}$. We choose $\sigma^2\in\{0,0.01,\ldots,0.1\}$. In total, we obtain 1056 point sets.

\subsection{Evaluation in Higher Dimensional Data}
\begin{figure}[!htbp]
	\centering
	\stackunder{\epsfig{figure=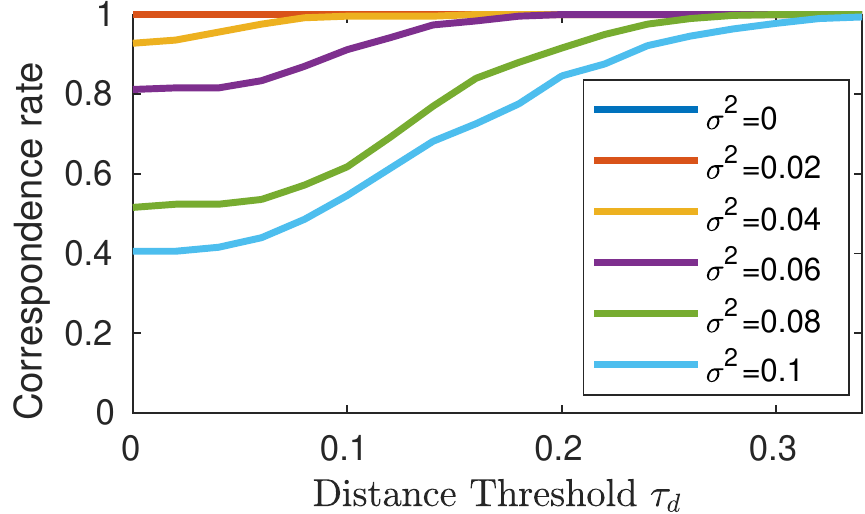,width=0.9\linewidth}}{$d=6$}
	\stackunder{\epsfig{figure=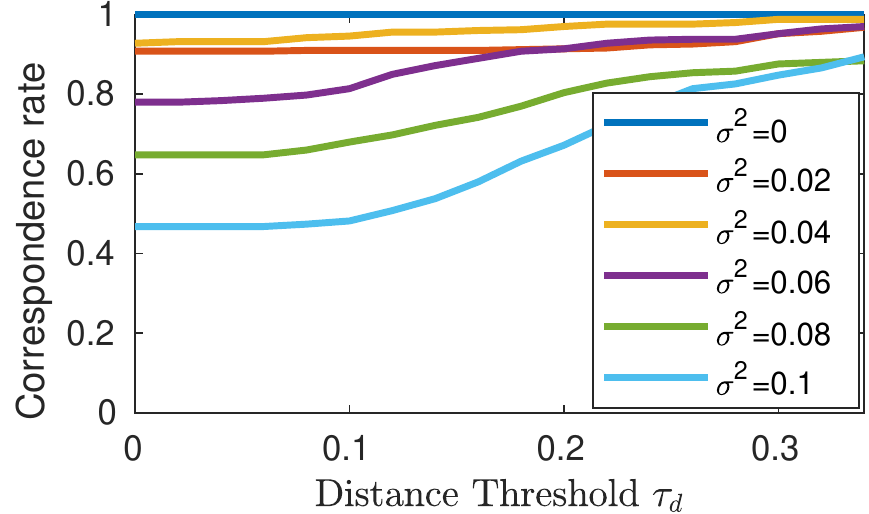,width=0.9\linewidth}}{$d=8$}
	\caption{Correspondence rate vs distance threshold curves for $d=6$ and $d=8$.}
	\label{fig:eval_6_8_c}
\end{figure}

\begin{figure}[!htbp]
	\centering
	\stackunder{\epsfig{figure=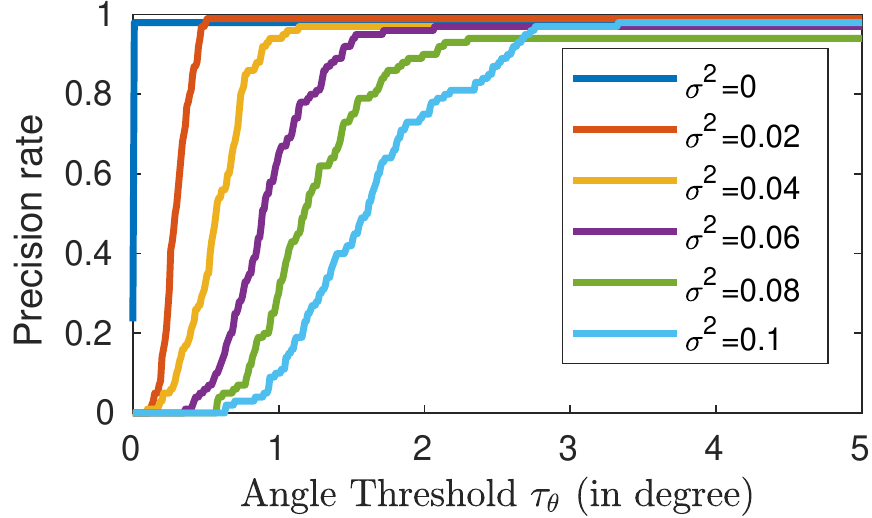,width=0.9\linewidth}}{$d=6$}
	\stackunder{\epsfig{figure=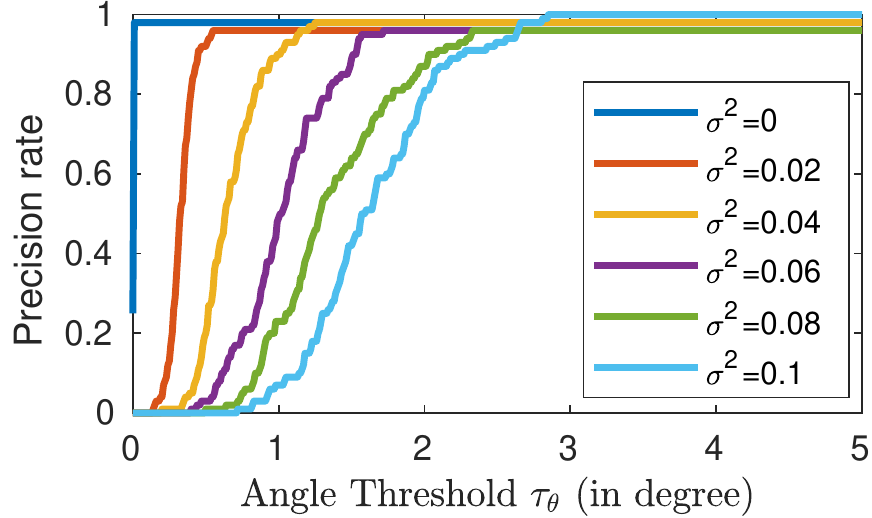,width=0.9\linewidth}}{$d=8$}
	\caption{Precision rate vs angle threshold curves for $d=6$ and $d=8$.}
	\label{fig:eval_6_8_p}
\end{figure}

\textbf{Datasets.} Since datasets for higher dimensions ($d>3$) are not available with ground-truth reflection symmetry, we synthetically create  datasets as follows. For the case $d=6$ and $d=8$, we create mirror symmetric point clouds using Definition 1, with $n\in\{50, 100, 150, 200, 250, 300\}$ and $0\leq x,y\leq 1$. For each $n$, we take 20 random symmetry plane normals. We choose $\sigma^2\in\{0,0.02,0.04,\ldots,0.1\}$ to get 6 different perturbations. In total, we have 720 sets for evaluation. For all these point clouds, we have the ground-truth correspondences between the symmetric points and the normals to the ground-truth  symmetry planes. \\
 \textbf{Evaluation of correspondences.} In order to evaluate the performance, we measure the correspondence rate which is the number of correct correspondences out of the estimated correspondences.  Let $(i,i_{\text{e}}^\prime)$ be the estimated correspondence and let $(i,i_{\text{g}}^\prime)$ be the ground-truth correspondence. Then, we decide if the estimated correspondence $(i,i_{\text{e}}^\prime)$ is correct based on a distance threshold $\tau_d$. If the  distance $\|\mathbf{x}_{i_{\text{e}}^\prime}-\mathbf{x}_{i_{\text{g}}^\prime}\|_2$ between the points $\mathbf{x}_{i_\text{e}^\prime}$ and $\mathbf{x}_{i_\text{g}^\prime}$ is less than the distance threshold $\tau_d$, then the correspondence $(i,i_{\text{e}}^\prime)$ is correct and otherwise, incorrect. For a given threshold $\tau_d$, we count the correspondences $(i,i_{\text{e}}^\prime)$ for which the condition $\|\mathbf{x}_{i_{\text{e}}^\prime}-\mathbf{x}_{i_{\text{g}}^\prime}\|_2<\tau_d$ holds true. In Fig. \ref{fig:eval_6_8_c}, we show the correspondence rate vs the distance threshold curves for the different perturbation radius $\sigma^2\in\{0,0.02,0.04,\ldots,0.1\}$ and for $d=6$ and $d=8$. We vary the distance threshold as $\tau_d\in\{0,0.01,0.02,\ldots,0.34\}$.  We observe that the correspondence rate increases as the distance threshold increases and the correspondence rate decreases as the perturbation radius increases for both $d=6$ and $d=8$. \\
 \textbf{Evaluation of symmetry plane.} To evaluate the performance of the reflection plane detection in higher dimensional point clouds  ($d>3$), instead of finding $d-1$ points on the estimated hyperplane (since finding $d-1$ points could be difficult), we measure the distance between their normals. Without loss of generality, we prepare the dataset  such that the reflection symmetry plane passes through the origin. Now, let $\boldsymbol{\eta}^\text{g}$ and $\boldsymbol{\eta}^\text{e}$ be the unit normals to the ground-truth and the estimated reflection symmetry planes, respectively. Then, we declare the estimated reflection symmetry plane to be correct, if $\cos^{-1}(|(\boldsymbol{\eta}^\text{g})^\top\boldsymbol{\eta}^\text{e}|)<\tau_\theta$. We vary the angle threshold $\tau_\theta$ in the range $[0^\circ,5^\circ]$ with a step size of $0.01^\circ$.   In Fig. \ref{fig:eval_6_8_p}, we show the precision rate vs the angle threshold $\tau_\theta$ curves for  different perturbation radius $\sigma^2\in\{0,0.02,0.04,\ldots,0.1\}$ and for $d=6$ and $d=8$. We observe that the precision rate increases as the angle threshold increases and  decreases as the perturbation radius increases for both $d=6$ and $d=8$. 

\begin{figure}[!htbp]
	\centering
	\stackunder{\epsfig{figure=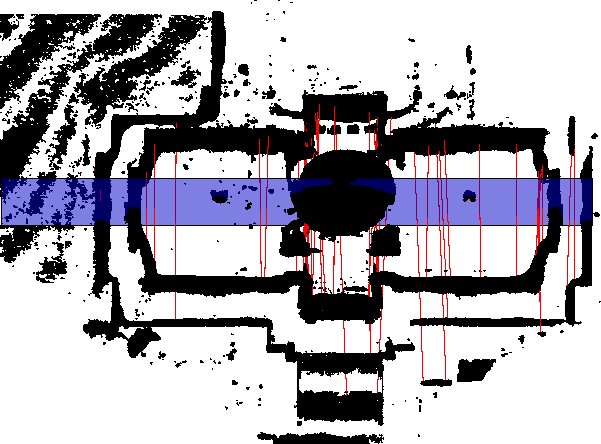,width=0.85\linewidth}}{}
	\stackunder{\epsfig{figure=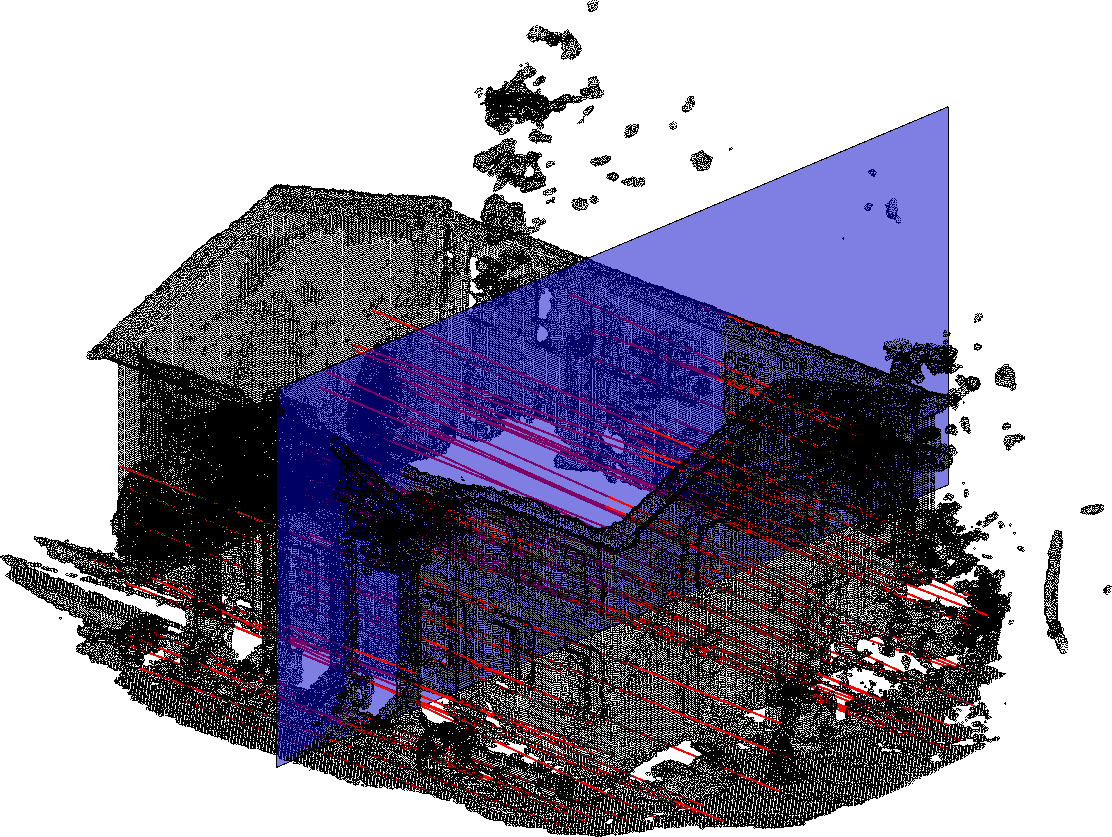,width=0.9\linewidth}}{}
	\caption{Detected reflection symmetry on two real 3D scans of buildings from the dataset \cite{Funk_2017_ICCV_Workshops}.}
	\label{fig:rs}
\end{figure} 
\begin{figure*}[htbp]
	\centering
	\includegraphics[width=0.16\linewidth]{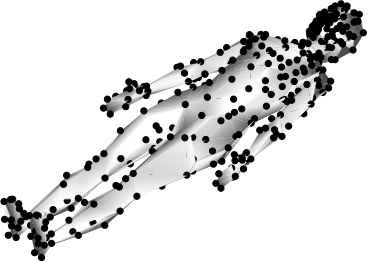}
	\includegraphics[width=0.18\linewidth]{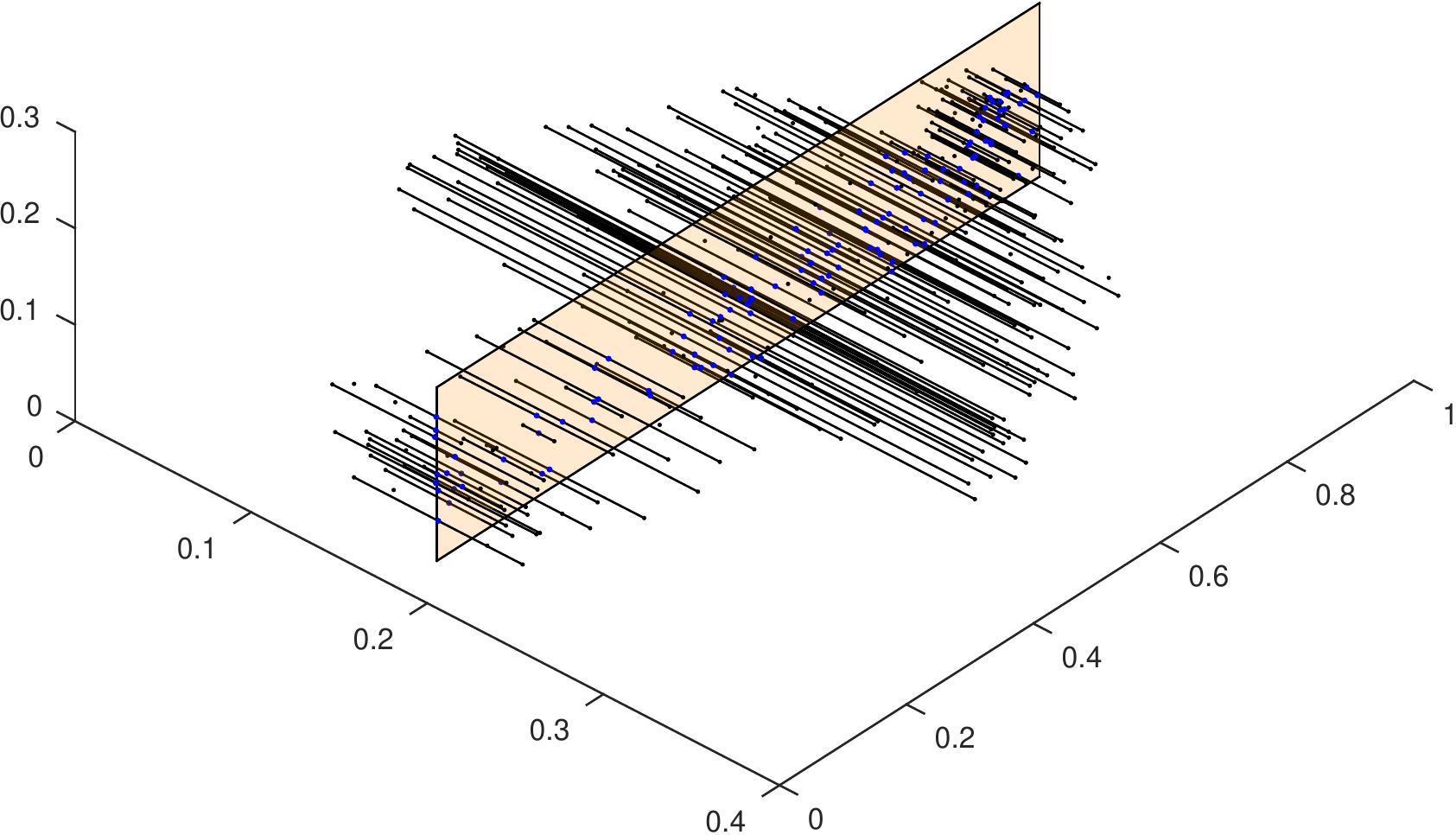}
	\includegraphics[width=0.1\linewidth]{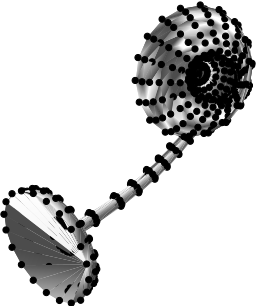}                  
	\includegraphics[width=0.18\linewidth]{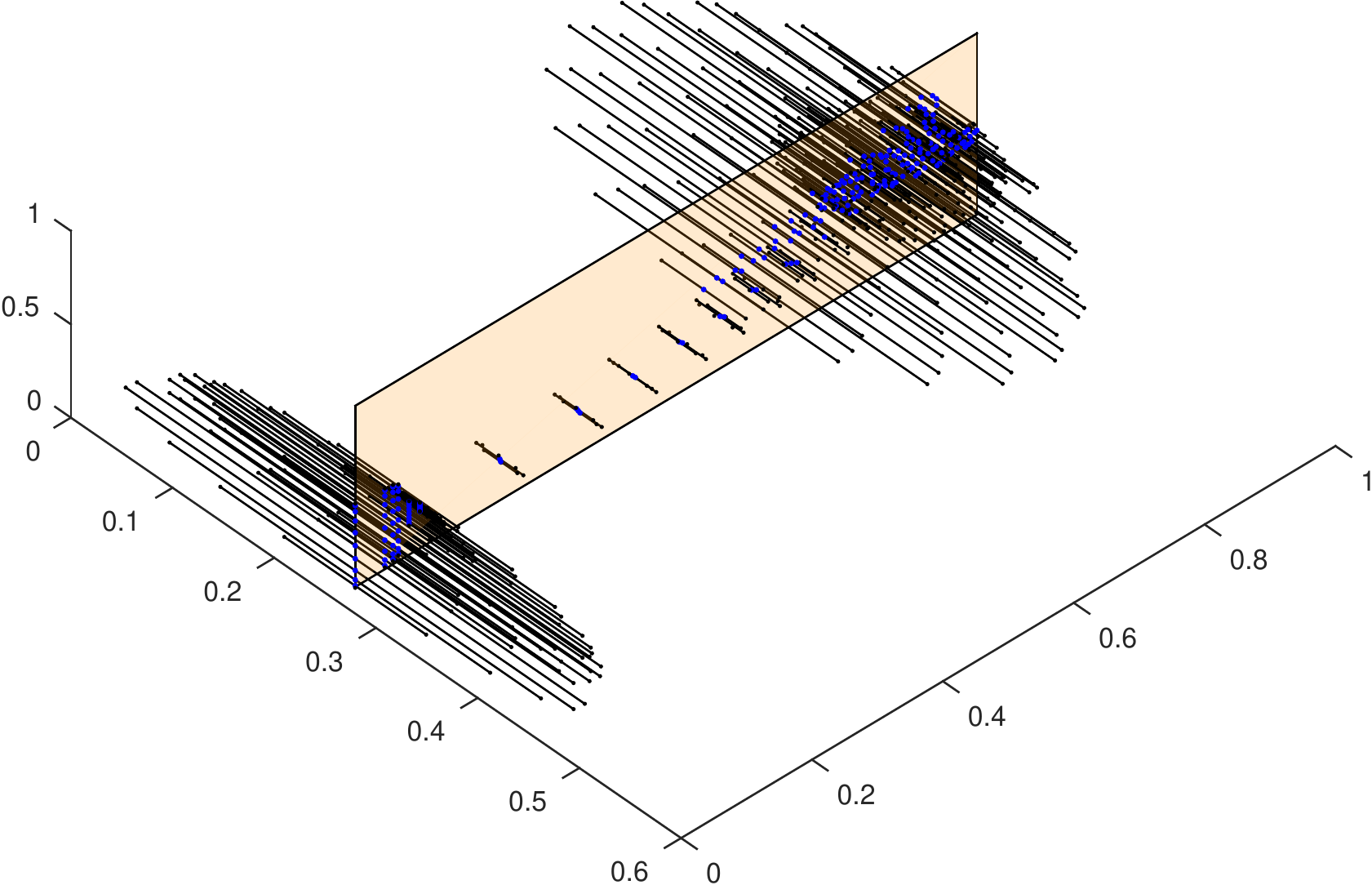}
	\includegraphics[width=0.17\linewidth]{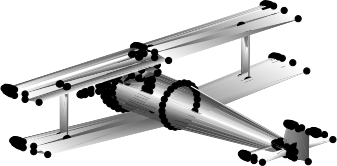}
	\includegraphics[width=0.18\linewidth]{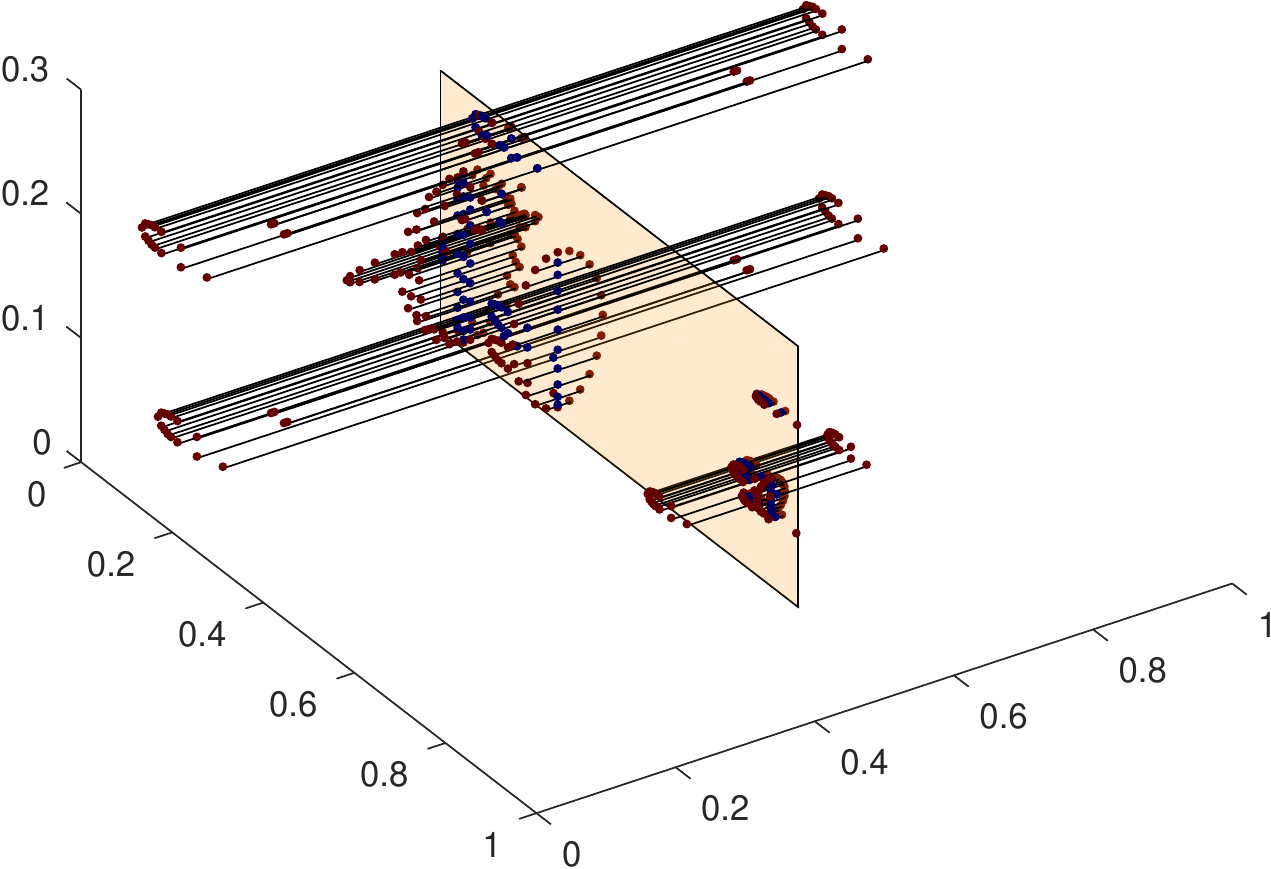}
	\includegraphics[width=0.16\linewidth]{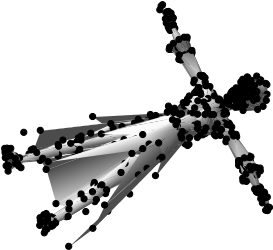}
	\includegraphics[width=0.18\linewidth]{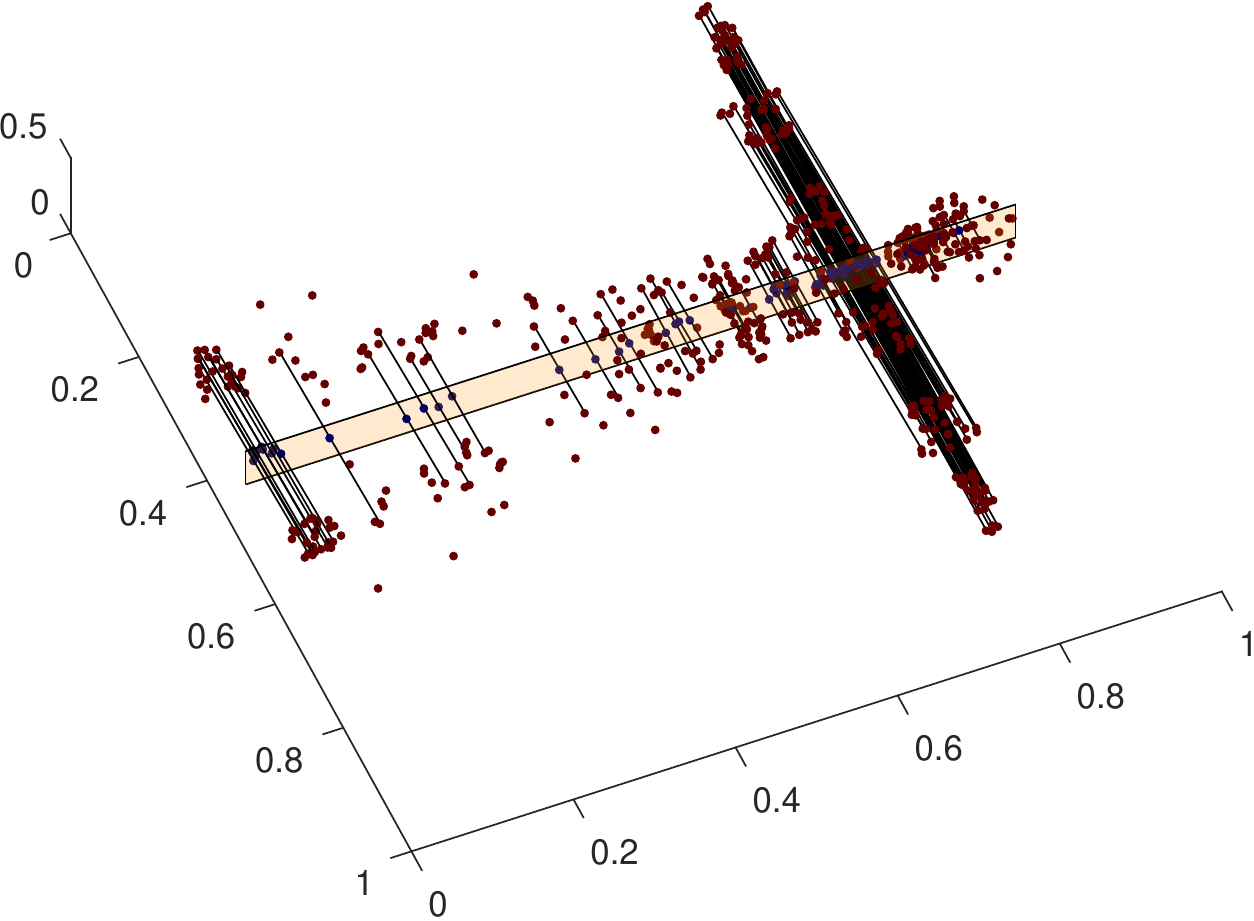}  
	\includegraphics[width=0.12\linewidth]{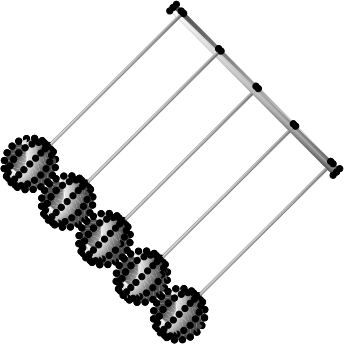}    
	\includegraphics[width=0.18\linewidth]{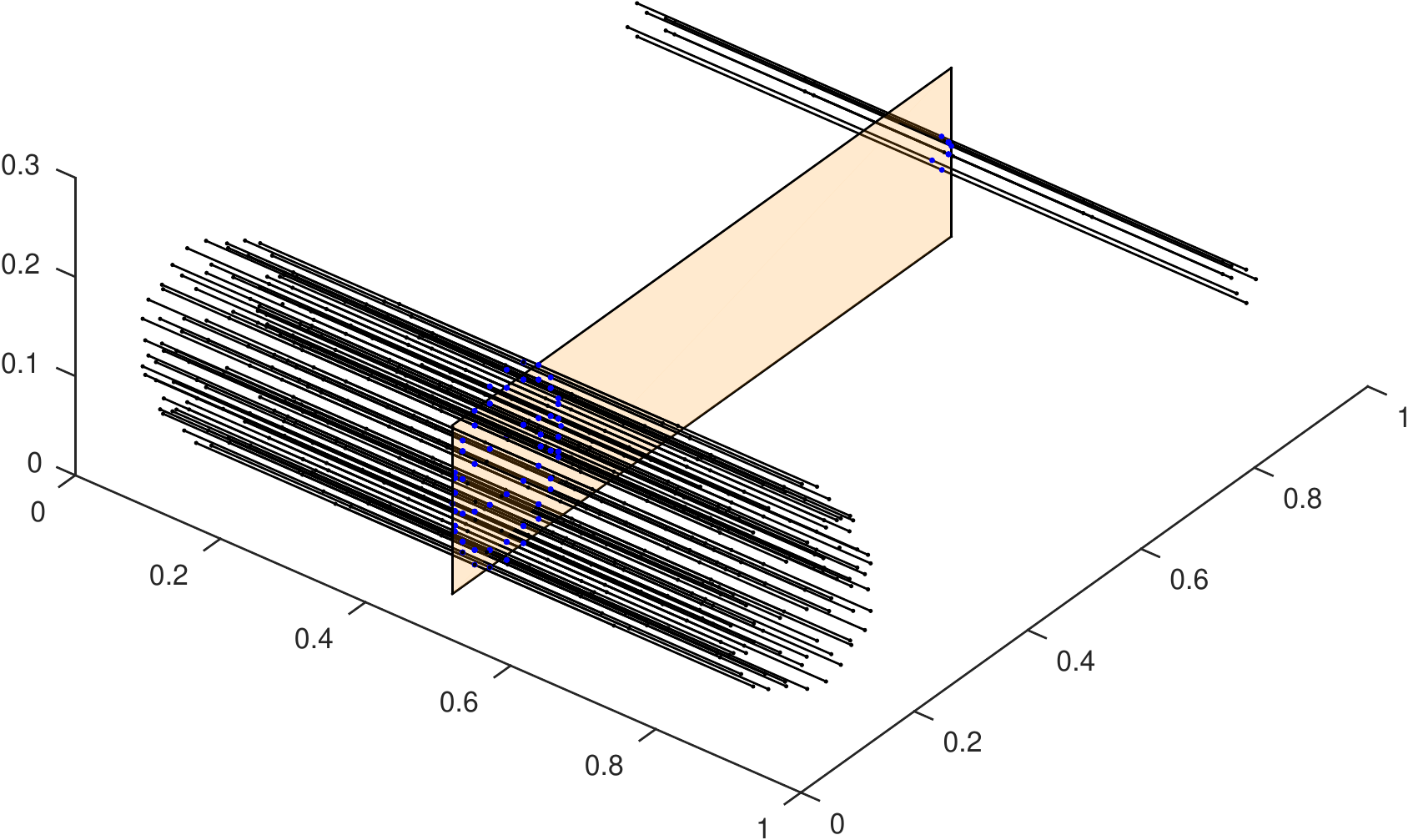}
	\includegraphics[width=0.12\linewidth]{m1667_3.png}
	\includegraphics[width=0.18\linewidth]{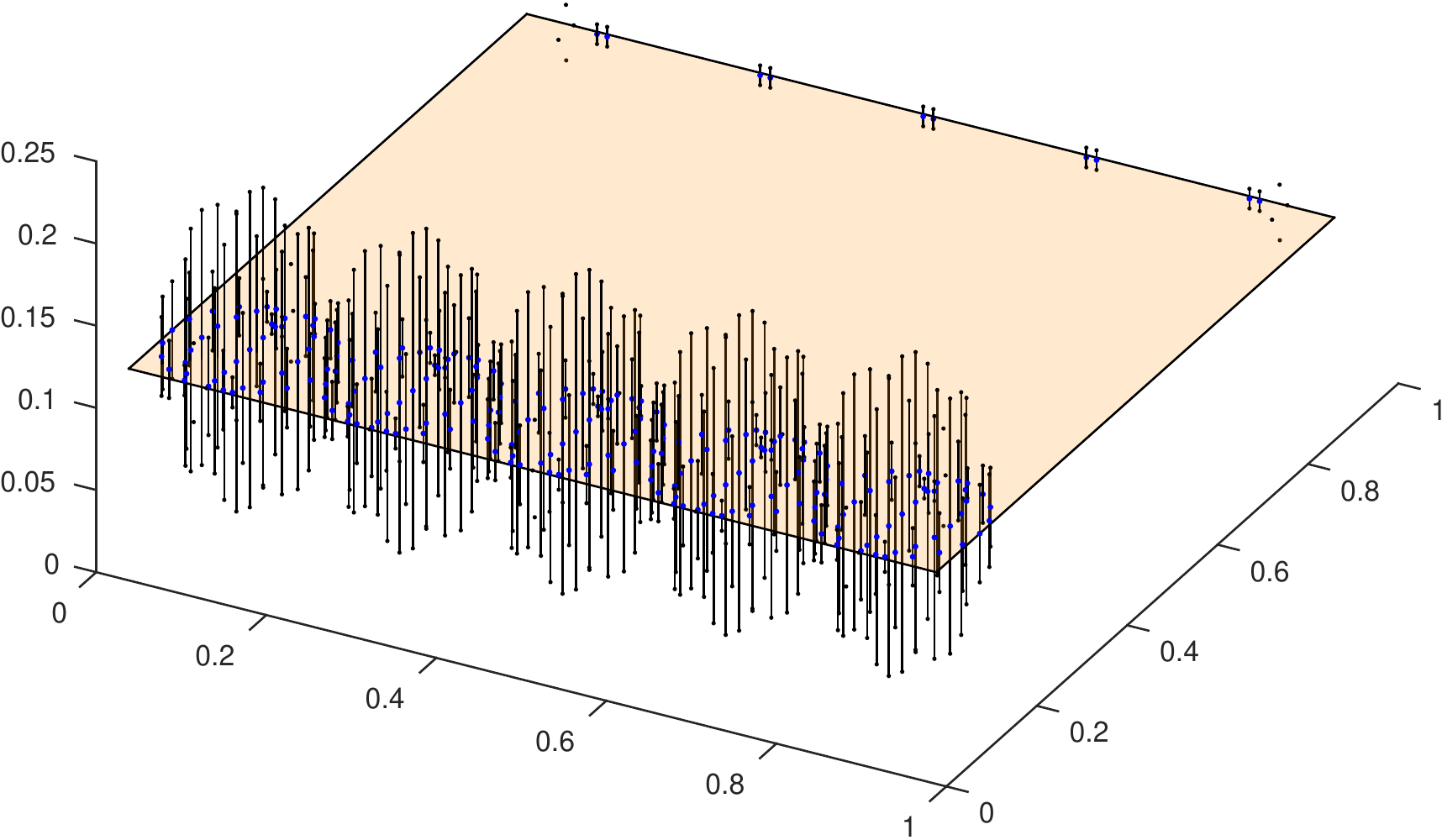}             
	\caption{Results of symmetry detection in the 3D object models from the dataset \cite{shilane2004princeton}. In the first and third columns we show the point set on the original surface. And in the second and fourth columns we show the detected reflection symmetry.  The correspondences are shown by joining the mirror symmetric points by the black colored lines. The Reflection symmetry plane is shown in light brown color. The mid-points of the mirror symmetric points are show in blue color.  Here, we show the surface for visualization purpose only.}
	\label{fig:im3d_2}
\end{figure*}
\subsection{Results} 
\begin{figure*}[!htbp]
	\centering
	\includegraphics[width=0.11\linewidth]{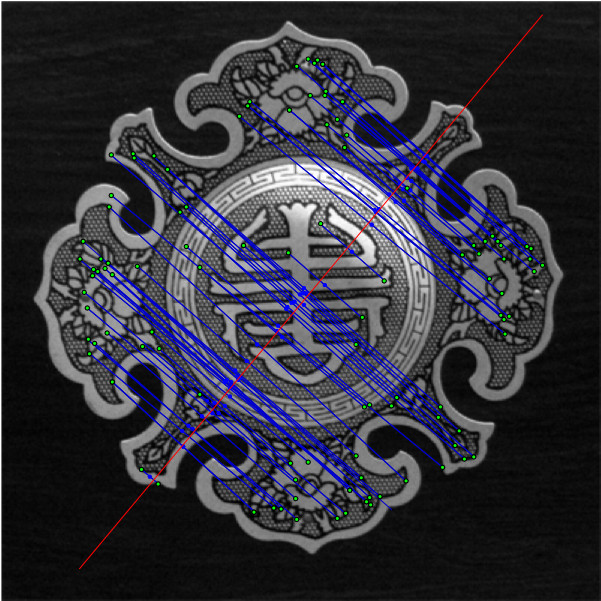}
	\includegraphics[width=0.115\linewidth]{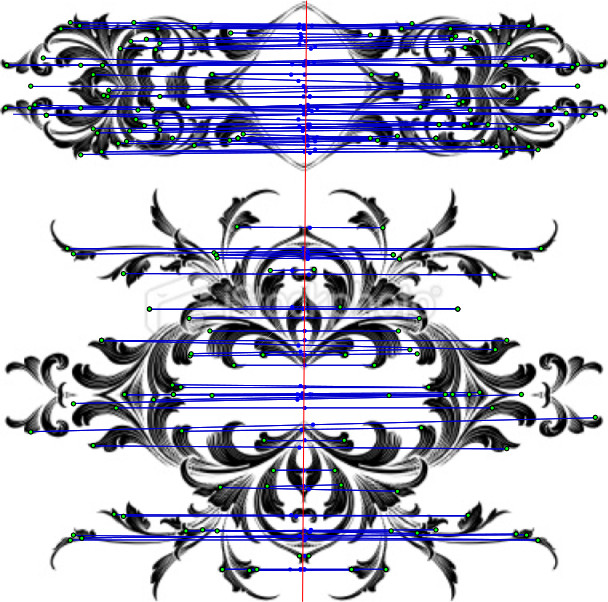}
	\includegraphics[width=0.11\linewidth]{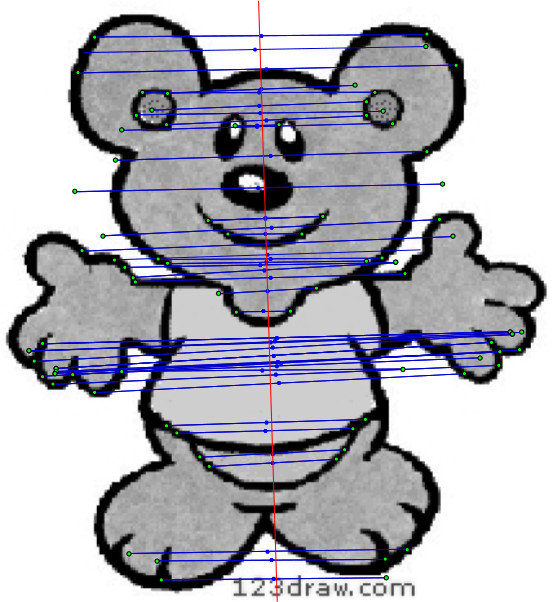}
	\includegraphics[width=0.15\linewidth]{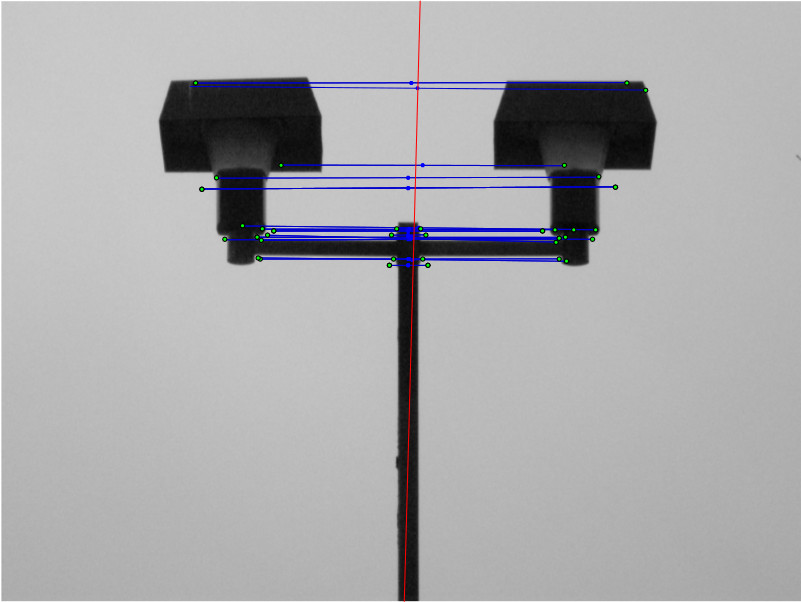}
	\includegraphics[width=0.15\linewidth]{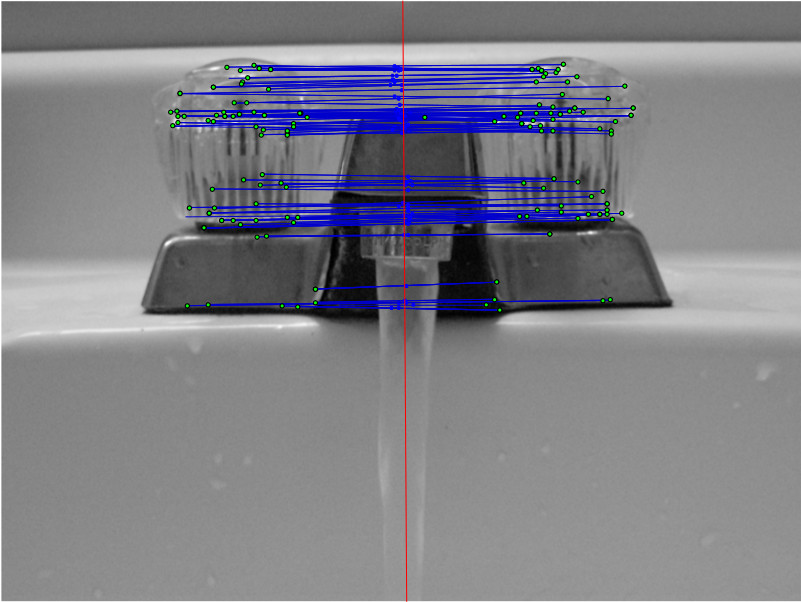}
	\includegraphics[width=0.08\linewidth]{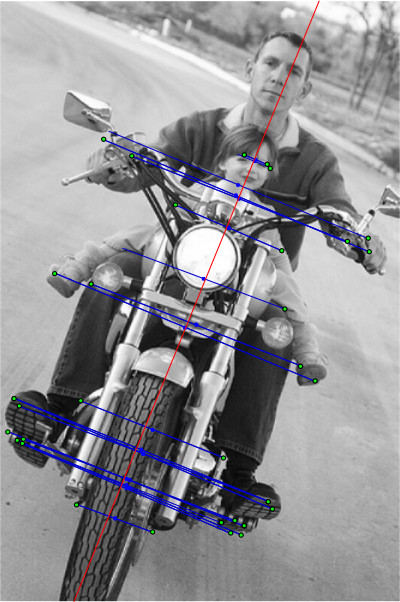}
	\includegraphics[width=0.09\linewidth]{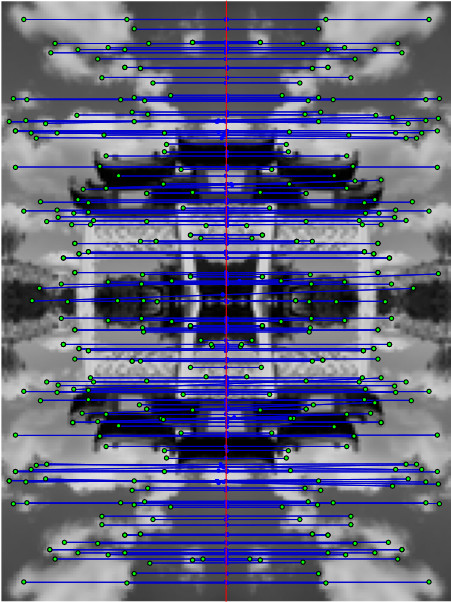}
	\includegraphics[width=0.084\linewidth]{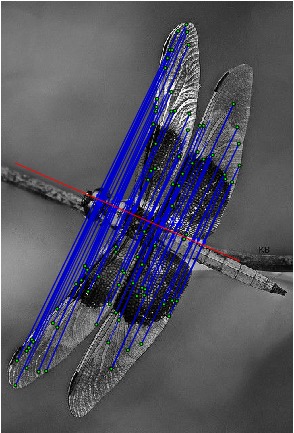}
	\includegraphics[width=0.066\linewidth]{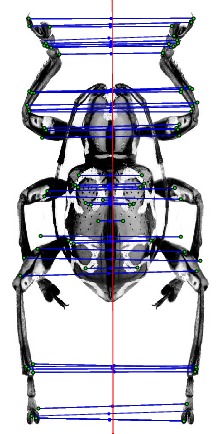}
	\includegraphics[width=0.076\linewidth]{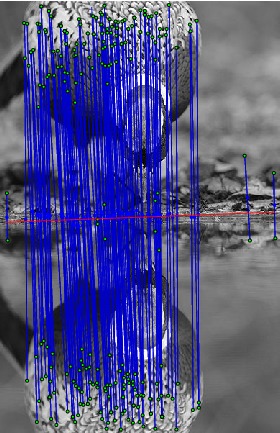}
	\includegraphics[width=0.103\linewidth]{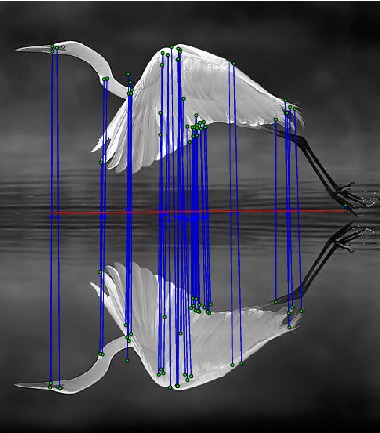}
	\includegraphics[width=0.155\linewidth]{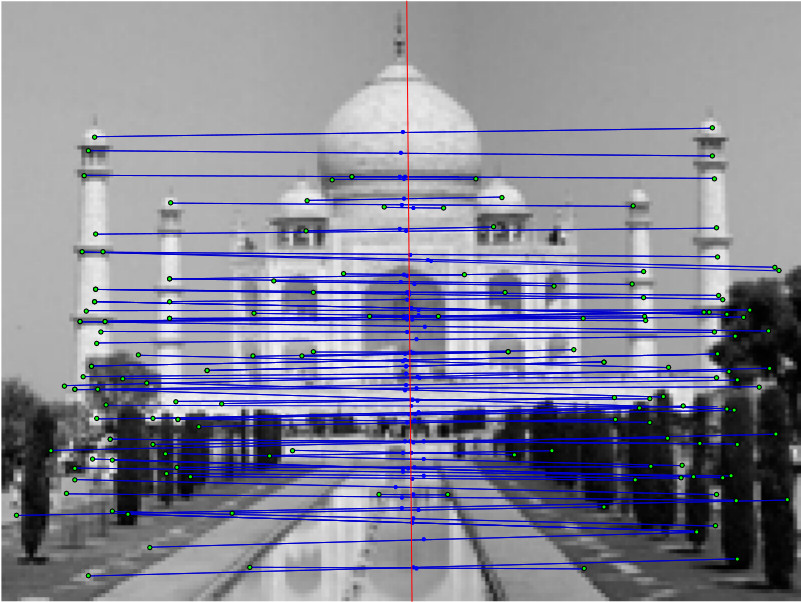}
	\includegraphics[width=0.205\linewidth]{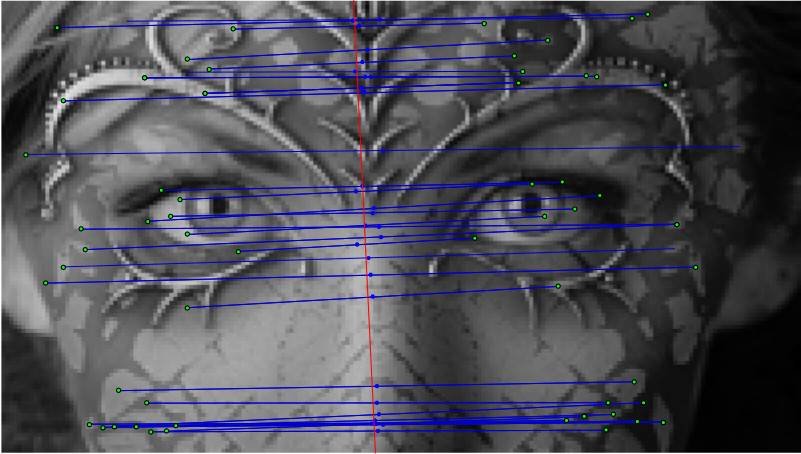}
	\includegraphics[width=0.168\linewidth]{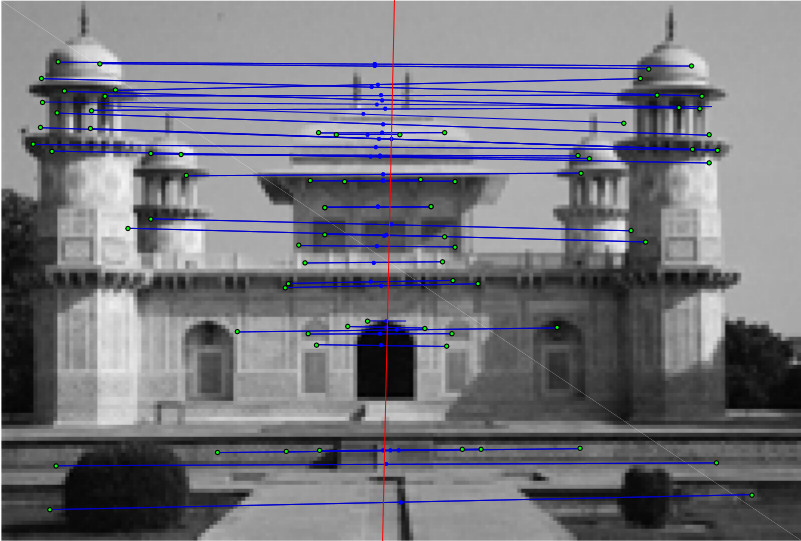}
	\includegraphics[width=0.153\linewidth]{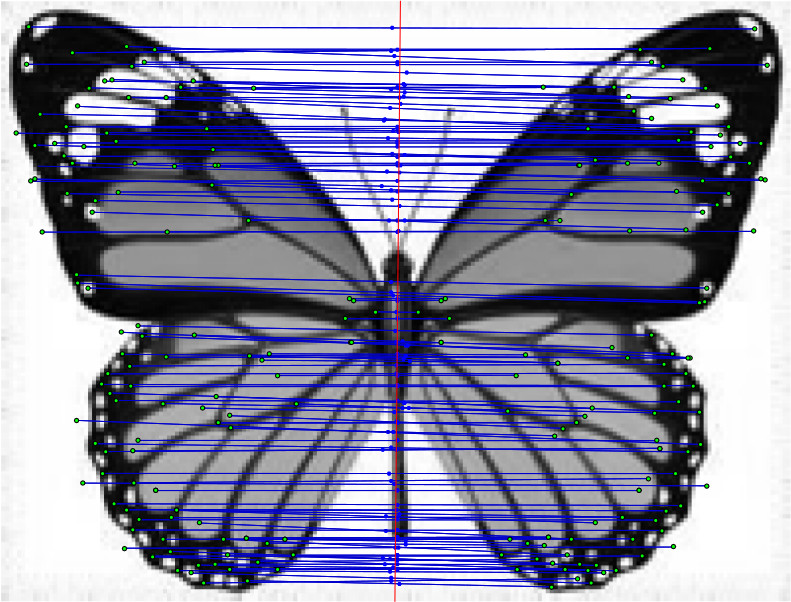}
	\epsfig{figure=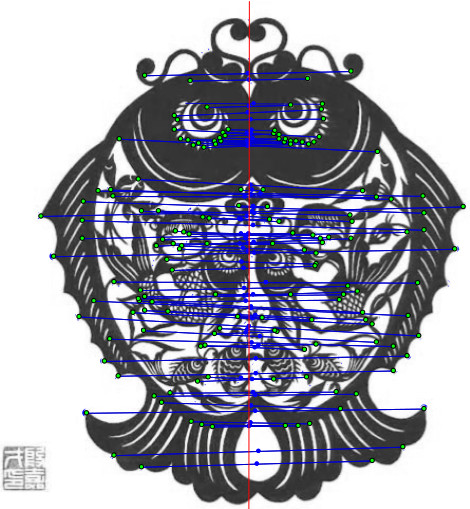,width=0.1\linewidth}
	\epsfig{figure=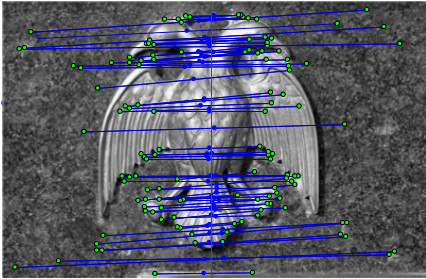,width=0.153\linewidth}
	\epsfig{figure=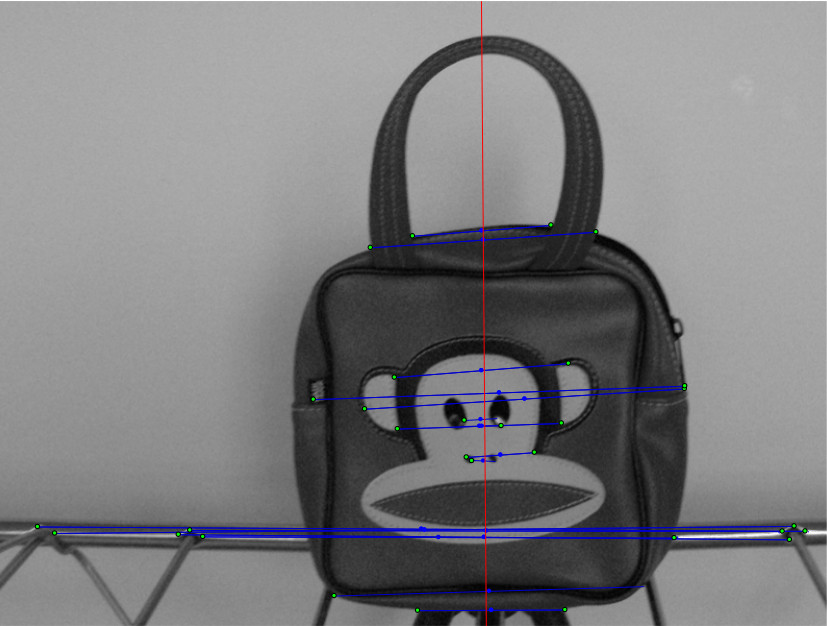,width=0.138\linewidth}
	\epsfig{figure=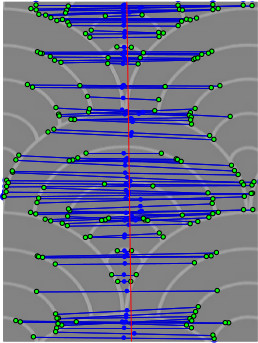,width=0.08\linewidth}
	\epsfig{figure=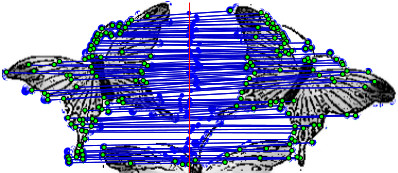,width=0.22\linewidth}
	\epsfig{figure=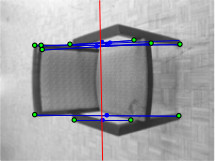,width=0.13\linewidth}
	\epsfig{figure=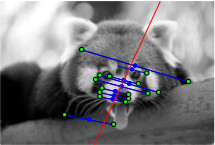,width=0.14\linewidth}
	\caption{Results of symmetry detection in real images from the dataset \cite{liu2013symmetry,cicconet2016convolutional,rauschert2011symmetry1}. We show the set $\mathcal{S}$ using green points, the reflection symmetry axis by a red line, and the correspondences between the mirror symmetric points by the blue lines.}
	\label{fig:im_2d}
\end{figure*}

In Fig. \ref{fig:rs},  we show the detected reflection symmetry for two real 3D scans of buildings from the dataset \cite{Funk_2017_ICCV_Workshops}. In Fig. \ref{fig:im3d_2}, we present the results for the case $d=3$.  The point cloud in Fig. \ref{fig:rs}(a) contains 912045 points and the point cloud in Fig. \ref{fig:rs}(b) contains 767474 points. Since the computational complexity is $O(n^{3.5})+O(nd^2)$, the computation time and space requirement (storing the matrices $\mathbf{A}_1$ and $\mathbf{A}_2$) are very high. Therefore, in order to compute the reflection symmetry in these scans, we randomly sample around 600 points. In both cases, we show the reflection symmetry plane by the blue color and estimated pairs of reflective symmetric points by the red colored line segment joining them.  In order to make our algorithm robust to the part removal, we simply put the extra constraint $\mathbf{e}^\top\mathbf{Pe}\leq 2k$ in ILP defined in Equation \eqref{eq:17} which limits the number of pairs to at most $k$.  For $d=2$, we detect reflection symmetry in the set of corner points in a real image. In order to determine the symmetry axis, we use Theorem 2. For $d=3$, we use existing standard 3D models dataset \cite{shilane2004princeton}. In order to calculate the symmetry axis in an image using the proposed approach, we first find the set of corner points \cite{harris1988combined}. This set may contain the corners not lying on the symmetric object. Therefore, we apply the proposed approach with RANSAC \cite{fischler1981random}. We compare the proposed results with  the results of two descriptor based methods \cite{loy2006detecting} and \cite{atadjanov2016reflection}. We evaluate on real and synthetic images containing single symmetric object from the dataset \cite{rauschert2011symmetry1}. In TABLE \ref{tab1}, we present the precision and the recall rates. We observe that for synthetic images, the precision rate is very high for the proposed approach because most of the corner points lie on the symmetric object. Whereas, in real images, the set of corner points contains many outlier corners which leads to the degraded performance. Precision rates for the proposed approach are higher than that for the methods \cite{loy2006detecting} and \cite{atadjanov2016reflection}. The recall rates are better than that of the method \cite{loy2006detecting} and comparable to that of the method \cite{atadjanov2016reflection}. This leads to the conclusion that symmetry detection can be performed even when the feature descriptors are not available. In Fig. \ref{fig:im_2d}, we show the results on the datasets \cite{cicconet2016convolutional,liu2013symmetry}, and \cite{rauschert2011symmetry1}. The last two images show the failure cases from the datasets \cite{rauschert2011symmetry1}. The reason could be that the pixels which are responsible for symmetry detection such as pixels on eyes and ear tips in the second image are not detected in the corner point detection step.\begin{table}[!htbp]
	\begin{center}
		\caption{Precision and recall rates for the methods \cite{loy2006detecting}, \cite{atadjanov2016reflection}, and the proposed approach on the dataset \cite{rauschert2011symmetry1}.}
		\label{tab1}
		\begin{tabular}{cc|c|c|c|c|c|}
			\cline{2-7}
			\multicolumn{1}{ c|  }{} &\multicolumn{3}{ c| }{Precision} & \multicolumn{3}{ c| }{Recall} \\ \cline{2-7}
			\multicolumn{1}{ c|  }{} &\cite{loy2006detecting} &\cite{atadjanov2016reflection}&Ours &\cite{loy2006detecting} &\cite{atadjanov2016reflection}&Ours\\ \cline{1-7}
			\multicolumn{1}{ |c| }{Real Images} & 0.21& 0.30& \textbf{0.42} & 0.75 & \textbf{0.95}& 0.93    \\ \cline{1-7}
			\multicolumn{1}{ |c| }{Synthetic Images}& 0.28 & 0.29& \textbf{0.73}& 0.93& \textbf{1.00}& 0.96 \\ \cline{1-7}
		\end{tabular}
	\end{center}
\end{table}\\

\begin{figure}[!htbp]
	\centering
	\stackunder{\epsfig{figure=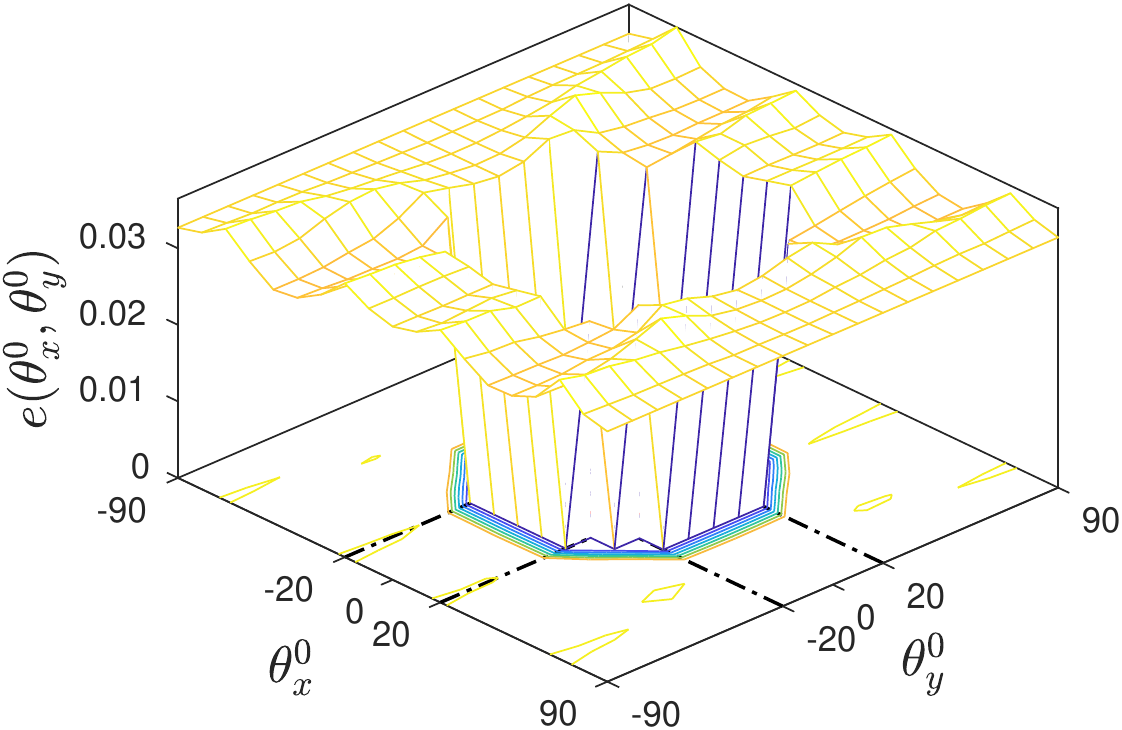,width=0.7\linewidth}}{(a)}
	\caption{Average error $e(\theta_x^0,\theta_y^0)$ vs the initialization angles $(\theta_x^0,\theta_y^0)$.}
	\label{fig:err}
\end{figure}

\textbf{ Influence of Different Initializations}.  We first create the following dataset of 3D point clouds. We create 5000 point clouds $\{S_i\}_{i=1}^{5000}$ with known ground-truth symmetries as discussed in \S 5.2. We keep 500 points in each point cloud. Without loss of generality, we choose the reflection symmetry plane such that it makes $90^\circ$ angle with $x$-axis and $y$-axis, i.e., the $x$-$y$ plane. For each point cloud, we initialize the variable $\mathbf{t}_i^0=\text{mean}(S_i)$ and $(\theta^0_x,\theta^0_y)$ on every point of the grid domain $\{-90^\circ,-80^\circ,\ldots,+80^\circ,+90^\circ\}\times\{-90^\circ,-80^\circ,\ldots,+80^\circ,+90^\circ\}$. We then run our approach and measure the error at the convergence $e_i(\theta_x^0,\theta_y^0)=\left\|\mathbf{R}^\star_x\mathbf{R}_y^\star\mathbf{E}(\mathbf{R}^\star_x\mathbf{R}_y^\star)^\top(\mathbf{X}_i-\mathbf{t}^\star\mathbf{e}^\top)+\mathbf{t}^\star\mathbf{e}^\top-\mathbf{X}_i\mathbf{P}^\star\right\|_\text{F}^2$ for each initialization $(\theta_x^0,\theta_y^0)$. Then, we find the average error $e(\theta_x^0,\theta_y^0)=\frac{1}{5000}\sum_{i=1}^{5000}e_i(\theta_x^0,\theta_y^0)$. Here, $\mathbf{R}_x$ and $\mathbf{R}_y$ are defined as follows.
$$
\mathbf{R}_x=\begin{bmatrix}1 & 0&0\\ 0&\cos\theta^0_x&-\sin\theta^0_x\\0&\sin\theta^0_x&\cos\theta^0_x\end{bmatrix},\mathbf{R}_y=\begin{bmatrix}\cos\theta^0_y&0&-\sin\theta^0_y\\0 &1&0\\\sin\theta^0_y&0&\cos\theta^0_y\end{bmatrix}.
$$
In Fig. \ref{fig:err}, we show the average error $e(\theta_x^0,\theta_y^0)$. We observe that if the initialization $(\theta_x^0,\theta_y^0)$ is far away from the global optimum $(0^\circ,0^\circ)$, then the error  is very high. As the distance between the initialization angles $(\theta_x^0,\theta_y^0)$ and the global optimum angles $(0^\circ,0^\circ)$ decrease, the error $e(\theta_x^0,\theta_y^0)$ remains approximately constant and suddenly drops to near zero after a particular distance. This indicates that, if the initialization angles are within a particular distance from the global optimum, then our approach always find the global optimum solution. This empirical result concurs with the result we already proved in Theorems 5 and 6. 
\section{Conclusion}\label{sec:CF}
In this work, we have developed a theory for establishing the correspondences between the mirror symmetric points in $\mathbb{R}^d$. We, further, determine the reflection symmetry transformation in a volumetric set of points in $\mathbb{R}^d$ containing a perturbed reflection symmetry pattern using optimization on Riemannian manifold. We have shown that our method is robust to a significant amount of perturbation and achieves 100\% accuracy for no perturbation. We have further shown the significance of this work by detecting reflection symmetry in real images and comparing with state-of-the-art methods. The proposed approach is particularly suitable for detecting reflection symmetry of objects in applications where obtaining a robust local descriptor is highly challenging. The linear assignment problem is a time consuming step which restricts us to apply it on the large point sets. However, a proper sampling method can be employed to reduce the size of the point set without losing the symmetry present in the point set. We believe that the fundamental theory and algorithm developed in this work will pave the way for researchers to exploit them for scenarios where estimating feature descriptors is a challenging task. 

Our approach detects single reflection symmetry plane of an object. Consider the third row of Fig. 11 in which there are multiple reflection symmetry planes present. In such cases, the detected reflection symmetry plane will be the one to which the initialized plane is the closest. For example, in the third row of Fig. 11, we have shown both the reflection symmetry planes detected depending on different initializations. This may not be a proper way of detecting multiple symmetries, though this is an interesting direction. We would like to extend our approach for the detection of multiple reflection symmetry planes of a symmetric object exhibiting multiple symmetries or a point cloud containing more than one symmetric objects. 

\section{Appendix}
\subsection*{A1. Euclidean gradient of the function $\bar{f}$ with respect to the variable  $\mathbf{t}$ (Equation (10))}

We write the cost function as follows.
\begin{eqnarray*}
	\bar{f}(\mathbf{R,t,P})&=&\|\mathbf{T}\mathbf{E}\mathbf{T}^\top(\mathbf{X}-\mathbf{t}\mathbf{e}^\top)-(\mathbf{X}\mathbf{P}-\mathbf{t}\mathbf{e}^\top)\|_\text{F}^2\\
	&=&\|(\mathbf{T}\mathbf{E}\mathbf{T}^\top\mathbf{X}-\mathbf{X}\mathbf{P})+(\mathbf{I}_d-\mathbf{T}\mathbf{E}\mathbf{T}^\top)\mathbf{t}\mathbf{e}^\top\|_\text{F}^2.
\end{eqnarray*}

We note  that
\begin{eqnarray*} (\mathbf{I}_d-\mathbf{T}\mathbf{E}\mathbf{T}^\top)^\top(\mathbf{I}_d-\mathbf{T}\mathbf{E}\mathbf{T}^\top)&=&2(\mathbf{I}_d-\mathbf{T}\mathbf{E}\mathbf{T}^\top).
\end{eqnarray*}
Therefore, we have (the terms which are not functions of $\mathbf{t}$ are not shown)
$$
\bar{f}(\mathbf{R,t,P})=\text{trace}(2\mathbf{e}\mathbf{t}^\top(\mathbf{I}_d-\mathbf{T}\mathbf{E}\mathbf{T}^\top)\mathbf{t}\mathbf{e}^\top
$$
$$+2(\mathbf{X}^\top\mathbf{T}\mathbf{E}\mathbf{T}^\top-\mathbf{P}^\top\mathbf{X}^\top)(\mathbf{I}_d-\mathbf{T}\mathbf{E}\mathbf{T}^\top)\mathbf{t}\mathbf{e}^\top).
$$

Now taking the derivative with respect to $\mathbf{t}$ we have,
$$
\nabla_{\mathbf{t}}\bar{f}=2(\mathbf{I}_d-\mathbf{T}\mathbf{E}\mathbf{T}^\top)\mathbf{t}\mathbf{e}^\top\mathbf{e}+2(\mathbf{e}^\top\mathbf{e}\mathbf{t}^\top(\mathbf{I}_d-\mathbf{T}\mathbf{E}\mathbf{T}^\top))^\top
$$
$$+2(\mathbf{e}^\top(\mathbf{X}^\top\mathbf{T}\mathbf{E}\mathbf{T}^\top-\mathbf{P}^\top\mathbf{X}^\top)(\mathbf{I}_d-\mathbf{T}\mathbf{E}\mathbf{T}^\top))^\top
$$
$$=4(\mathbf{I}_d-\mathbf{T}\mathbf{E}\mathbf{T}^\top)\mathbf{t}\mathbf{e}^\top\mathbf{e}+2(\mathbf{I}_d-\mathbf{T}\mathbf{E}\mathbf{T}^\top)(\mathbf{T}\mathbf{E}\mathbf{T}^\top\mathbf{X}-\mathbf{X}\mathbf{P})\mathbf{e}
$$

Here we have that $(\mathbf{I}_d-\mathbf{T}\mathbf{E}\mathbf{T}^\top)\mathbf{T}\mathbf{E}\mathbf{T}^\top=-(\mathbf{I}_d-\mathbf{T}\mathbf{E}\mathbf{T}^\top)$. Therefore,
\begin{eqnarray*}\nabla_{\mathbf{t}}\bar{f}&=&4(\mathbf{I}_d-\mathbf{T}\mathbf{E}\mathbf{T}^\top)\mathbf{t}\mathbf{e}^\top\mathbf{e}-2(\mathbf{I}_d-\mathbf{T}\mathbf{E}\mathbf{T}^\top)(\mathbf{X}+\mathbf{X}\mathbf{P})\mathbf{e}\\
	&=&2(\mathbf{I}_d-\mathbf{T}\mathbf{E}\mathbf{T}^\top)(2\mathbf{t}\mathbf{e}^\top\mathbf{e}-\mathbf{X}\mathbf{e}-\mathbf{X}\mathbf{P}\mathbf{e}).
\end{eqnarray*}
\subsection*{A2. Euclidean gradient of the function $\bar{f}$ with respect to the variable $\mathbf{R}_j$ (Equation (11))}
Let us consider the cost function as defined in equation (5) (in main manuscript):
$$\bar{f}(\mathbf{R,t,P})=\|\big(\prod\limits_{u=1}^{d-1}\mathbf{R}_u\big)\mathbf{E}\big(\prod\limits_{u=1}^{d-1}\mathbf{R}_u\big)^\top(\mathbf{X}-\mathbf{t}\mathbf{e}^\top)+\mathbf{t}\mathbf{e}^\top-\mathbf{X}\mathbf{P}\|_\text{F}^2.$$
Now, let us define  $\mathbf{T}=\prod\limits_{u=1}^{d-1}\mathbf{R}_u$, $\mathbf{U}=\mathbf{X}-\mathbf{t}\mathbf{e}^\top$ and $\mathbf{V}=\mathbf{X}\mathbf{P}-\mathbf{t}\mathbf{e}^\top$.

Then the cost function becomes.
\begin{eqnarray*}
	\bar{f}(\mathbf{R,t,P})&=&\|\mathbf{T}\mathbf{E}\mathbf{T}^\top\mathbf{U}-\mathbf{V}\|_\text{F}^2\\
	&=&\text{trace}((\mathbf{T}\mathbf{E}\mathbf{T}^\top\mathbf{U}-\mathbf{V})^\top(\mathbf{T}\mathbf{E}\mathbf{T}^\top\mathbf{U}-\mathbf{V}))\\
	&=&\text{trace}((\mathbf{U}^\top\mathbf{T}\mathbf{E}\mathbf{T}^\top-\mathbf{V}^\top)(\mathbf{T}\mathbf{E}\mathbf{T}^\top\mathbf{U}-\mathbf{V}))\\
	&=&\text{trace}(\mathbf{U}^\top\mathbf{T}\mathbf{E}\mathbf{T}^\top\mathbf{T}\mathbf{E}\mathbf{T}^\top\mathbf{U}-2\mathbf{U}^\top\mathbf{T}\mathbf{E}\mathbf{T}^\top\mathbf{V}\\
	&&+\mathbf{V}^\top\mathbf{V})
\end{eqnarray*}

Here we note that $\mathbf{T}\mathbf{E}\mathbf{T}^\top\mathbf{T}\mathbf{E}\mathbf{T}^\top=\mathbf{I}_d$, therefore
\begin{eqnarray*}
	\bar{f}(\mathbf{R,t,P})&=&\text{trace}(\mathbf{U}^\top\mathbf{U}-2\mathbf{U}^\top\mathbf{T}\mathbf{E}\mathbf{T}^\top\mathbf{V}+\mathbf{V}^\top\mathbf{V}).
\end{eqnarray*}
Now taking the classical gradient of $\bar{f}$ with respect to $\mathbf{R}_j$ we have. (We follow \cite{petersen2008matrix} for the necessary properties.)
$$
\frac{\partial \bar{f}}{\partial \mathbf{R}_j}=-2\frac{\partial}{\partial \mathbf{R}_j}\text{trace}(\mathbf{U}^\top\big(\prod\limits_{u=1}^{d-1}\mathbf{R}_u\big)\mathbf{E}\big(\prod\limits_{u=1}^{d-1}\mathbf{R}_u\big)^\top\mathbf{V})
$$
$$=-2\frac{\partial}{\partial \mathbf{R}_j}\text{trace}(\mathbf{U}^\top\big(\prod\limits_{u=1}^{j-1}\mathbf{R}_u\big)\mathbf{R}_j\big(\prod\limits_{u=j+1}^{d-1}\mathbf{R}_u\big)\mathbf{E}
$$
$$\big(\prod\limits_{j+1}^{d-1}\mathbf{R}_u\big)^\top\mathbf{R}_j^\top(\prod\limits_{u=1}^{j-1}\mathbf{R}_u)^\top\mathbf{V})$$
$$
=-2\big(\big(\prod\limits_{j+1}^{d-1}\mathbf{R}_u\big)\mathbf{E}\big(\prod\limits_{u=1}^{d-1}\mathbf{R}_u\big)^\top\mathbf{V}\mathbf{U}^\top\big(\prod\limits_{u=1}^{j-1}\mathbf{R}_u\big)\big)^\top
$$
$$-2\big(\prod\limits_{u=1}^{j-1}\mathbf{R}_u\big)^\top\mathbf{V}\mathbf{U}^\top\big(\prod\limits_{u=1}^{d-1}\mathbf{R}_u\big)\mathbf{E}\big(\prod\limits_{u=j+1}^{d-1}\mathbf{R}_u\big)^\top
$$
$$=-2\big(\prod\limits_{u=1}^{j-1}\mathbf{R}_u\big)^\top\mathbf{U}\mathbf{V}^\top\big(\prod\limits_{u=1}^{d-1}\mathbf{R}_u\big)\mathbf{E}\big(\prod\limits_{j+1}^{d-1}\mathbf{R}_u\big)^\top
$$
$$-2\big(\prod\limits_{u=1}^{j-1}\mathbf{R}_u\big)^\top\mathbf{V}\mathbf{U}^\top\big(\prod\limits_{u=1}^{d-1}\mathbf{R}_u\big)\mathbf{E}\big(\prod\limits_{u=j+1}^{d-1}\mathbf{R}_u\big)^\top
$$
$$=-2\big(\prod\limits_{u=1}^{j-1}\mathbf{R}_u\big)^\top(\mathbf{U}\mathbf{V}^\top+\mathbf{V}\mathbf{U}^\top)\big(\prod\limits_{u=1}^{d-1}\mathbf{R}_u\big)\mathbf{E}\big(\prod\limits_{u=j+1}^{d-1}\mathbf{R}_u\big)^\top
$$
$$=-2\big(\prod\limits_{u=1}^{j-1}\mathbf{R}_u\big)^\top\mathbf{A}\big(\prod\limits_{u=1}^{d-1}\mathbf{R}_u\big)\mathbf{E}\big(\prod\limits_{u=j+1}^{d-1}\mathbf{R}_u\big)^\top
$$

Where 
\begin{eqnarray*}
	\mathbf{A}&=&(\mathbf{V}\mathbf{U}^\top+\mathbf{U}\mathbf{V}^\top)\\
	&=&(\mathbf{X}\mathbf{P}-\mathbf{t}\mathbf{e}^\top)(\mathbf{X}-\mathbf{t}\mathbf{e}^\top)^\top+(\mathbf{X}-\mathbf{t}\mathbf{e}^\top)(\mathbf{X}\mathbf{P}-\mathbf{t}\mathbf{e}^\top)^\top.\end{eqnarray*}
\subsection*{A3. The Riemannian gradient of the function $f$ with respect to the variable  $\mathbf{t}$ (Equation (12))}
Using the definition, as defined in main paper, of Riemannian gradient $\boldsymbol{\xi}_\mathbf{t}(\mathbf{t})$ of the function $f$ with respect to the variable $\mathbf{t}$ we have  
$$
\boldsymbol{\xi}_\mathbf{t}(\mathbf{t})=\mathbb{P}_{\mathbf{t}}(\nabla_{\mathbf{t}}\bar{f})=\nabla_{\mathbf{t}}\bar{f}$$
$$=2\big(\mathbf{I}_d-\big(\prod_{u=1}^{d-1}\mathbf{R}_u\big)\mathbf{E}\big(\prod_{u=1}^{d-1}\mathbf{R}_u\big)^\top\big)(2n\mathbf{t}-\mathbf{X}\mathbf{e}-\mathbf{X}\mathbf{P}\mathbf{e}).
$$
\subsection*{A4. The Riemannian gradient of the function $f$ with respect to the variable $\mathbf{R}_j$ (Equation (13))}

Using the definition, as defined in main paper, of Riemannian gradient $\boldsymbol{\xi}_{\mathbf{R}_j}(\mathbf{R}_j)$ of the function $f$ with respect to the variable $\mathbf{R}_j$ we have  
$$
\boldsymbol{\xi}_{\mathbf{R}_j}(\mathbf{R}_j)=\mathbb{P}_{\mathbf{R}_j}(\nabla_{\mathbf{R}_j}\bar{f})=\mathbf{R}_j\text{skew}(\mathbf{R}_j^\top\nabla_{\mathbf{R}_j}\bar{f}).
$$
$$
\boldsymbol{\xi}_{\mathbf{R}_j}(\mathbf{R}_j)=\mathbf{R}_j\text{skew}(\mathbf{R}_j^\top\nabla_{\mathbf{R}_j}\bar{f}).
$$
$$
\mathbf{R}_j^\top\nabla_{\mathbf{R}_j}\bar{f}=-2\big(\prod\limits_{u=1}^{j}\mathbf{R}_u\big)^\top\mathbf{A}\big(\prod\limits_{u=1}^{d-1}\mathbf{R}_u)\mathbf{E}\big(\prod\limits_{u=j+1}^{d-1}\mathbf{R}_u\big)^\top
$$
$$
\nabla_{\mathbf{R}_j}\bar{f}^\top\mathbf{R}_j=-2\big(\prod\limits_{u=j+1}^{d-1}\mathbf{R}_u\big)\mathbf{E}\big(\prod\limits_{u=1}^{d-1}\mathbf{R}_u\big)^\top\mathbf{A}^\top\big(\prod\limits_{u=1}^{j}\mathbf{R}_u\big).
$$
Therefore,

$$
\nonumber\boldsymbol{\xi}_{\mathbf{R}_j}(\mathbf{R}_j)=\mathbf{R}_j\frac{\mathbf{R}_j^\top\nabla_{\mathbf{R}_j}\bar{f}-\nabla_{\mathbf{R}_j}\bar{f}^\top\mathbf{R}_j}{2}
$$
$$=-\mathbf{R}_j\big(\prod\limits_{u=1}^{j}\mathbf{R}_u\big)^\top\mathbf{A}\big(\prod\limits_{u=1}^{d-1}\mathbf{R}_u\big)\mathbf{E}\big(\prod\limits_{u=j+1}^{d-1}\mathbf{R}_u\big)^\top
$$
\begin{equation}
+\mathbf{R}_j\big(\prod\limits_{u=j+1}^{d-1}\mathbf{R}_u\big)\mathbf{E}\big(\prod\limits_{u=1}^{d-1}\mathbf{R}_u\big)^\top\mathbf{A}^\top\big(\prod\limits_{u=1}^{j}\mathbf{R}_u\big).
\label{eq:20}
\end{equation}
\subsection*{A5. The Riemannian Hessian of the function $f$ with respect to $\mathbf{R}_j$ (Equation (14))}
Next, we determine the Riemannian Hessian of the function $f$. In order to determine the $j^\text{th}$ component $\text{Hess}_{\mathbf{R}_j}(f(\mathbf{R,t}))[\mathbf{R}_j\Omega_j]=\mathbb{P}_{\mathbf{R}_j}(\text{D} \boldsymbol{\xi}_{\mathbf{R}_j}(\mathbf{R}_j)[\mathbf{R}_j\mathbf{\Omega}_j])$, of the Riemannian Hessian, we first find the classical derivative $ \text{D} \boldsymbol{\xi}_{\mathbf{R}_j}(\mathbf{R}_j)[\mathbf{R}_j\mathbf{\Omega}_j]$ of the Riemannian gradient $\boldsymbol{\xi}_{\mathbf{R}_j}(\mathbf{R}_j)$ in the direction $\mathbf{R}_j\mathbf{\Omega}_j$ and then we apply the projection operator $\mathbb{P}_{\mathbf{R}_j}$. Now using Equation \ref{eq:20} we have
$$
\boldsymbol{\xi}_{\mathbf{R}_j}(\mathbf{R}_j)=-\mathbf{R}_j\big(\prod\limits_{u=1}^{j}\mathbf{R}_u\big)^\top\mathbf{A}\big(\prod\limits_{u=1}^{d-1}\mathbf{R}_u\big)\mathbf{E}\big(\prod\limits_{u=j+1}^{d-1}\mathbf{R}_u\big)^\top
$$
$$
+\mathbf{R}_j\big(\prod\limits_{u=j+1}^{d-1}\mathbf{R}_u\big)\mathbf{E}\big(\prod\limits_{u=1}^{d-1}\mathbf{R}_u\big)^\top\mathbf{A}^\top\big(\prod\limits_{u=1}^{j}\mathbf{R}_u\big)
$$
$$=-\mathbf{R}_j\mathbf{R}_j^\top\big(\prod\limits_{u=1}^{j-1}\mathbf{R}_u\big)^\top\mathbf{A}\big(\prod\limits_{u=1}^{j-1}\mathbf{R}_u\big)\mathbf{R}_j\big(\prod\limits_{j+1}^{d-1}\mathbf{R}_u\big)\mathbf{E}\big(\prod\limits_{j+1}^{d-1}\mathbf{R}_u\big)^\top
$$
$$+\mathbf{R}_j\big(\prod\limits_{u=j+1}^{d-1}\mathbf{R}_u\big)\mathbf{E}\big(\prod\limits_{u=j+1}^{d-1}\mathbf{R}_u\big)^\top\mathbf{R}_j^\top\big(\prod\limits_{u=1}^{j-1}\mathbf{R}_u\big)^\top\mathbf{A}^\top\big(\prod\limits_{u=1}^{j-1}\mathbf{R}_u\big)\mathbf{R}_j
$$
$$=-\mathbf{R}_j\mathbf{R}_j^\top\mathbf{B}_2\mathbf{R}_j\mathbf{B}_1+\mathbf{R}_j\mathbf{B}_1\mathbf{R}_j^\top\mathbf{B}_2\mathbf{R}_j.
$$
Here, $$\mathbf{B}_1=\big(\prod\limits_{u=j+1}^{d-1}\mathbf{R}_u)\mathbf{E}\big(\prod\limits_{u=j+1}^{d-1}\mathbf{R}_u)^\top,
$$
$$\mathbf{B}_2=\big(\prod\limits_{u=1}^{j-1}\mathbf{R}_u\big)^\top\mathbf{A}^\top\big(\prod\limits_{u=1}^{j-1}\mathbf{R}_u\big).$$

Now 
$$
\text{D} \boldsymbol{\xi}_{\mathbf{R}_j}(\mathbf{R}_j)[\mathbf{R}_j\mathbf{\Omega}_j]=\frac{d}{dt}\boldsymbol{\xi}_{\mathbf{R}_j}(\mathbf{R}_j+t\mathbf{R}_j\mathbf{\Omega}_j)\mid_{t=0}
$$
$$
=\frac{d}{dt}(-(\mathbf{R}_j+t\mathbf{R}_j\mathbf{\Omega}_j)(\mathbf{R}_j+t\mathbf{R}_j\mathbf{\Omega}_j)^\top\mathbf{B}_2(\mathbf{R}_j+t\mathbf{R}_j\mathbf{\Omega}_j)\mathbf{B}_1)\mid_{t=0}
$$
$$+\frac{d}{dt}((\mathbf{R}_j+t\mathbf{R}_j\mathbf{\Omega}_j)\mathbf{B}_1(\mathbf{R}_j+t\mathbf{R}_j\mathbf{\Omega}_j)^\top\mathbf{B}_2(\mathbf{R}_j+t\mathbf{R}_j\mathbf{\Omega}_j))\mid_{t=0}.
$$
The first term is equal to

$$
-\mathbf{R}_j\mathbf{R}_j^\top\mathbf{B}_2\mathbf{R}_j\mathbf{\Omega}_j\mathbf{B}_1.
$$
The second term is equal to

$$
\mathbf{R}_j\mathbf{B}_1\mathbf{R}_j^\top\mathbf{B}_2\mathbf{R}_j\mathbf{\Omega}_j-\mathbf{R}_j\mathbf{B}_1\mathbf{\Omega}_j\mathbf{R}_j^\top\mathbf{B}_2\mathbf{R}_j+\mathbf{R}_j\mathbf{\Omega}_j\mathbf{B}_1\mathbf{R}_j^\top\mathbf{B}_2\mathbf{R}_j.
$$
Therefore,
$$
\text{D} \boldsymbol{\xi}_{\mathbf{R}_j}(\mathbf{R}_j)[\mathbf{R}_j\mathbf{\Omega}_j]=-\mathbf{R}_j\mathbf{R}_j^\top\mathbf{B}_2\mathbf{R}_j\mathbf{\Omega}_j\mathbf{B}_1+
(\mathbf{R}_j\mathbf{B}_1\mathbf{R}_j^\top\mathbf{B}_2\mathbf{R}_j\mathbf{\Omega}_j$$
$$-\mathbf{R}_j\mathbf{B}_1\mathbf{\Omega}_j\mathbf{R}_j^\top\mathbf{B}_2\mathbf{R}_j+\mathbf{R}_j\mathbf{\Omega}_j\mathbf{B}_1\mathbf{R}_j^\top\mathbf{B}_2\mathbf{R}_j)
\mathbf{R}_j^\top\text{D}\boldsymbol{\xi}_{\mathbf{R}_j}(\mathbf{R}_j)[\mathbf{R}_j\mathbf{\Omega}_j]$$
$$=-\mathbf{R}_j^\top\mathbf{R}_j\mathbf{R}_j^\top\mathbf{B}_2\mathbf{R}_j\mathbf{\Omega}_j\mathbf{B}_1+\mathbf{R}_j^\top(\mathbf{R}_j\mathbf{B}_1\mathbf{R}_j^\top\mathbf{B}_2\mathbf{R}_j\mathbf{\Omega}_j
$$
$$-\mathbf{R}_j\mathbf{B}_1\mathbf{\Omega}_j\mathbf{R}_j^\top\mathbf{B}_2\mathbf{R}_j+\mathbf{R}_j\mathbf{\Omega}_j\mathbf{B}_1\mathbf{R}_j^\top\mathbf{B}_2\mathbf{R}_j)$$
$$=-\mathbf{R}_j^\top\mathbf{B}_2\mathbf{R}_j\mathbf{\Omega}_j\mathbf{B}_1+
(\mathbf{B}_1\mathbf{R}_j^\top\mathbf{B}_2\mathbf{R}_j\mathbf{\Omega}_j$$
$$-\mathbf{B}_1\mathbf{\Omega}_j\mathbf{R}_j^\top\mathbf{B}_2\mathbf{R}_j+\mathbf{\Omega}_j\mathbf{B}_1\mathbf{R}_j^\top\mathbf{B}_2\mathbf{R}_j)
(\text{D}\boldsymbol{\xi}_{\mathbf{R}_j}(\mathbf{R}_j)[\mathbf{R}_j\mathbf{\Omega}_j])^\top\mathbf{R}_j
$$
$$=\mathbf{B}_1^\top\mathbf{\Omega}_j\mathbf{R}_j^\top\mathbf{B}_2^\top\mathbf{R}_j\mathbf{R}_j^\top\mathbf{R}_j+
(-\mathbf{\Omega}_j\mathbf{R}_j^\top\mathbf{B}_2^\top\mathbf{R}_j\mathbf{B}_1^\top\mathbf{R}_j^\top
$$
$$+\mathbf{R}_j^\top\mathbf{B}_2^\top\mathbf{R}_j\mathbf{\Omega}_j\mathbf{B}_1^\top\mathbf{R}_j^\top-\mathbf{R}_j^\top\mathbf{B}_2^\top\mathbf{R}_j\mathbf{B}_1^\top\mathbf{\Omega}_j\mathbf{R}_j^\top)\mathbf{R}_j
$$
$$=\mathbf{B}_1^\top\mathbf{\Omega}_j\mathbf{R}_j^\top\mathbf{B}_2^\top\mathbf{R}_j+
(-\mathbf{\Omega}_j\mathbf{R}_j^\top\mathbf{B}_2^\top\mathbf{R}_j\mathbf{B}_1^\top$$
$$+\mathbf{R}_j^\top\mathbf{B}_2^\top\mathbf{R}_j\mathbf{\Omega}_j\mathbf{B}_1^\top-\mathbf{R}_j^\top\mathbf{B}_2^\top\mathbf{R}_j\mathbf{B}_1^\top\mathbf{\Omega}_j)
$$

$$
\mathbf{R}_j^\top\text{D} \boldsymbol{\xi}_{\mathbf{R}_j}(\mathbf{R}_j)[\mathbf{R}_j\mathbf{\Omega}_j]-(\text{D} \boldsymbol{\xi}_{\mathbf{R}_j}(\mathbf{R}_j)[\mathbf{R}_j\mathbf{\Omega}_j])^\top\mathbf{R}_j$$
$$=-\mathbf{R}_j^\top\mathbf{B}_2\mathbf{R}_j\mathbf{\Omega}_j\mathbf{B}_1+\mathbf{B}_1\mathbf{R}_j^\top\mathbf{B}_2\mathbf{R}_j\mathbf{\Omega}_j-\mathbf{B}_1\mathbf{\Omega}_j\mathbf{R}_j^\top\mathbf{B}_2\mathbf{R}_j
$$
$$+\mathbf{\Omega}_j\mathbf{B}_1\mathbf{R}_j^\top\mathbf{B}_2\mathbf{R}_j-\mathbf{B}_1^\top\mathbf{\Omega}_j\mathbf{R}_j^\top\mathbf{B}_2^\top\mathbf{R}_j+
\mathbf{\Omega}_j\mathbf{R}_j^\top\mathbf{B}_2^\top\mathbf{R}_j\mathbf{B}_1^\top
$$
$$-\mathbf{R}_j^\top\mathbf{B}_2^\top\mathbf{R}_j\mathbf{\Omega}_j\mathbf{B}_1^\top+\mathbf{R}_j^\top\mathbf{B}_2^\top\mathbf{R}_j\mathbf{B}_1^\top\mathbf{\Omega}_j
$$
$$=\mathbf{B}_1[\mathbf{R}_j^\top\mathbf{B}_2^\top\mathbf{R}_j, \mathbf{\Omega}_j]-[\mathbf{R}_j^\top\mathbf{B}_2^\top\mathbf{R}_j, \mathbf{\Omega}_j]\mathbf{B}_1+
$$
$$
[\mathbf{\Omega}_j,\mathbf{B}_1]\mathbf{R}_j^\top\mathbf{B}_2\mathbf{R}_j-\mathbf{R}_j^\top\mathbf{B}_2^\top\mathbf{R}_j[\mathbf{\Omega}_j,\mathbf{B}_1]\\
$$
$$=[\mathbf{B}_1,[\mathbf{R}_j^\top\mathbf{B}_2\mathbf{R}_j, \mathbf{\Omega}_j]]+[[\mathbf{\Omega}_j,\mathbf{B}_1],\mathbf{R}_j^\top\mathbf{B}_2\mathbf{R}_j].
$$
Here $[.,.]$ is the Lie bracket and defined as $[\mathbf{U},\mathbf{V}]=\mathbf{UV}-\mathbf{VU}$ for any two matrices $\mathbf{U}$ and $\mathbf{V}$.

$$
\text{Hess}_{\mathbf{R}_j}(f(\mathbf{R,t}))[\mathbf{R}_j\Omega_j]=\mathbb{P}_{\mathbf{R}_j}(\text{D} \boldsymbol{\xi}_{\mathbf{R}_j}(\mathbf{R}_j)[\mathbf{R}_j\mathbf{\Omega}_j])
$$
$$=\mathbf{R}_j\text{skew}(\mathbf{R}_j^\top\text{D} \boldsymbol{\xi}_{\mathbf{R}_j}(\mathbf{R}_j)[\mathbf{R}_j\mathbf{\Omega}_j])
$$
$$=\frac{1}{2}\mathbf{R}_j(\mathbf{R}_j^\top\text{D} \boldsymbol{\xi}_{\mathbf{R}_j}(\mathbf{R}_j)[\mathbf{R}_j\mathbf{\Omega}_j]-(\text{D} \boldsymbol{\xi}_{\mathbf{R}_j}(\mathbf{R}_j)[\mathbf{R}_j\mathbf{\Omega}_j])^\top\mathbf{R}_j)
$$
$$=\frac{1}{2}\mathbf{R}_j(
[\mathbf{B}_1,[\mathbf{R}_j^\top\mathbf{B}_2\mathbf{R}_j, \mathbf{\Omega}_j]]+[[\mathbf{\Omega}_j,\mathbf{B}_1],\mathbf{R}_j^\top\mathbf{B}_2\mathbf{R}_j]).
$$

\subsection*{A6. The Riemannian Hessian of the function $f$ with respect to $\mathbf{t}$ (Equation (15))}
In a similar way, we determine the second component, $\mathbb{P}_{\mathbf{t}}(\text{D} \boldsymbol{\xi}_\mathbf{t}(\mathbf{t})[\boldsymbol{\eta}_\mathbf{t}])$, of the Riemannian Hessian. Since $\mathbb{R}^d$ is a vector space we have $\mathbb{P}_{\mathbf{t}}(\text{D} \boldsymbol{\xi}_\mathbf{t}(\mathbf{t})[\boldsymbol{\eta}_\mathbf{t}])=\text{D} \boldsymbol{\xi}_\mathbf{t}(\mathbf{t})[\boldsymbol{\eta}_\mathbf{t}]$

$$\text{D} \boldsymbol{\xi}_\mathbf{t}(\mathbf{t})[\boldsymbol{\eta}_\mathbf{t}]=\frac{d}{dq}\boldsymbol{\xi}_\mathbf{t}(\mathbf{t}+q\boldsymbol{\eta}_\mathbf{t})\mid_{q=0}
$$ 
$$
=4n\big(\mathbf{I}_d-\big(\prod_{u=1}^{d-1}\mathbf{R}_u\big)\mathbf{E}\big(\prod_{u=1}^{d-1}\mathbf{R}_u\big)^\top\big)\boldsymbol{\eta}_\mathbf{t}.
$$
Therefore
\begin{equation}
\text{Hess}_{\mathbf{t}}(f(\mathbf{R,t}))[\boldsymbol{\eta}_\mathbf{t}]=4n\big(\mathbf{I}_d-\big(\prod_{u=1}^{d-1}\mathbf{R}_u\big)\mathbf{E}\big(\prod_{u=1}^{d-1}\mathbf{R}_u\big)^\top\big)\boldsymbol{\eta}_\mathbf{t}.
\label{eq:23}
\end{equation}

\subsection*{A7. Steps of Theorem 5}
Showing the  fact $$\text{trace}(\mathbf{\Omega}_j^\top\mathbf{R}_j^\top\mathbf{H}[\mathbf{R}_j\mathbf{\Omega}_j])=4\text{trace}(\mathbf{R}_j^\top\mathbf{B}_2\mathbf{R}_j\mathbf{\Omega}_j(\mathbf{B}_1\mathbf{\Omega}_j-\mathbf{\Omega}_j\mathbf{B}_1)).$$
Now
$$
\text{trace}(\mathbf{\Omega}_j^\top\mathbf{R}_j^\top\mathbf{H}[\mathbf{R}_j\mathbf{\Omega}_j])=\text{trace}(-\mathbf{\Omega}_j^\top\mathbf{R}_j^\top\mathbf{B}_2\mathbf{R}_j\mathbf{\Omega}_j\mathbf{B}_1
$$
$$+\mathbf{\Omega}_j^\top\mathbf{B}_1\mathbf{R}_j^\top\mathbf{B}_2\mathbf{R}_j\mathbf{\Omega}_j-\mathbf{\Omega}_j^\top\mathbf{B}_1\mathbf{\Omega}_j\mathbf{R}_j^\top\mathbf{B}_2\mathbf{R}_j
+\mathbf{\Omega}_j^\top\mathbf{\Omega}_j\mathbf{B}_1\mathbf{R}_j^\top\mathbf{B}_2\mathbf{R}_j$$
$$-\mathbf{\Omega}_j^\top\mathbf{B}_1^\top\mathbf{\Omega}_j\mathbf{R}_j^\top\mathbf{B}_2^\top\mathbf{R}_j+\mathbf{\Omega}_j^\top\mathbf{\Omega}_j\mathbf{R}_j^\top\mathbf{B}_2^\top\mathbf{R}_j\mathbf{B}_1^\top
$$
$$-\mathbf{\Omega}_j^\top\mathbf{R}_j^\top\mathbf{B}_2^\top\mathbf{R}_j\mathbf{\Omega}_j\mathbf{B}_1^\top+\mathbf{\Omega}_j^\top\mathbf{R}_j^\top\mathbf{B}_2^\top\mathbf{R}_j\mathbf{B}_1^\top\mathbf{\Omega}_j)
$$
$$=\text{trace}(\mathbf{\Omega}_j\mathbf{R}_j^\top\mathbf{B}_2\mathbf{R}_j\mathbf{\Omega}_j\mathbf{B}_1-\mathbf{\Omega}_j\mathbf{B}_1\mathbf{R}_j^\top\mathbf{B}_2\mathbf{R}_j\mathbf{\Omega}_j
$$
$$+\mathbf{\Omega}_j\mathbf{B}_1\mathbf{\Omega}_j\mathbf{R}_j^\top\mathbf{B}_2\mathbf{R}_j-\mathbf{\Omega}_j\mathbf{\Omega}_j\mathbf{B}_1\mathbf{R}_j^\top\mathbf{B}_2\mathbf{R}_j
$$
$$+\mathbf{\Omega}_j\mathbf{B}_1\mathbf{\Omega}_j\mathbf{R}_j^\top\mathbf{B}_2\mathbf{R}_j-
\mathbf{\Omega}_j\mathbf{\Omega}_j\mathbf{R}_j^\top\mathbf{B}_2\mathbf{R}_j\mathbf{B}_1
$$
$$+\mathbf{\Omega}_j\mathbf{R}_j^\top\mathbf{B}_2\mathbf{R}_j\mathbf{\Omega}_j\mathbf{B}_1-\mathbf{\Omega}_j\mathbf{R}_j^\top\mathbf{B}_2\mathbf{R}_j\mathbf{B}_1\mathbf{\Omega}_j)
$$
$$
=\text{trace}(4\mathbf{R}_j^\top\mathbf{B}_2\mathbf{R}_j\mathbf{\Omega}_j\mathbf{B}_1\mathbf{\Omega}_j-2\mathbf{R}_j^\top\mathbf{B}_2\mathbf{R}_j\mathbf{\Omega}_j\mathbf{\Omega}_j\mathbf{B}_1
$$
$$-2\mathbf{R}_j^\top\mathbf{B}_2\mathbf{R}_j\mathbf{B}_1\mathbf{\Omega}_j\mathbf{\Omega}_j)
$$
Since, $$\text{trace}(\mathbf{R}_j^\top\mathbf{B}_2\mathbf{R}_j\mathbf{B}_1\mathbf{\Omega}_j\mathbf{\Omega}_j)=\text{trace}((\mathbf{B}_1\mathbf{\Omega}_j\mathbf{\Omega}_j)^\top\mathbf{R}_j^\top\mathbf{B}_2\mathbf{R}_j)
$$
$$=\text{trace}(\mathbf{\Omega}_j\mathbf{\Omega}_j\mathbf{B}_1\mathbf{R}_j^\top\mathbf{B}_2\mathbf{R}_j)
$$
$$=\text{trace}(\mathbf{R}_j^\top\mathbf{B}_2\mathbf{R}_j\mathbf{\Omega}_j\mathbf{\Omega}_j\mathbf{B}_1),$$

we have
$$
\text{trace}(\mathbf{\Omega}_j^\top\mathbf{R}_j^\top\mathbf{H}[\mathbf{R}_j\mathbf{\Omega}_j])=\text{trace}(4\mathbf{R}_j^\top\mathbf{B}_2\mathbf{R}_j\mathbf{\Omega}_j\mathbf{B}_1\mathbf{\Omega}_j
$$
$$-4\mathbf{R}_j^\top\mathbf{B}_2\mathbf{R}_j\mathbf{\Omega}_j\mathbf{\Omega}_j\mathbf{B}_1)
$$
$$
=4\text{ trace}(\mathbf{R}_j^\top\mathbf{B}_2\mathbf{R}_j(\mathbf{\Omega}_j\mathbf{B}_1\mathbf{\Omega}_j-\mathbf{\Omega}_j\mathbf{\Omega}_j\mathbf{B}_1)).
$$

\end{document}